\pdfoutput=1

\documentclass[11pt]{article}

\newcommand{\cmulogo}{%
  \raisebox{-0.15ex}{\includegraphics[height=1em]{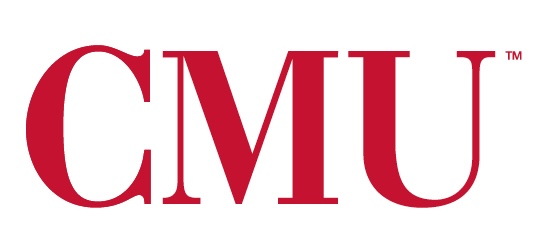}}%
  \hspace{0.15em}%
  \raisebox{0.1ex}{\includegraphics[height=0.7em]{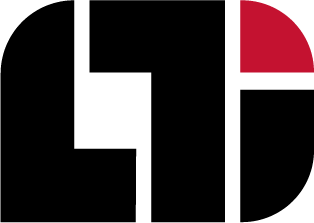}}%
}

\usepackage[final]{acl}

\usepackage{times}
\usepackage{latexsym}

\usepackage[T1]{fontenc}

\usepackage[utf8]{inputenc}

\usepackage{microtype}

\usepackage{inconsolata}

\usepackage{graphicx}

\usepackage{amssymb,bm,mathtools,color}

\usepackage[normalem]{ulem}
\usepackage{enumitem}

\usepackage{booktabs}
\usepackage{multirow}
\usepackage{tabularx}
\usepackage{multicol}
\usepackage{colortbl}
\usepackage{hhline}
\usepackage{stfloats}
\usepackage{subcaption}
\usepackage{threeparttable}
\usepackage{pifont}
\usepackage{amsmath}
\usepackage{amssymb}
\usepackage{ifthen}
\usepackage{dsfont}

\usepackage{dcolumn}
\usepackage{siunitx}
\usepackage{makecell,rotating,array}

\usepackage{listings}
\usepackage{xcolor}
\usepackage{ltablex}
\usepackage{longtable}
\usepackage{multicol}
\usepackage{wrapfig}
\usepackage{threeparttable}

\usepackage{tikz}
\usetikzlibrary{bayesnet}
\usepackage[linguistics]{forest}

\usepackage[ruled,vlined]{algorithm2e}

\usepackage{caption}
\usepackage{subcaption}

\usepackage{cancel}

\usepackage{xspace}
\usepackage{comment}
\usepackage{color,soul}

\usepackage{float}

\usepackage[noabbrev,capitalize]{cleveref} 

\crefname{page}{page}{pages}
\crefname{footnote}{footnote}{footnotes}   
\crefname{equation}{equation}{equations}   
\crefname{line}{line}{lines}               
\crefrangeformat{line}{lines #3#1#4--#5#2#6}
\crefname{lstlsting}{Listing}{Listings}   
\crefname{section}{\S}{\S\S}
\Crefname{section}{\S}{\S\S}    
\crefformat{section}{#2\S#1#3}  
\Crefformat{section}{#2\S#1#3}
\crefrangeformat{section}{\S\S#3#1#4--#5#2#6}
\Crefrangeformat{section}{\S\S#3#1#4--#5#2#6}
\crefmultiformat{section}{\S#2#1#3}{ and~\S#2#1#3}{, \S#2#1#3}{ and~\S#2#1#3}
\Crefmultiformat{section}{\S#2#1#3}{ and~\S#2#1#3}{, \S#2#1#3}{ and~\S#2#1#3}
\crefrangemultiformat{section}{\S\S#3#1#4--#5#2#6}{ and~\S\S#3#1#4--#5#2#6}{, \S\S#3#1#4--#5#2#6}{ and~\S\S#3#1#4--#5#2#6}
\Crefrangemultiformat{section}{\S\S#3#1#4--#5#2#6}{ and~\S\S#3#1#4--#5#2#6}{, \S\S#3#1#4--#5#2#6}{ and~\S\S#3#1#4--#5#2#6}

\newcommand{\eg}{\emph{e.g.}, }
\newcommand{\ie}{\emph{i.e.}}

\usepackage[]{todonotes}

\usepackage[]{todonotes}  
\makeatletter
\newcommand*\iftodonotes{\if@todonotes@disabled\expandafter\@secondoftwo\else\expandafter\@firstoftwo\fi}  
\makeatother

\title{Model Internal Sleuthing: Finding Lexical Identity and Inflectional Features in Modern Language Models}

\author{\
Michael Li$^{\text{\cmulogo}}$%
\ \ \ \ \ 
\textbf{Nishant Subramani$^{\text{\cmulogo}}$}%
\\
$^\text{\cmulogo}$Carnegie Mellon University - Language Technologies Institute \\
$\texttt{\{ml6, nishant2\}@cs.cmu.edu}$}

\begin{document}
\maketitle
\begin{abstract}
Large transformer-based language models dominate modern NLP, yet our understanding of how they encode linguistic information relies primarily on studies of early models like \texttt{BERT} and \texttt{GPT-2}.
We systematically probe 25 models from \texttt{BERT Base} to \texttt{Qwen2.5-7B} focusing on two linguistic properties: lexical identity and inflectional features across 6 diverse languages. 
We find a consistent pattern: inflectional features are linearly decodable throughout the model, while lexical identity is prominent early but increasingly weakens with depth.
Further analysis of the representation geometry reveals that models with aggressive mid-layer dimensionality compression show reduced steering effectiveness in those layers, despite probe accuracy remaining high. 
Pretraining analysis shows that inflectional structure stabilizes early while lexical identity representations continue evolving.
Taken together, our findings suggest that transformers maintain inflectional features across layers, while trading off lexical identity for compact, predictive representations.
Our code is available at \url{https://github.com/ml5885/model_internal_sleuthing}
\end{abstract}

\section{Introduction}

Large transformer-based language models (LMs) are widely used for tasks such as text generation, question answering, and code completion~\citep{workshop2023bloom176bparameteropenaccessmultilingual, groeneveld-etal-2024-olmo, grattafiori2024llama3herdmodels, hui2024qwen2}.
However, how these models internally represent linguistic information remains an active research area. Prior work suggests a hierarchical organization where different layers specialize in capturing distinct levels of linguistic structure~\citep{jawahar-etal-2019-bert, tenney-etal-2019-bert, rogers-etal-2020-primer}. 
However, these studies focus only on first-generation LMs such as \texttt{BERT} and \texttt{GPT-2}~\citep{devlin-etal-2019-bert, radford2019language}. 
Since then, language technology has transformed dramatically --- today's models differ in architecture (encoder-only, decoder-only, encoder-decoder), pretraining objectives (masked vs.~causal language modeling), training data volume (billions vs.~trillions of tokens), and post-training adaptation~\citep{brown2020languagemodelsfewshotlearners, groeneveld-etal-2024-olmo, lambert2025tulu3pushingfrontiers}.
We ask: where and how do modern LMs encode lexical identity and inflectional morphology, and how do these representations vary with model scale and architecture?

To answer these questions we systematically probe 25 pretrained models ranging from \texttt{BERT Base} to \texttt{Llama-3.1 8B}, spanning multiple architectures, sizes, and training regimes. We train simple classifiers at each layer to predict word-level lexical identity and inflectional features, and evaluate where this information emerges and how linearly accessible they are.
We focus on two linguistic properties: \emph{lexical identity} and \emph{inflectional features}, which help disentangle meaning from surface form.
Consider the words \textit{walk}, \textit{walked}, \textit{jump}, and \textit{jumped}.
Do language models group words by shared meaning (\textit{walk}, \textit{walked}) or by shared grammar (\textit{walked}, \textit{jumped})?
More broadly, where and how do LMs encode a word's lexeme and its inflectional features?

To test whether observed patterns generalize beyond English, we examine six typologically diverse languages: English, Chinese, German, French, Russian, and Turkish. We also investigate where lexical and inflectional information reside (attention heads vs.~residual streams), evaluate the impact of editing activations via steering vectors, and track when these representations emerge during pretraining. To the best of our knowledge, this is the first systematic analysis of how lexical identity and inflectional features are encoded across 25 modern language models spanning multiple architectures, scales, and training regimes in six typologically diverse languages. We find that:

\begin{enumerate}[leftmargin=1cm, itemsep=1pt, topsep=1pt]
    \item Lexical identity information is encoded prominently in early layers and becomes increasingly non-linear deeper in the network, whereas inflectional information remains linearly accessible across all layers.
    \item Across languages, we find that linguistic encoding strength varies with morphological typology, declining most sharply in Turkish.
    \item Lexical and inflectional information emerge early in pretraining and reside primarily in the residual stream; inflectional features occupy compact, steerable subspaces that enable effective interventions.
\end{enumerate}

\section{Probe Design and Metrics}
\label{sec:model_internal_sleuthing}

\begin{figure*}[t]
  \centering
  \includegraphics[width=\textwidth]{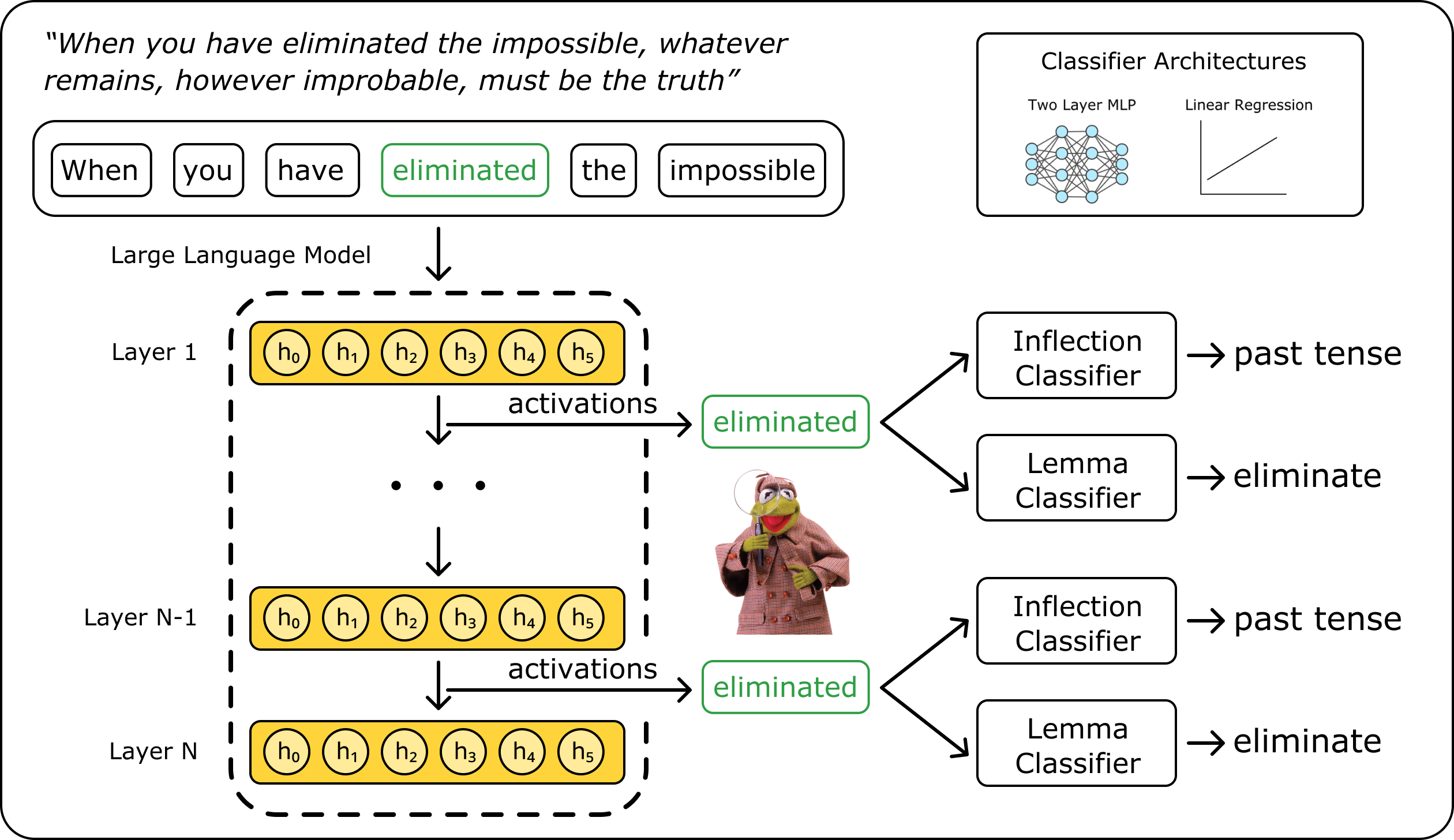}
  \caption{Overview of our probing methodology. We extract hidden state activations from each model layer for target words and train simple linear and shallow non-linear classifiers to predict word-level lexical identity and inflectional features. We compute selectivity using control labels and analyze how linear vs. non-linear accessibility varies with depth.}
  \label{fig:classifier_pipeline}
  \vspace{-1em}
\end{figure*}

We investigate how language models encode linguistic information using simple classifiers (\emph{probes}) trained on activations from individual layers. We train probes to predict each word's lexeme (\eg \textit{walk} as the base form of \textit{walked}) and its inflectional features (\eg plural, past tense).

\subsection{Probe architectures}

For each layer of a model we extract residual-stream representations for a target word and train two classifiers: a linear regression probe and non-linear multi-layer perceptron (MLP) probe. The linear probe measures how well information is linearly separable in the representation space, while the non-linear probe relaxes this linear assumption and searches for a non-linear decision boundary. Comparing these probes allows us to infer whether a property is encoded \emph{linearly} or \emph{non-linearly}. Training details are provided in Appendix~\ref{sec:probe_details}.

\subsubsection{Linear Regression Classifier}

Consistent with best practices for probing \citep{hewitt-liang-2019-designing, liu-etal-2019-linguistic}, we use a ridge-regularized linear regression classifier.  Given training representations $X_{\text{train}} \in \mathbb{R}^{m \times d}$ and one-hot encoded labels $Y_{\text{train}} \in \mathbb{R}^{m \times c}$, the optimal weight matrix $W \in \mathbb{R}^{d \times c}$ is obtained in closed form as:
\begin{equation}\label{eq:ridge}
    W = \bigl(X_{\text{train}}^\top X_{\text{train}} + \lambda I\bigr)^{-1} X_{\text{train}}^\top Y_{\text{train}},
\end{equation}

\noindent where $\lambda$ controls the strength of $\ell_2$ regularization and $I$ is the identity matrix.  Predictions on test representations $X_{\text{test}}$ are then given by $\hat{Y}_{\text{test}} = X_{\text{test}} W$.

\subsubsection{MLP Classifier}

To test for non-linear separability, we train a simple two-layer MLP with ReLU activation:
\begin{equation}
    \hat{Y} = \mathrm{softmax}\Bigl( \mathrm{ReLU}(X W_1) W_2 \Bigr),
\end{equation}
\noindent Here, $W_1 \in \mathbb{R}^{d \times h}$ and $W_2 \in \mathbb{R}^{h \times c}$ are learned weight matrices, $h$ is the hidden dimension (we use $h=64$), and biases are omitted for brevity. 

\subsection{Metrics}
We define two metrics for quantifying signal and nonlinearity across depth: selectivity and the linear separability gap.

\paragraph{Selectivity.} Probes may simply memorize training data rather than extracting true linguistic information from the representations. To account for this, we construct control tasks following \citet{hewitt-liang-2019-designing}, assigning each unique word form a random class label and training identical probes on these labels. We call this set the control set. We define selectivity at layer $\ell$ as the difference between real and control accuracies:
\begin{align}
\label{eq:selectivity}
\text{Sel}_\ell = \text{Acc}^{\text{real}}_\ell - \text{Acc}^{\text{control}}_\ell
\end{align}
Higher values mean the classifier is extracting true linguistic information rather than just memorizing.

\paragraph{Linear separability gap.}
To compare how much linguistic signal each probe type extracts, we compute the difference in selectivity:
\begin{align}
\label{eq:linear_separability_gap}
\text{Gap}_\ell = \text{Sel}^{\text{nonlin}}_\ell - \text{Sel}^{\text{linear}}_\ell
\end{align}
Negative gap values indicate that additional (MLP) probe capacity captures spurious correlations rather than linguistic structure.

\section{Experiments}
\label{sec:experiments}
Using the methodology introduced in Section~\cref{sec:model_internal_sleuthing}, we describe the components of our experimental setup: the datasets, model suite, and procedure for extracting token-level representations.

\subsection{Datasets}

For our analysis of lexical identity and inflectional features, we use Universal Dependencies corpora across six languages - English, Chinese, German, French, Russian, Turkish~\citep{nivre-etal-2016-universal}. We select GUM for English~\citep{Zeldes2017}, GSD for Chinese/German/French~\citep{mcdonald-etal-2013-universal, guillaume-etal-2019-conversion}, SynTagRus for Russian~\citep{droganova2018data}, and IMST for Turkish~\citep{sulubacak-etal-2016-universal}.~\footnote{See Appendix~\cref{sec:dataset_appendix} for complete details including dataset statistics, tokenization information, and visualizations for all languages}

\begin{table}[t]
  \renewcommand\arraystretch{0.95}
  \setlength{\tabcolsep}{3pt}
  \centering
  \resizebox{\linewidth}{!}{%
  \begin{tabular}{lrrr}
    \toprule
    \textbf{Model} & \textbf{Parameters} & \textbf{Pretraining Data} & \textbf{Layers} \\
    \midrule
    \textbf{Encoder-only} & & & \\
    \midrule
    \hspace{5pt}\texttt{BERT Base} & 110M & 12.6B tokens\footnotemark[1] & 12 \\
    \hspace{5pt}\texttt{BERT Large} & 340M & 12.6B tokens\footnotemark[1] & 24 \\
    \hspace{5pt}\texttt{DeBERTa V3 Large} & 418M & 32B tokens\footnotemark[1] & 24 \\
    \midrule
    \textbf{Decoder-only} & & & \\
    \midrule
    \hspace{5pt}\texttt{GPT 2 Small} & 124M & 8B tokens\footnotemark[1] & 12 \\
    \hspace{5pt}\texttt{GPT 2 Large} & 708M & 8B tokens\footnotemark[1] & 36 \\
    \hspace{5pt}\texttt{GPT 2 XL} & 1.5B & 8B tokens\footnotemark[1] & 48 \\ 
    \hspace{5pt}\texttt{Goldfish English 1000mb} & 124M & 200M tokens & 12 \\
    \hspace{5pt}\texttt{Goldfish Chinese 1000mb} & 124M & 200M tokens & 12 \\
    \hspace{5pt}\texttt{Goldfish German 1000mb} & 124M & 200M tokens & 12 \\
    \hspace{5pt}\texttt{Goldfish French 1000mb} & 124M & 200M tokens & 12 \\
    \hspace{5pt}\texttt{Goldfish Russian 1000mb} & 124M & 200M tokens & 12 \\
    \hspace{5pt}\texttt{Goldfish Turkish 1000mb} & 124M & 200M tokens & 12 \\
    \hspace{5pt}\texttt{Pythia 6.9b} & 6900M & 300B tokens & 32 \\
    \hspace{5pt}\texttt{Pythia 6.9b Tulu} & 6900M & 300B tokens & 32 \\
    \hspace{5pt}\texttt{OLMo 2 7B} & 7300M & 4T tokens & 32 \\
    \hspace{5pt}\texttt{OLMo 2 7B Instruct} & 7300M & 4T tokens & 32 \\
    \hspace{5pt}\texttt{Gemma 2 2B} & 2610M & 2T tokens & 26 \\
    \hspace{5pt}\texttt{Gemma 2 2B Instruct} & 2610M & 2T tokens & 26 \\
    \hspace{5pt}\texttt{Qwen2.5 1.5B} & 1540M & 18T tokens & 28 \\
    \hspace{5pt}\texttt{Qwen2.5 1.5B Instruct} & 1540M & 18T tokens & 28 \\
    \hspace{5pt}\texttt{Qwen2.5 7B} & 7620M & 18T tokens & 28 \\
    \hspace{5pt}\texttt{Qwen2.5 7B Instruct} & 7620M & 18T tokens & 28 \\
    \hspace{5pt}\texttt{Llama 3.1 8B} & 8000M & 15T tokens & 32 \\
    \hspace{5pt}\texttt{Llama 3.1 8B Instruct} & 8000M & 15T tokens & 32 \\
    \midrule
    \textbf{Encoder-Decoder} & & & \\
    \midrule
    \hspace{5pt}\texttt{mT5-base} & 580M & 1T tokens & 12 \\
    \bottomrule
  \end{tabular}
  }%
  \caption{Overview of models used in this study. $^1$Converted from GB to tokens using the approximation that 1GB of data is approximately 200M tokens in English~\citep{chang2024goldfishmonolinguallanguagemodels}.}
  \label{tab:models}
\end{table}

\subsection{Models}

We study a diverse set of pretrained transformer language models spanning different architectures, sizes, and training regimes.~\cref{tab:models} lists all models used in this study (see~\cref{tab:hf_id_mapping} for the HuggingFace identifiers).

For English, we evaluate 19 models: all models in~\cref{tab:models} except the five non-English \texttt{Goldfish} models and \texttt{mT5-base}. For the five non-English languages (Chinese, German, French, Russian, Turkish), we use a set of models with explicit coverage of each target language: the corresponding monolingual \texttt{Goldfish <Language> 1000mb} model~\citep{chang2024goldfishmonolinguallanguagemodels}, multilingual \texttt{Qwen2.5-1.5B} (and instruct), multilingual \texttt{Qwen2.5-7B} (and instruct)~\citep{qwen2.5}, and the multilingual \texttt{mT5-base} model~\citep{xue-etal-2021-mt5}. This ensures that we evaluate models on languages they were trained on while maintaining sufficient coverage.
\vspace{-0.3em}

\subsection{Representation Extraction}

Each input to the model is a complete sentence from the corpus. We tokenize inputs with model-specific tokenizers and run a forward pass to collect residual-stream activations at the target word position from every layer.\footnote{We also experiment with attention head outputs (see \cref{sec:analysis} and \cref{sec:attention_analysis_appendix})}

For words split into multiple subwords, we use the last subword's representation, since this is where prior work suggests word-level meaning lives~\citep{kaplan2025from, feucht2025the}.
\section{Lexical Identity and Inflectional Features}
\label{sec:lexemes_and_inflections}

\begin{figure*}[htbp]
  \centering
  \includegraphics[width=\textwidth]{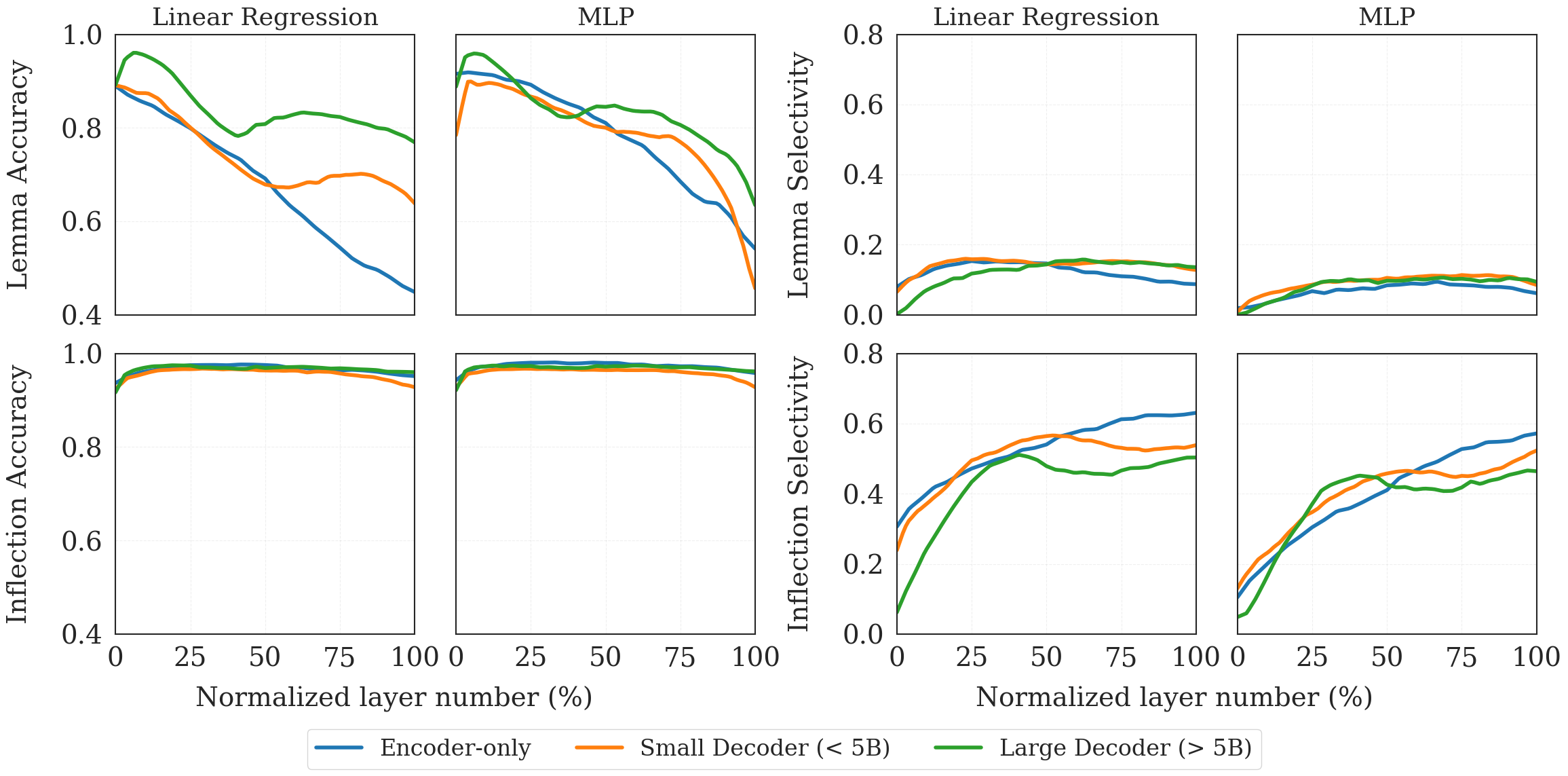}
  \caption{Lexeme and inflection probing results for English, \emph{averaged by model category}: encoder-only (\texttt{BERT}, \texttt{DeBERTa}), small decoder <5B (\texttt{GPT-2}, \texttt{Gemma-2-2B} (and instruct), \texttt{Qwen2.5-1.5B} (and instruct)), and large decoder >5B (\texttt{Pythia-6.9B}, \texttt{OLMo-2-7B}, \texttt{Llama-3.1-8B} and instruct versions). Columns show prediction accuracy (Linear vs.\ MLP probes) and selectivity scores (linguistic minus control accuracy). Note that for readability, the y-axis for accuracy starts at 0.4. Full (non-averaged) results for individual models are provided in Appendix~\cref{sec:full_linguistic_results}.
  }
  \label{fig:linguistic_accuracy_and_selectivity}
\end{figure*}

\subsection{Results}
\label{sec:results}

We report layer-wise accuracies for lexeme and inflection prediction across all datasets models described in \cref{sec:experiments}. Detailed layer-wise accuracy and selectivity tables are provided in Appendix~\cref{sec:detailed_layerwise_tables}.

\begin{figure*}[htbp]
    \centering
    \includegraphics[width=\textwidth]{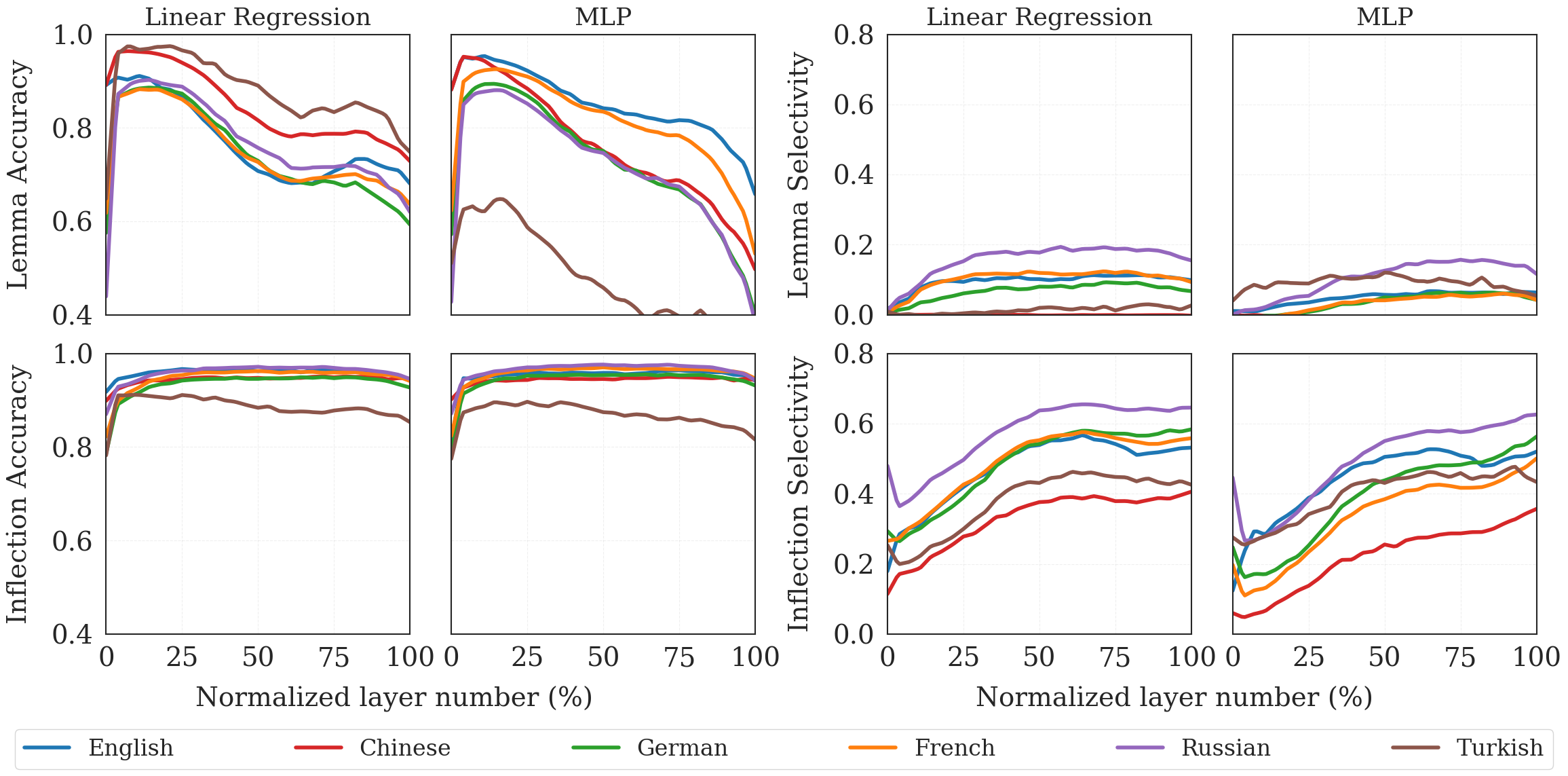}
    \caption{Cross-linguistic probing results \emph{averaged across all models within each language}.
    Columns show lexeme and inflection accuracy (Linear vs.\ MLP) followed by selectivity scores. Note that for readability, the y-axis for accuracy starts at 0.4. Full (non-averaged) results for individual models are provided in Appendix~\cref{sec:full_linguistic_results}.}
    \label{fig:multilingual_accuracy_and_selectivity}
\end{figure*}

\paragraph{Lexeme.} Lexeme accuracy under linear regression starts high (0.8--1.0) and decreases with depth in all English model families (\cref{fig:linguistic_accuracy_and_selectivity}, top left). Encoder-only models show the strongest decrease, while small decoders decline more gradually and large decoders maintain higher accuracy in deeper layers. Across languages (\cref{fig:multilingual_accuracy_and_selectivity}, top left), Turkish shows the largest drop (0.95 to 0.25), while Russian and Chinese retain 0.6--0.8 accuracy in later layers. MLP accuracy is similar but slightly higher than linear at most depths (middle column). Selectivity for lexeme remains close to zero across depths and languages (right column), indicating that high lexeme accuracy early in the network is mostly driven by surface correlations rather than strongly selective lexical structure.

\paragraph{Inflection.} Inflectional features remain readable across all layers and architectures. For English, linear regression accuracy stays near 0.9--1.0 throughout the layers (\cref{fig:linguistic_accuracy_and_selectivity}, bottom left). This pattern holds cross-linguistically (\cref{fig:multilingual_accuracy_and_selectivity}, bottom left): English, Chinese, German, French, and Russian exceed 0.9 accuracy at most depths, while Turkish is slightly lower, hovering around 0.8--0.9. MLP probes follow the same pattern (middle column). Selectivity scores for inflection remain positive (0.4--0.6) across models and languages (right column), with Russian and German at the upper end, supporting the view that inflectional features are encoded in stable, linearly accessible subspaces.

\paragraph{Probe error analysis.} Frequency strongly correlates with probe accuracy for both tasks,~\ie~frequent lexemes and inflectional categories achieve high accuracy, while rarer ones account for most of the errors.
For inflection, comparative and superlative degrees and low-frequency verb forms are the most error-prone categories. Turkish shows the strongest sensitivity to frequency, likely due to its morphological complexity creating a long tail of rare forms. A detailed breakdown by part of speech and inflectional category is given in Appendix~\cref{sec:classifier_metrics_appendix}.

\subsection{Analysis}
\label{sec:analysis}

Our results show that lexical identity is encoded strongly in early layers but becomes less accessible in later layers, whereas inflectional features remain robustly decodable throughout the model. We run a variety of experiments to further investigate this.

\textbf{Linear probes are more selective than MLP probes.}
We quantify the relationship between probe type and selectivity via the linear separability gap (defined in~\cref{eq:linear_separability_gap}); detailed plots appear in Appendix~\cref{sec:linear_separability_gap}. As shown in \cref{fig:selectivity_gap_summary}, the gap is negative for both tasks across layer depths, indicating that linear probes achieve higher selectivity than MLP probes. For lexeme, the gap is consistently negative, even though MLP probes achieve higher accuracy. This suggests that while non-linear probes can extract more lexeme information, some of that additional signal reflects memorization rather than genuine linguistic structure. For inflection, the gap is also negative on average but exhibits substantially higher variance. This suggests that in some cases both probe types extract similar amounts of selective linguistic signal (gap near zero), while in others linear probes are considerably more selective.

\begin{figure}[t]
  \centering
  \includegraphics[width=\linewidth]{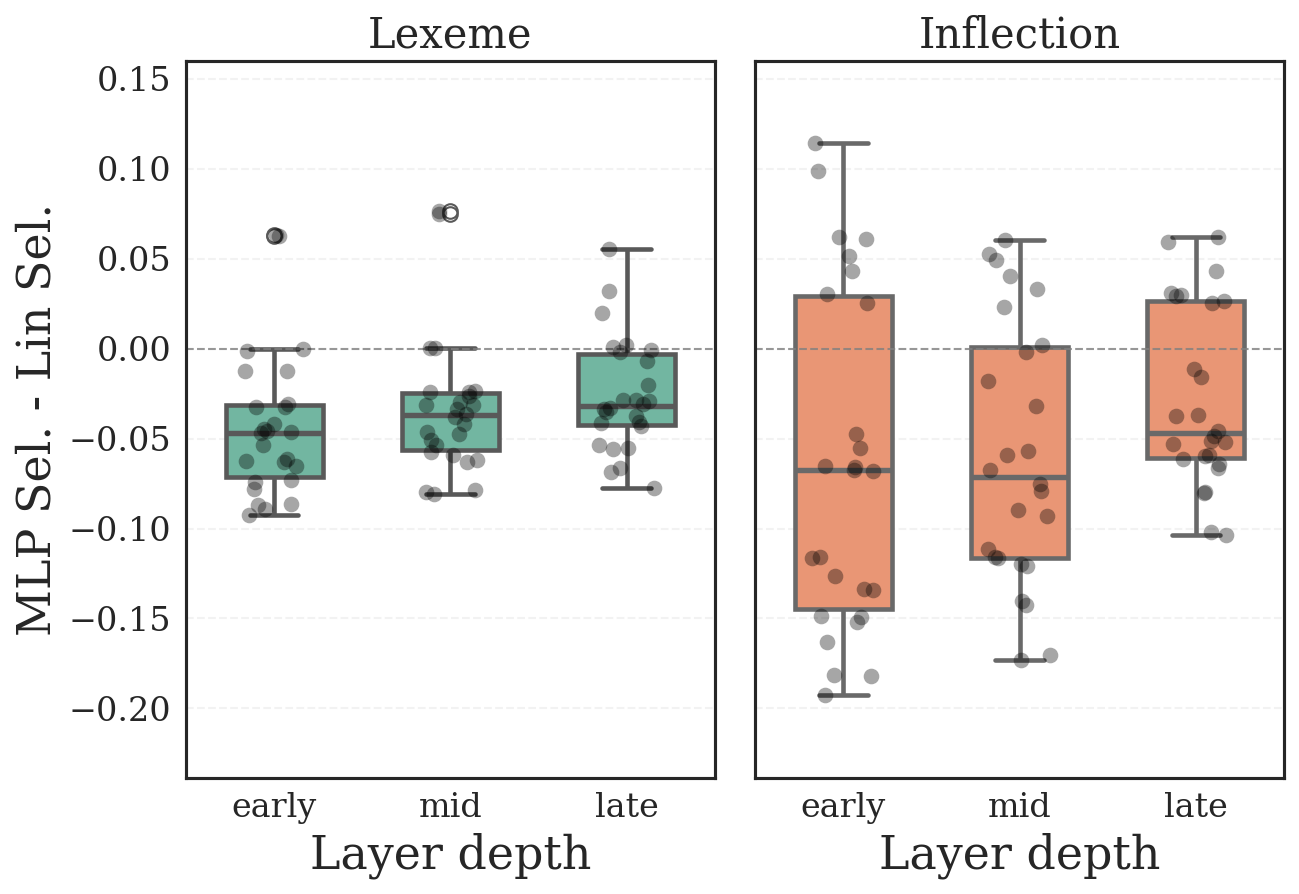}
  \caption{Linear separability gap (difference in probe selectivity) for lexeme and inflection, aggregated across models and languages and grouped by layer depth (early, mid, late). Each point represents a single model-language pair.}
  \label{fig:selectivity_gap_summary}
\end{figure}

\textbf{Some models show extreme mid-layer dimensionality compression; others maintain stable dimensionality.}
To characterize representation geometry, we measure the linear effective dimensionality across layers, building on prior work showing that neural network loss landscapes and language model fine-tuning subspaces often have low intrinsic dimensionality~\citep{li2018measuring, aghajanyan-etal-2021-intrinsic}. Following \citet{NEURIPS2019_48c8c396}, who referred to this as effective dimension and approximated it linearly via PCA, we estimate how many PCA components (as a fraction of the full basis) are needed to explain fixed variance thresholds on our dataset of collected activations. As noted by \citet{NEURIPS2019_48c8c396}, this PCA-based linear approximation serves as an upper bound for the unconstrained effective dimension.

\begin{figure*}[htbp]
  \centering
  \includegraphics[width=\textwidth]{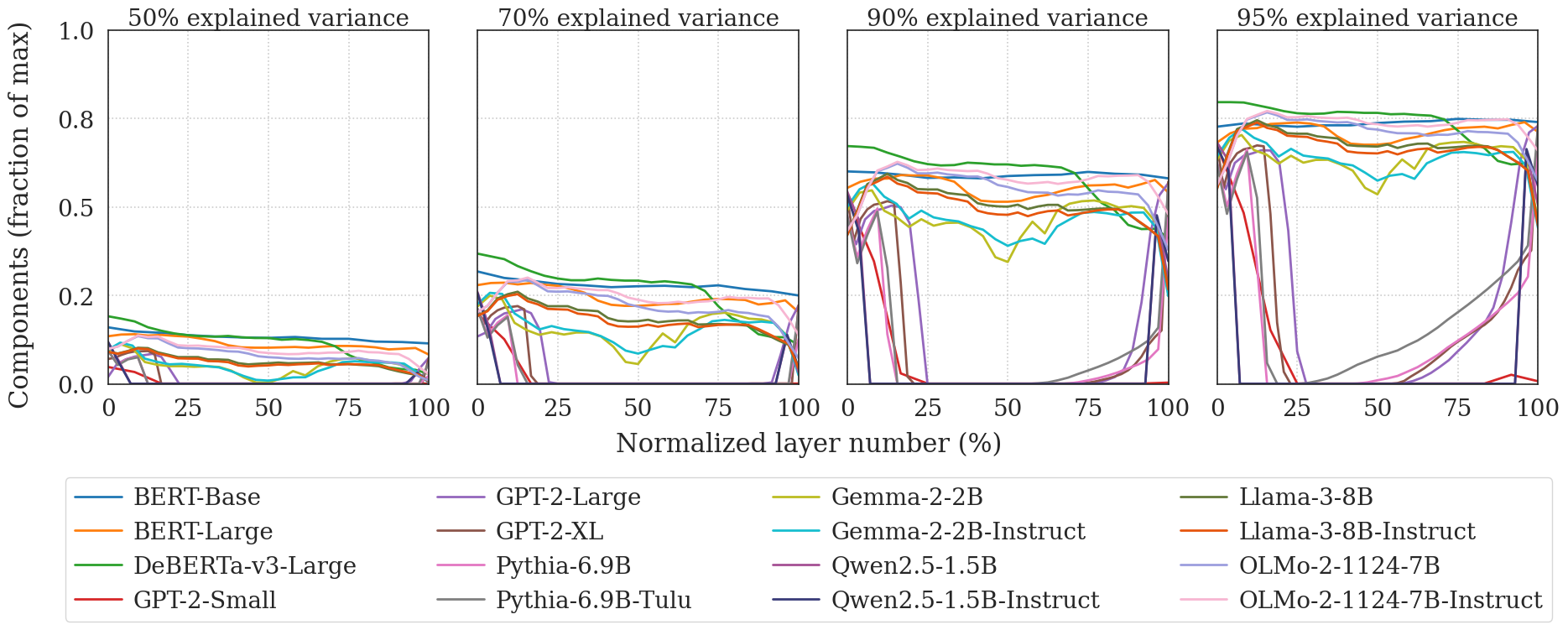}
  \caption{Linear effective dimensionality across layers. Lines show fraction of PCA components needed to reach variance thresholds (50--95\%). Full results appear in~\cref{sec:intrinsic_dim_appendix}.}
  \label{fig:intrinsic_dim_thresholds}
\end{figure*}

\Cref{fig:intrinsic_dim_thresholds} summarizes these trajectories across thresholds (50--95\%), and full results are provided in Appendix~\cref{sec:intrinsic_dim_appendix}. In models with gradual compression, the curves decline smoothly with layer index, indicating steady consolidation without a single dominant bottleneck; this behavior appears in encoder-only models (\texttt{BERT}, \texttt{DeBERTa}) and several contemporary decoders (\texttt{Gemma}, \texttt{Llama}, \texttt{OLMo-2}). However, we also observe a sharp mid-layer collapse for \texttt{GPT-2}, \texttt{Qwen2.5}, and \texttt{Pythia}: even very high variance thresholds (90\%, 95\%) become explainable by very few components before re-expanding in the very last layers. When inspecting activation statistics (Appendix~\cref{sec:massive_activations_appendix}), we find that this collapse coincides with outlier dimensions of unusually large magnitudes: \texttt{Qwen2.5-1.5B} reaches maximum absolute activations of $\sim$8000 in its middle layers, whereas \texttt{Llama-3-8B} remains around $\sim$30--40~\citep{rudman-etal-2023-outlier, sun2024massive}.

\begin{figure}[h]
  \centering
  \includegraphics[width=0.9\linewidth]{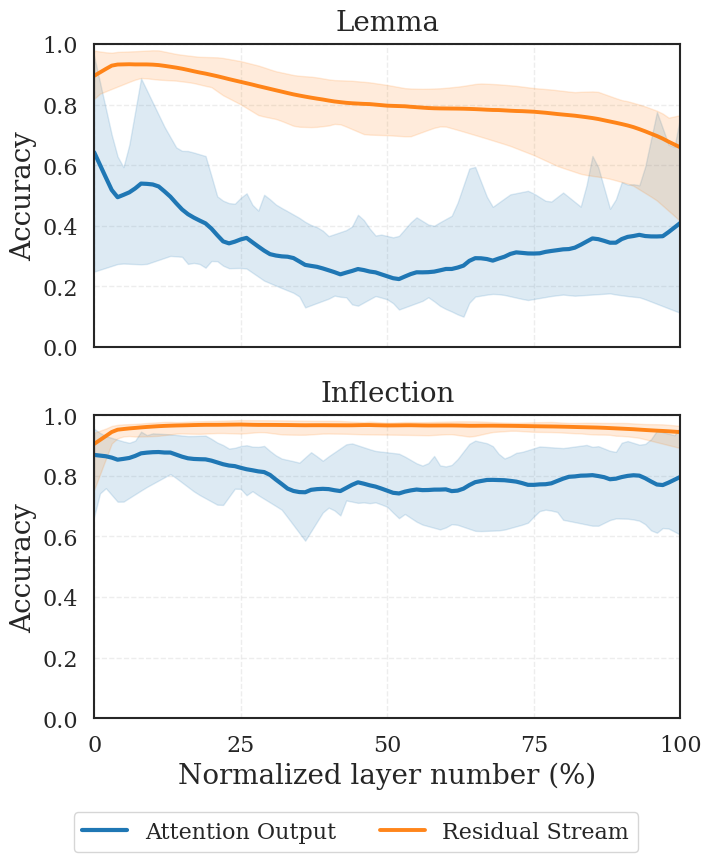}
  \caption{Probing accuracy for lexeme (top) and inflection (bottom) as a function of normalized layer depth, comparing attention-head outputs to residual-stream activations. Curves are averaged across models and probe type, with min-max shaded regions.}
  \label{fig:attention_accuracy_two_panel_averaged}
\end{figure}

\textbf{Residual streams retain more linguistic information than attention outputs.} 
We conduct a targeted probing experiment that compares \emph{attention-head outputs} to \emph{residual-stream activations} in \texttt{BERT} and contemporary decoders. \Cref{fig:attention_accuracy_two_panel_averaged} summarizes the effect averaged across models and probe types (the corresponding full, per-model curves are shown in \cref{fig:attention_bert_combined,fig:attention_other_combined} in Appendix \cref{sec:attention_analysis_appendix}). Across both lexeme and inflection, probes trained on attention outputs have lower accuracy than probes trained on the residual stream at nearly all depths. For lexeme, attention-based accuracy drops to roughly 0.2--0.4 in middle layers, while residual-stream accuracy remains closer to 0.6--0.9. For inflection, both components stay highly decodable (0.7--1.0), but residual streams still consistently outperform attention outputs, especially in middle layers. Selectivity mirrors these trends: lexeme selectivity stays near zero on attention outputs but is higher on residuals, while inflection selectivity reaches 0.4--0.5 in both streams with residuals slightly higher. Overall, these experiments support an interpretation in which attention emphasizes contextual aggregation, while the residual stream (including MLP mixing) more directly preserves token-level lexical and morphological information used by the probes, hinting that these could be better overall representations.

\begin{figure*}[!htbp]
  \centering
  \includegraphics[width=\textwidth]{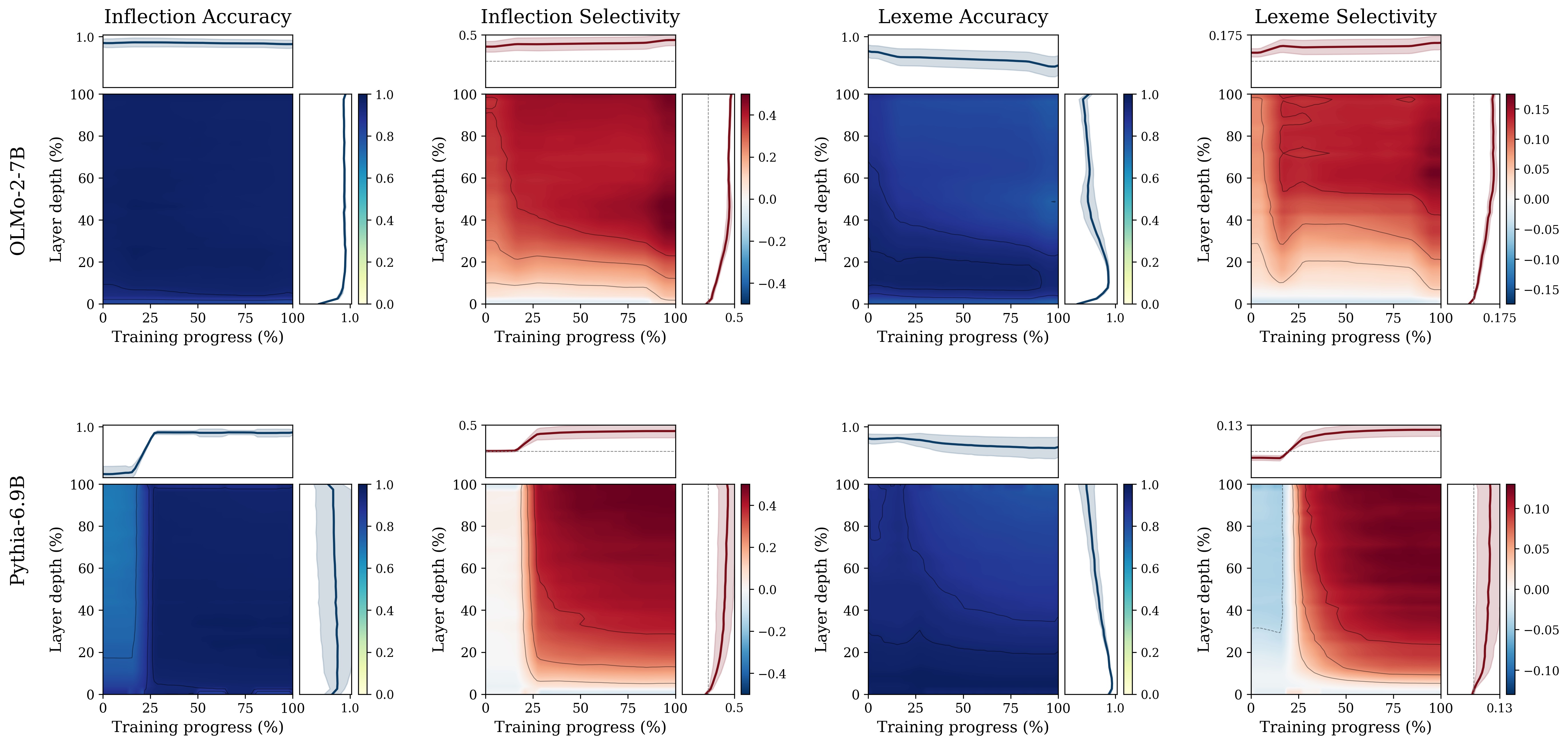}
  \caption{Checkpoint-by-layer heatmaps for \texttt{OLMo-2-7B} (top row) and \texttt{Pythia-6.9B} (bottom row). Columns show inflection accuracy, inflection selectivity, lexeme accuracy, and lexeme selectivity. The x-axis is checkpoint training progress and the y-axis is model layer depth.
  }
  \label{fig:checkpoint_layer_heatmaps}
\end{figure*}

\textbf{Inflection representations are highly steerable.}
We connect these representational measurements to causal control through inflection steering experiments (\eg singular vs.~plural). For each category pair, we compute a difference vector between mean hidden states and apply scaled interventions at each layer.~\footnote{This is identical to difference of means interventions with steering vectors~\citep{subramani-etal-2022-extracting}.}
To evaluate steering effectiveness, we apply these vectors to a set of test examples and use our trained linear classifier to assess whether the intervention successfully changes the predicted inflectional category. For example, when steering from past tense ('-ed') to gerund ('-ing'), we add the difference vector $\mathbf{v}_{\text{ing} - \text{ed}}$ to the hidden state of a word like "jumped" and measure whether the classifier now assigns higher probability to '-ing' than to '-ed'. 
\Cref{fig:steering_probability_change,fig:steering_flip_rate} show that across most architectures, even moderate intervention strengths ($\lambda=5$) produce large probability shifts and high flip rates, consistent with inflection being controllable along a small number of directions. One exception is \texttt{DeBERTa-v3-Large}, which shows a sudden drop in steering effectiveness around 75\% of model depth. The consistency of this pattern across all tested $\lambda$ values indicates that it reflects properties of the model's representational structure rather than the intervention strength $\lambda$.

\textbf{Inflection stabilizes early in training; lexeme continues to change.} 
Finally, we analyze pretraining dynamics by probing intermediate checkpoints for \texttt{OLMo-2-7B} and \texttt{Pythia-6.9B} (\cref{fig:olmo2_combined,fig:pythia_combined} in Appendix~\cref{sec:training_dynamics_appendix}). In both model families, inflection accuracy is already high at the earliest checkpoints and improves only slightly with further updates; inflection selectivity rises rapidly in the first few checkpoints and then stays near its final value. Lexical identity exhibits a different tradeoff: lemma/lexeme accuracy is highest in earlier checkpoints and tends to decline with additional training (and with depth), even as lexeme selectivity increases gradually over training, particularly in mid-to-late layers. \Cref{fig:checkpoint_layer_heatmaps} visualizes these dynamics jointly across training and depth: inflection reaches a stable regime early, while lexical identity remains more variable  and continues to be reshaped throughout training, particularly in later layers of decoder-only models. This is similar to findings from cognitive science that grammatical knowledge requires far less information storage than lexical semantics during human language acquisition \citep{10.1098/rsos.181393}.

\subsection{Discussion}
\label{sec:lemma_inflection_discussion}

We find a consistent trend across models and languages. Inflection stays highly decodable across depth and shows strongly positive selectivity, while lexeme accuracy starts high and drops in later layers and lexeme selectivity stays near zero. This suggests a simple story in which models choose to retain morphosyntactic features, since they constrain surface realization, while moving away from token identity as representations become more contextual. However, there could be other explanations: lexeme prediction has far higher cardinality than inflection, and frequency drives many of the remaining errors, so task difficulty could explain part of the gap.

We also see substantial variation in representation geometry across model families. Some models show a sharp mid-layer collapse in PCA-based effective dimensionality, and this collapse coincides with unusually large outlier activations \citep{rudman-etal-2023-outlier, sun2024massive}. Other models compress more gradually without a clear bottleneck.

Finally, we observe, based on the linear separability gap, that additional probe capacity tends to capture spurious correlations rather than linguistic structure. This aligns with~\citet{hewitt-liang-2019-designing}, who report a similar trend on \texttt{ELMo}, highlighting the need to report selectivity alongside accuracy when interpreting probing results.
\section{Related Work}
\label{sec:related_work}

\textbf{Probing for linguistic information.} 
Probing studies typically use supervised classifiers to predict linguistic properties from model representations \citep{alain2017understanding, adi2017finegrained}. Extensive work has established that early transformer models (\texttt{BERT}, \texttt{GPT-2}) learn hierarchical linguistic structures, with different layers specializing in different information types: lower layers capture surface features and morphology, middle layers encode syntax, and upper layers represent semantics and context \citep{jawahar-etal-2019-bert, tenney-etal-2019-bert, rogers-etal-2020-primer}. More relevant to our work, \citet{vulic-etal-2020-probing} found that lexical information concentrates in lower layers, while \citet{ethayarajh-2019-contextual} showed that representations become increasingly context-specific in higher layers.

For morphology specifically, \citet{Acs_Hamerlik_Schwartz_Smith_Kornai_2024} introduced an extensive multilingual probing dataset (247 tasks across 42 languages), finding that \texttt{mBERT} and \texttt{XLM-RoBERTa} encode morphosyntactic features strongly, with preceding context more informative than following context for morphological prediction. \citet{lasri-etal-2022-probing} distinguished between spurious and \emph{functional} encodings---those actually used by the model---demonstrating that BERT relies on linear representations of grammatical number for agreement tasks, with nouns and verbs encoded in disjoint subspaces.

\textbf{Representation dynamics in modern LLMs.} 
Recent research has extended these analyses to modern, larger-scale generative models. \citet{cheng2025emergence} identify a distinct high-dimensional abstraction phase in the early-to-middle layers of models like Llama and OLMo, suggesting that the transition from surface-level to abstract linguistic features occurs earlier than in previous architectures. Similarly, \citet{skean2025layer} demonstrate that intermediate layers in modern LLMs often encode richer task-transferable representations than final layers. These findings align with the layerwise dynamics we observe in Section~\cref{sec:experiments}.

\textbf{Activation steering and mechanistic interpretability.} 
Beyond probing, recent work has explored manipulating model behavior by intervening on internal representations, including steering vectors~\citep{subramani-etal-2022-extracting}, inference-time interventions~\citep{li2023inferencetime}, and representation editing~\citep{meng2022locating}. Mechanistic interpretability approaches aim to reverse-engineer learned algorithms \citep{elhage2021mathematical}, with recent work using sparse autoencoders to decompose representations into interpretable latent features \citep{cunningham2023sparseautoencodershighlyinterpretable, bricken2023monosemanticity}, providing clearer targets for interpretation than raw activations. See Appendix~\cref{sec:additional_related_work} for detailed discussion.

\section{Conclusion}
In this work, we analyzed 25 transformer models to understand how they encode two token-level linguistic properties: lexical identity and inflectional features. We find that these properties follow distinct representational trajectories: lexical identity is most linearly accessible in early layers but becomes increasingly entangled deeper in the network, while inflectional information remains robustly and linearly decodable across layers and languages. Additional analyses of residual streams, attention outputs, activation steering, and pretraining dynamics further reveal that inflection occupies compact, steerable subspaces that stabilize early in training, and that linear probes are more selective than MLP probes for capturing these properties. Collectively, these findings suggest that despite rapid advances in model scale and training, transformers converge on robust internal representations of core morphological properties.
\section{Limitations}
\label{sec:limitations}

\paragraph{Representation Extraction for Decoder Models} 
Our current approach for extracting word representations from decoder-only models uses the final subword token.
This assumption is an intuitive and natural choice, but may not be optimal for all architectures and models. 
Future work could develop better extraction methods that account for subword tokenization effects and leverage attention patterns to create more accurate word-level representations.

\paragraph{Form and Function in Inflection} 
Some languages contain cases where different grammatical functions share the same surface form (\eg infinitive and non-past verb forms in English). We do not explicitly examine these cases in our classification experiments, but these ambiguities create opportunities to better examine how models separate form from function across languages.

\paragraph{Indirect Nature of Classifiers}
While our classifier methodology follows established best practices~\citep{hewitt-liang-2019-designing, liu-etal-2019-linguistic}, we only detect correlations in hidden activations, not causal mechanisms.

\paragraph{Scope of Steering Experiments}
Our steering vector experiments measure changes in classifier performance rather than downstream model outputs. Evaluating effects on actual model generation would require more complex experimental designs to control for confounding factors and ensure that observed changes result from the intended representational modifications rather than other influences.
\section{Acknowledgements}

We thank David Mortensen and Mona Diab for their valuable guidance and feedback throughout this work. We also thank the anonymous reviewers for their thoughtful comments and suggestions, including the suggestion to extend our analysis to a multilingual setting, which greatly improved the paper.

\bibliography{custom}

@inproceedings{adi2017finegrained,
    title={Fine-grained Analysis of Sentence Embeddings Using Auxiliary Prediction Tasks},
    author={Yossi Adi and Einat Kermany and Yonatan Belinkov and Ofer Lavi and Yoav Goldberg},
    booktitle={5th International Conference on Learning Representations (Conference Track)},
    year={2017},
    url={https://openreview.net/forum?id=BJh6Ztuxl}
}

@inproceedings{
    alain2017understanding,
    title={Understanding intermediate layers using linear classifier probes},
    author={Guillaume Alain and Yoshua Bengio},
    booktitle={5th International Conference on Learning Representations (Workshop Track)},
    year={2017},
    url={https://openreview.net/forum?id=ryF7rTqgl}
}

@Article{Zeldes2017,
  author    = {Amir Zeldes},
  title     = {The {GUM} Corpus: Creating Multilayer Resources in the Classroom},
  journal   = {Language Resources and Evaluation},
  year      = {2017},
  volume    = {51},
  number    = {3},
  pages     = {581--612},
  doi       = {http://dx.doi.org/10.1007/s10579-016-9343-x}
}

@article{radford2019language,
  title={Language models are unsupervised multitask learners},
  author = {Radford, Alec and Wu, Jeff and Child, Rewon and Luan, David and Amodei, Dario and Sutskever, Ilya},
  journal={OpenAI blog},
  volume={1},
  number={8},
  pages={9},
  year={2019}
}

@misc{qwen2.5,
    title = {Qwen2.5: A Party of Foundation Models},
    url = {https://qwenlm.github.io/blog/qwen2.5/},
    author = {Qwen Team},
    month = {September},
    year = {2024}
}

@article{hui2024qwen2,
      title={Qwen2. 5-Coder Technical Report},
      author={Hui, Binyuan and Yang, Jian and Cui, Zeyu and Yang, Jiaxi and Liu, Dayiheng and Zhang, Lei and Liu, Tianyu and Zhang, Jiajun and Yu, Bowen and Dang, Kai and others},
      journal={arXiv preprint arXiv:2409.12186},
      year={2024}
}

@misc{chang2024goldfishmonolinguallanguagemodels,
      title={Goldfish: Monolingual Language Models for 350 Languages}, 
      author={Tyler A. Chang and Catherine Arnett and Zhuowen Tu and Benjamin K. Bergen},
      year={2024},
      eprint={2408.10441},
      archivePrefix={arXiv},
      primaryClass={cs.CL},
      url={https://arxiv.org/abs/2408.10441}, 
}

@misc{panickssery2024steeringllama2contrastive,
      title={Steering Llama 2 via Contrastive Activation Addition}, 
      author={Nina Panickssery and Nick Gabrieli and Julian Schulz and Meg Tong and Evan Hubinger and Alexander Matt Turner},
      year={2024},
      eprint={2312.06681},
      archivePrefix={arXiv},
      primaryClass={cs.CL},
      url={https://arxiv.org/abs/2312.06681},
}

@inproceedings{
    ilharco2023editing,
    title={Editing models with task arithmetic},
    author={Gabriel Ilharco and Marco Tulio Ribeiro and Mitchell Wortsman and Ludwig Schmidt and Hannaneh Hajishirzi and Ali Farhadi},
    booktitle={The Eleventh International Conference on Learning Representations },
    year={2023},
    url={https://openreview.net/forum?id=6t0Kwf8-jrj}
}

@misc{cunningham2023sparseautoencodershighlyinterpretable,
      title={Sparse Autoencoders Find Highly Interpretable Features in Language Models}, 
      author={Hoagy Cunningham and Aidan Ewart and Logan Riggs and Robert Huben and Lee Sharkey},
      year={2023},
      eprint={2309.08600},
      archivePrefix={arXiv},
      primaryClass={cs.LG},
      url={https://arxiv.org/abs/2309.08600}, 
}

@article{bricken2023monosemanticity,
   title={Towards Monosemanticity: Decomposing Language Models With Dictionary Learning},
   author={Bricken, Trenton and Templeton, Adly and Batson, Joshua and Chen, Brian and Jermyn, Adam and Conerly, Tom and Turner, Nick and Anil, Cem and Denison, Carson and Askell, Amanda and Lasenby, Robert and Wu, Yifan and Kravec, Shauna and Schiefer, Nicholas and Maxwell, Tim and Joseph, Nicholas and Hatfield-Dodds, Zac and Tamkin, Alex and Nguyen, Karina and McLean, Brayden and Burke, Josiah E and Hume, Tristan and Carter, Shan and Henighan, Tom and Olah, Christopher},
   year={2023},
   journal={Transformer Circuits Thread},
   note={https://transformer-circuits.pub/2023/monosemantic-features/index.html}
}

@inproceedings{
    meng2022locating,
    title={Locating and Editing Factual Associations in {GPT}},
    author={Kevin Meng and David Bau and Alex J Andonian and Yonatan Belinkov},
    booktitle={Advances in Neural Information Processing Systems},
    editor={Alice H. Oh and Alekh Agarwal and Danielle Belgrave and Kyunghyun Cho},
    year={2022},
    url={https://openreview.net/forum?id=-h6WAS6eE4}
}

@inproceedings{vig2020investigating,
    author = {Vig, Jesse and Gehrmann, Sebastian and Belinkov, Yonatan and Qian, Sharon and Nevo, Daniel and Singer, Yaron and Shieber, Stuart},
    booktitle = {Advances in Neural Information Processing Systems},
    editor = {H. Larochelle and M. Ranzato and R. Hadsell and M.F. Balcan and H. Lin},
    pages = {12388--12401},
    publisher = {Curran Associates, Inc.},
    title = {Investigating Gender Bias in Language Models Using Causal Mediation Analysis},
    url = {https://proceedings.neurips.cc/paper_files/paper/2020/file/92650b2e92217715fe312e6fa7b90d82-Paper.pdf},
    volume = {33},
    year = {2020}
    }

@inproceedings{
    geiger2021causal,
    title={Causal Abstractions of Neural Networks},
    author={Atticus Geiger and Hanson Lu and Thomas F Icard and Christopher Potts},
    booktitle={Advances in Neural Information Processing Systems},
    editor={A. Beygelzimer and Y. Dauphin and P. Liang and J. Wortman Vaughan},
    year={2021},
    url={https://openreview.net/forum?id=RmuXDtjDhG}
}

@inproceedings{
    li2023inferencetime,
    title={Inference-Time Intervention: Eliciting Truthful Answers from a Language Model},
    author={Kenneth Li and Oam Patel and Fernanda Vi{\'e}gas and Hanspeter Pfister and Martin Wattenberg},
    booktitle={Thirty-seventh Conference on Neural Information Processing Systems},
    year={2023},
    url={https://openreview.net/forum?id=aLLuYpn83y}
}

@misc{grattafiori2024llama3herdmodels,
      title={The Llama 3 Herd of Models}, 
      author={Team Llama},
      year={2024},
      eprint={2407.21783},
      archivePrefix={arXiv},
      primaryClass={cs.AI},
      url={https://arxiv.org/abs/2407.21783}, 
}

@misc{workshop2023bloom176bparameteropenaccessmultilingual,
      title={BLOOM: A 176B-Parameter Open-Access Multilingual Language Model}, 
      author={BigScience Workshop},
      year={2023},
      eprint={2211.05100},
      archivePrefix={arXiv},
      primaryClass={cs.CL},
      url={https://arxiv.org/abs/2211.05100}, 
}

@article{elhage2021mathematical,
   title={A Mathematical Framework for Transformer Circuits},
   author={Elhage, Nelson and Nanda, Neel and Olsson, Catherine and Henighan, Tom and Joseph, Nicholas and Mann, Ben and Askell, Amanda and Bai, Yuntao and Chen, Anna and Conerly, Tom and DasSarma, Nova and Drain, Dawn and Ganguli, Deep and Hatfield-Dodds, Zac and Hernandez, Danny and Jones, Andy and Kernion, Jackson and Lovitt, Liane and Ndousse, Kamal and Amodei, Dario and Brown, Tom and Clark, Jack and Kaplan, Jared and McCandlish, Sam and Olah, Chris},
   year={2021},
   journal={Transformer Circuits Thread},
   note={https://transformer-circuits.pub/2021/framework/index.html}
}

@misc{brown2020languagemodelsfewshotlearners,
      title={Language Models are Few-Shot Learners}, 
      author={Tom B. Brown and Benjamin Mann and Nick Ryder and Melanie Subbiah and Jared Kaplan and Prafulla Dhariwal and Arvind Neelakantan and Pranav Shyam and Girish Sastry and Amanda Askell and Sandhini Agarwal and Ariel Herbert-Voss and Gretchen Krueger and Tom Henighan and Rewon Child and Aditya Ramesh and Daniel M. Ziegler and Jeffrey Wu and Clemens Winter and Christopher Hesse and Mark Chen and Eric Sigler and Mateusz Litwin and Scott Gray and Benjamin Chess and Jack Clark and Christopher Berner and Sam McCandlish and Alec Radford and Ilya Sutskever and Dario Amodei},
      year={2020},
      eprint={2005.14165},
      archivePrefix={arXiv},
      primaryClass={cs.CL},
      url={https://arxiv.org/abs/2005.14165}, 
}

@misc{lambert2025tulu3pushingfrontiers,
      title={Tulu 3: Pushing Frontiers in Open Language Model Post-Training}, 
      author={Nathan Lambert and Jacob Morrison and Valentina Pyatkin and Shengyi Huang and Hamish Ivison and Faeze Brahman and Lester James V. Miranda and Alisa Liu and Nouha Dziri and Shane Lyu and Yuling Gu and Saumya Malik and Victoria Graf and Jena D. Hwang and Jiangjiang Yang and Ronan Le Bras and Oyvind Tafjord and Chris Wilhelm and Luca Soldaini and Noah A. Smith and Yizhong Wang and Pradeep Dasigi and Hannaneh Hajishirzi},
      year={2025},
      eprint={2411.15124},
      archivePrefix={arXiv},
      primaryClass={cs.CL},
      url={https://arxiv.org/abs/2411.15124}, 
}

@inproceedings{droganova2018data,
  title={Data conversion and consistency of monolingual corpora: Russian UD treebanks},
  author={Droganova, Kira and Lyashevskaya, Olga and Zeman, Daniel},
  booktitle={Proceedings of the 17th international workshop on treebanks and linguistic theories (tlt 2018)},
  volume={155},
  pages={53--66},
  year={2018},
  organization={Link{\"o}ping University Electronic Press Link{\"o}ping, Sweden}
}

@inproceedings{
cheng2025emergence,
title={Emergence of a High-Dimensional Abstraction Phase in Language Transformers},
author={Emily Cheng and Diego Doimo and Corentin Kervadec and Iuri Macocco and Lei Yu and Alessandro Laio and Marco Baroni},
booktitle={The Thirteenth International Conference on Learning Representations},
year={2025},
url={https://openreview.net/forum?id=0fD3iIBhlV}
}

@inproceedings{
skean2025layer,
title={Layer by Layer: Uncovering Hidden Representations in Language Models},
author={Oscar Skean and Md Rifat Arefin and Dan Zhao and Niket Nikul Patel and Jalal Naghiyev and Yann LeCun and Ravid Shwartz-Ziv},
booktitle={Forty-second International Conference on Machine Learning},
year={2025},
url={https://openreview.net/forum?id=WGXb7UdvTX}
}

@misc{nostalgebraist2020logitlens,
  author = {nostalgebraist},
  title = {Interpreting {GPT}: The Logit Lens},
  year = {2020},
  howpublished = {\url{https://www.lesswrong.com/posts/AcKRB8wDpdaN6v6ru/interpreting-gpt-the-logit-lens}},
  note = {LessWrong blog post}
}

@inproceedings{
sun2024massive,
title={Massive Activations in Large Language Models},
author={Mingjie Sun and Xinlei Chen and J Zico Kolter and Zhuang Liu},
booktitle={First Conference on Language Modeling},
year={2024},
url={https://openreview.net/forum?id=F7aAhfitX6}
}

@article{
Acs_Hamerlik_Schwartz_Smith_Kornai_2024,
title={Morphosyntactic probing of multilingual BERT models},
volume={30},
DOI={10.1017/S1351324923000190},
number={4},
journal={Natural Language Engineering},
author={Acs, Judit and Hamerlik, Endre and Schwartz, Roy and Smith, Noah A. and Kornai, Andras},
year={2024},
pages={753–792}}

@inproceedings{NEURIPS2019_48c8c396,
 author = {Subramani, Nishant and Bowman, Samuel and Cho, Kyunghyun},
 booktitle = {Advances in Neural Information Processing Systems},
 editor = {H. Wallach and H. Larochelle and A. Beygelzimer and F. d\textquotesingle Alch\'{e}-Buc and E. Fox and R. Garnett},
 pages = {},
 publisher = {Curran Associates, Inc.},
 title = {Can Unconditional Language Models Recover Arbitrary Sentences?},
 url = {https://proceedings.neurips.cc/paper_files/paper/2019/file/48c8c3963853fff20bd9e8bee9bd4c07-Paper.pdf},
 volume = {32},
 year = {2019}
}

@inproceedings{
feucht2025the,
title={The Dual-Route Model of Induction},
author={Sheridan Feucht and Eric Todd and Byron C Wallace and David Bau},
booktitle={Second Conference on Language Modeling},
year={2025},
url={https://openreview.net/forum?id=bNTrKqqnG9}
}

@inproceedings{
kaplan2025from,
title={From Tokens to Words: On the Inner Lexicon of {LLM}s},
author={Guy Kaplan and Matanel Oren and Yuval Reif and Roy Schwartz},
booktitle={The Thirteenth International Conference on Learning Representations},
year={2025},
url={https://openreview.net/forum?id=328vch6tRs}
}

@article{10.1098/rsos.181393,
    author = {Mollica, Francis and Piantadosi, Steven T.},
    title = {Humans store about 1.5 megabytes of information during language acquisition},
    journal = {Royal Society Open Science},
    volume = {6},
    number = {3},
    pages = {181393},
    year = {2019},
    month = {03},
    issn = {2054-5703},
    doi = {10.1098/rsos.181393},
    url = {https://doi.org/10.1098/rsos.181393},
    eprint = {https://royalsocietypublishing.org/rsos/article-pdf/doi/10.1098/rsos.181393/976518/rsos.181393.pdf},
}

@inproceedings{
    li2018measuring,
    title={Measuring the Intrinsic Dimension of Objective Landscapes},
    author={Chunyuan Li and Heerad Farkhoor and Rosanne Liu and Jason Yosinski},
    booktitle={International Conference on Learning Representations},
    year={2018},
    url={https://openreview.net/forum?id=ryup8-WCW},
}

@article{DBLP:journals/corr/abs-2008-09049,
  author       = {Nishant Subramani and
                  Nivedita Suresh},
  title        = {Discovering Useful Sentence Representations from Large Pretrained
                  Language Models},
  journal      = {CoRR},
  volume       = {abs/2008.09049},
  year         = {2020},
  url          = {https://arxiv.org/abs/2008.09049},
  eprinttype   = {arXiv},
  eprint       = {2008.09049},
  bibsource    = {dblp computer science bibliography, https://dblp.org}
}

@article{liu2025llm,
  title={LLM Microscope: What Model Internals Reveal About Answer Correctness and Context Utilization},
  author={Liu, Jiarui and Jain, Jivitesh and Diab, Mona and Subramani, Nishant},
  journal={arXiv preprint arXiv:2510.04013},
  year={2025}
}

@string{acl = {Association for Computational Linguistics}}

@string{anth = {https://aclanthology.org/}}

@inproceedings{aghajanyan-etal-2021-intrinsic,title = "Intrinsic Dimensionality Explains the Effectiveness of Language Model Fine-Tuning",author = "Aghajanyan, Armen and Gupta, Sonal and Zettlemoyer, Luke",editor = "Zong, Chengqing and Xia, Fei and Li, Wenjie and Navigli, Roberto",booktitle = "Proceedings of the 59th Annual Meeting of the Association for Computational Linguistics and the 11th International Joint Conference on Natural Language Processing (Volume 1: Long Papers)",month = aug,year = "2021",address = "Online",publisher = acl,url = anth # {2021.acl-long.568/},doi = "10.18653/v1/2021.acl-long.568",pages = "7319--7328"}

@inproceedings{dang-etal-2025-tokenization,title = "Tokenization and Morphology in Multilingual Language Models: A Comparative Analysis of m{T}5 and {B}y{T}5",author = "Dang, Thao Anh and Raviv, Limor and Galke, Lukas",editor = "Abbas, Mourad and Yousef, Tariq and Galke, Lukas",booktitle = "Proceedings of the 8th International Conference on Natural Language and Speech Processing (ICNLSP-2025)",month = aug,year = "2025",address = "Southern Denmark University, Odense, Denmark",publisher = acl,url = anth # {2025.icnlsp-1.24/},pages = "242--257"}

@inproceedings{devlin-etal-2019-bert,title = "{BERT}: Pre-training of Deep Bidirectional Transformers for Language Understanding",author = "Devlin, Jacob and Chang, Ming-Wei and Lee, Kenton and Toutanova, Kristina",editor = "Burstein, Jill and Doran, Christy and Solorio, Thamar",booktitle = "Proceedings of the 2019 Conference of the North {A}merican Chapter of the Association for Computational Linguistics: Human Language Technologies, Volume 1 (Long and Short Papers)",month = jun,year = "2019",address = "Minneapolis, Minnesota",publisher = acl,url = anth # {N19-1423/},doi = "10.18653/v1/N19-1423",pages = "4171--4186"}

@article{elazar-etal-2021-amnesic,title = "Amnesic Probing: Behavioral Explanation with Amnesic Counterfactuals",author = "Elazar, Yanai and Ravfogel, Shauli and Jacovi, Alon and Goldberg, Yoav",editor = "Roark, Brian and Nenkova, Ani",journal = "Transactions of the Association for Computational Linguistics",volume = "9",year = "2021",address = "Cambridge, MA",publisher = "MIT Press",url = anth # {2021.tacl-1.10/},doi = "10.1162/tacl_a_00359",pages = "160--175"}

@inproceedings{ethayarajh-2019-contextual,title = "How Contextual are Contextualized Word Representations? {C}omparing the Geometry of {BERT}, {ELM}o, and {GPT}-2 Embeddings",author = "Ethayarajh, Kawin",editor = "Inui, Kentaro and Jiang, Jing and Ng, Vincent and Wan, Xiaojun",booktitle = "Proceedings of the 2019 Conference on Empirical Methods in Natural Language Processing and the 9th International Joint Conference on Natural Language Processing (EMNLP-IJCNLP)",month = nov,year = "2019",address = "Hong Kong, China",publisher = acl,url = anth # {D19-1006/},doi = "10.18653/v1/D19-1006",pages = "55--65"}

@inproceedings{groeneveld-etal-2024-olmo,title = "{OLM}o: Accelerating the Science of Language Models",author = "Groeneveld, Dirk and Beltagy, Iz and Walsh, Evan and Bhagia, Akshita and Kinney, Rodney and Tafjord, Oyvind and Jha, Ananya and Ivison, Hamish and Magnusson, Ian and Wang, Yizhong and Arora, Shane and Atkinson, David and Authur, Russell and Chandu, Khyathi and Cohan, Arman and Dumas, Jennifer and Elazar, Yanai and Gu, Yuling and Hessel, Jack and Khot, Tushar and Merrill, William and Morrison, Jacob and Muennighoff, Niklas and Naik, Aakanksha and Nam, Crystal and Peters, Matthew and Pyatkin, Valentina and Ravichander, Abhilasha and Schwenk, Dustin and Shah, Saurabh and Smith, William and Strubell, Emma and Subramani, Nishant and Wortsman, Mitchell and Dasigi, Pradeep and Lambert, Nathan and Richardson, Kyle and Zettlemoyer, Luke and Dodge, Jesse and Lo, Kyle and Soldaini, Luca and Smith, Noah and Hajishirzi, Hannaneh",editor = "Ku, Lun-Wei and Martins, Andre and Srikumar, Vivek",booktitle = "Proceedings of the 62nd Annual Meeting of the Association for Computational Linguistics (Volume 1: Long Papers)",month = aug,year = "2024",address = "Bangkok, Thailand",publisher = acl,url = anth # {2024.acl-long.841/},doi = "10.18653/v1/2024.acl-long.841",pages = "15789--15809"}

@article{guillaume-etal-2019-conversion,title = "Conversion et am{\'e}liorations de corpus du fran{\c{c}}ais annot{\'e}s en {U}niversal {D}ependencies [Conversion and Improvement of {U}niversal {D}ependencies {F}rench corpora]",author = "Guillaume, Bruno and de Marneffe, Marie-Catherine and Perrier, Guy",editor = "Candito, Marie and Liberman, Mark",journal = "Traitement Automatique des Langues",volume = "60",number = "2",year = "2019",address = "France",publisher = "ATALA (Association pour le Traitement Automatique des Langues)",url = anth # {2019.tal-2.4/},pages = "71--95",language = "fra"}

@inproceedings{he-etal-2024-decoding,title = "Decoding Probing: Revealing Internal Linguistic Structures in Neural Language Models Using Minimal Pairs",author = "He, Linyang and Chen, Peili and Nie, Ercong and Li, Yuanning and Brennan, Jonathan R.",editor = "Calzolari, Nicoletta and Kan, Min-Yen and Hoste, Veronique and Lenci, Alessandro and Sakti, Sakriani and Xue, Nianwen",booktitle = "Proceedings of the 2024 Joint International Conference on Computational Linguistics, Language Resources and Evaluation (LREC-COLING 2024)",month = may,year = "2024",address = "Torino, Italia",publisher = "ELRA and ICCL",url = anth # {2024.lrec-main.402/},pages = "4488--4497"}

@inproceedings{hewitt-liang-2019-designing,title = "Designing and Interpreting Probes with Control Tasks",author = "Hewitt, John and Liang, Percy",editor = "Inui, Kentaro and Jiang, Jing and Ng, Vincent and Wan, Xiaojun",booktitle = "Proceedings of the 2019 Conference on Empirical Methods in Natural Language Processing and the 9th International Joint Conference on Natural Language Processing (EMNLP-IJCNLP)",month = nov,year = "2019",address = "Hong Kong, China",publisher = acl,url = anth # {D19-1275/},doi = "10.18653/v1/D19-1275",pages = "2733--2743"}

@inproceedings{jawahar-etal-2019-bert,title = "What Does {BERT} Learn about the Structure of Language?",author = "Jawahar, Ganesh and Sagot, Beno{\^i}t and Seddah, Djam{\'e}",editor = "Korhonen, Anna and Traum, David and M{\`a}rquez, Llu{\'i}s",booktitle = "Proceedings of the 57th Annual Meeting of the Association for Computational Linguistics",month = jul,year = "2019",address = "Florence, Italy",publisher = acl,url = anth # {P19-1356/},doi = "10.18653/v1/P19-1356",pages = "3651--3657"}

@inproceedings{lasri-etal-2022-probing,title = "Probing for the Usage of Grammatical Number",author = "Lasri, Karim and Pimentel, Tiago and Lenci, Alessandro and Poibeau, Thierry and Cotterell, Ryan",editor = "Muresan, Smaranda and Nakov, Preslav and Villavicencio, Aline",booktitle = "Proceedings of the 60th Annual Meeting of the Association for Computational Linguistics (Volume 1: Long Papers)",month = may,year = "2022",address = "Dublin, Ireland",publisher = acl,url = anth # {2022.acl-long.603/},doi = "10.18653/v1/2022.acl-long.603",pages = "8818--8831"}

@inproceedings{liu-etal-2019-linguistic,title = "Linguistic Knowledge and Transferability of Contextual Representations",author = "Liu, Nelson F. and Gardner, Matt and Belinkov, Yonatan and Peters, Matthew E. and Smith, Noah A.",editor = "Burstein, Jill and Doran, Christy and Solorio, Thamar",booktitle = "Proceedings of the 2019 Conference of the North {A}merican Chapter of the Association for Computational Linguistics: Human Language Technologies, Volume 1 (Long and Short Papers)",month = jun,year = "2019",address = "Minneapolis, Minnesota",publisher = acl,url = anth # {N19-1112/},doi = "10.18653/v1/N19-1112",pages = "1073--1094"}

@inproceedings{mcdonald-etal-2013-universal,title = "{U}niversal {D}ependency Annotation for Multilingual Parsing",author = {McDonald, Ryan and Nivre, Joakim and Quirmbach-Brundage, Yvonne and Goldberg, Yoav and Das, Dipanjan and Ganchev, Kuzman and Hall, Keith and Petrov, Slav and Zhang, Hao and T{\"a}ckstr{\"o}m, Oscar and Bedini, Claudia and Bertomeu Castell{\'o}, N{\'u}ria and Lee, Jungmee},editor = "Schuetze, Hinrich and Fung, Pascale and Poesio, Massimo",booktitle = "Proceedings of the 51st Annual Meeting of the Association for Computational Linguistics (Volume 2: Short Papers)",month = aug,year = "2013",address = "Sofia, Bulgaria",publisher = acl,url = anth # {P13-2017/},pages = "92--97"}

@inproceedings{niu-etal-2022-bert,title = "Does {BERT} Rediscover a Classical {NLP} Pipeline?",author = "Niu, Jingcheng and Lu, Wenjie and Penn, Gerald",editor = "Calzolari, Nicoletta and Huang, Chu-Ren and Kim, Hansaem and Pustejovsky, James and Wanner, Leo and Choi, Key-Sun and Ryu, Pum-Mo and Chen, Hsin-Hsi and Donatelli, Lucia and Ji, Heng and Kurohashi, Sadao and Paggio, Patrizia and Xue, Nianwen and Kim, Seokhwan and Hahm, Younggyun and He, Zhong and Lee, Tony Kyungil and Santus, Enrico and Bond, Francis and Na, Seung-Hoon",booktitle = "Proceedings of the 29th International Conference on Computational Linguistics",month = oct,year = "2022",address = "Gyeongju, Republic of Korea",publisher = "International Committee on Computational Linguistics",url = anth # {2022.coling-1.278/},pages = "3143--3153"}

@inproceedings{nivre-etal-2016-universal,title = "{U}niversal {D}ependencies v1: A Multilingual Treebank Collection",author = "Nivre, Joakim and de Marneffe, Marie-Catherine and Ginter, Filip and Goldberg, Yoav and Haji{\v{c}}, Jan and Manning, Christopher D. and McDonald, Ryan and Petrov, Slav and Pyysalo, Sampo and Silveira, Natalia and Tsarfaty, Reut and Zeman, Daniel",editor = "Calzolari, Nicoletta and Choukri, Khalid and Declerck, Thierry and Goggi, Sara and Grobelnik, Marko and Maegaard, Bente and Mariani, Joseph and Mazo, Helene and Moreno, Asuncion and Odijk, Jan and Piperidis, Stelios",booktitle = "Proceedings of the Tenth International Conference on Language Resources and Evaluation ({LREC}'16)",month = may,year = "2016",address = "Portoro{\v{z}}, Slovenia",publisher = "European Language Resources Association (ELRA)",url = anth # {L16-1262/},pages = "1659--1666"}

@inproceedings{pimentel-etal-2020-information,title = "Information-Theoretic Probing for Linguistic Structure",author = "Pimentel, Tiago and Valvoda, Josef and Maudslay, Rowan Hall and Zmigrod, Ran and Williams, Adina and Cotterell, Ryan",editor = "Jurafsky, Dan and Chai, Joyce and Schluter, Natalie and Tetreault, Joel",booktitle = "Proceedings of the 58th Annual Meeting of the Association for Computational Linguistics",month = jul,year = "2020",address = "Online",publisher = acl,url = anth # {2020.acl-main.420/},doi = "10.18653/v1/2020.acl-main.420",pages = "4609--4622"}

@article{rogers-etal-2020-primer,title = "A Primer in {BERT}ology: What We Know About How {BERT} Works",author = "Rogers, Anna and Kovaleva, Olga and Rumshisky, Anna",editor = "Johnson, Mark and Roark, Brian and Nenkova, Ani",journal = "Transactions of the Association for Computational Linguistics",volume = "8",year = "2020",address = "Cambridge, MA",publisher = "MIT Press",url = anth # {2020.tacl-1.54/},doi = "10.1162/tacl_a_00349",pages = "842--866"}

@inproceedings{rudman-etal-2023-outlier,title = "Outlier Dimensions Encode Task Specific Knowledge",author = "Rudman, William and Chen, Catherine and Eickhoff, Carsten",editor = "Bouamor, Houda and Pino, Juan and Bali, Kalika",booktitle = "Proceedings of the 2023 Conference on Empirical Methods in Natural Language Processing",month = dec,year = "2023",address = "Singapore",publisher = acl,url = anth # {2023.emnlp-main.901/},doi = "10.18653/v1/2023.emnlp-main.901",pages = "14596--14605"}

@inproceedings{subramani-etal-2022-extracting,title = "Extracting Latent Steering Vectors from Pretrained Language Models",author = "Subramani, Nishant and Suresh, Nivedita and Peters, Matthew",editor = "Muresan, Smaranda and Nakov, Preslav and Villavicencio, Aline",booktitle = "Findings of the Association for Computational Linguistics: ACL 2022",month = may,year = "2022",address = "Dublin, Ireland",publisher = acl,url = anth # {2022.findings-acl.48/},doi = "10.18653/v1/2022.findings-acl.48",pages = "566--581"}

@inproceedings{subramani-etal-2025-mice,title = "{MICE} for {CAT}s: Model-Internal Confidence Estimation for Calibrating Agents with Tools",author = "Subramani, Nishant and Eisner, Jason and Svegliato, Justin and Van Durme, Benjamin and Su, Yu and Thomson, Sam",editor = "Chiruzzo, Luis and Ritter, Alan and Wang, Lu",booktitle = "Proceedings of the 2025 Conference of the Nations of the Americas Chapter of the Association for Computational Linguistics: Human Language Technologies (Volume 1: Long Papers)",month = apr,year = "2025",address = "Albuquerque, New Mexico",publisher = acl,url = anth # {2025.naacl-long.615/},doi = "10.18653/v1/2025.naacl-long.615",pages = "12362--12375",ISBN = "979-8-89176-189-6"}

@inproceedings{sulubacak-etal-2016-universal,title = "{U}niversal {D}ependencies for {T}urkish",author = {Sulubacak, Umut and Gokirmak, Memduh and Tyers, Francis and {\c{C}}{\"o}ltekin, {\c{C}}a{\u{g}}r{\i} and Nivre, Joakim and Eryi{\u{g}}it, G{\"u}l{\c{s}}en},editor = "Matsumoto, Yuji and Prasad, Rashmi",booktitle = "Proceedings of {COLING} 2016, the 26th International Conference on Computational Linguistics: Technical Papers",month = dec,year = "2016",address = "Osaka, Japan",publisher = "The COLING 2016 Organizing Committee",url = anth # {C16-1325/},pages = "3444--3454"}

@inproceedings{tenney-etal-2019-bert,title = "{BERT} Rediscovers the Classical {NLP} Pipeline",author = "Tenney, Ian and Das, Dipanjan and Pavlick, Ellie",editor = "Korhonen, Anna and Traum, David and M{\`a}rquez, Llu{\'i}s",booktitle = "Proceedings of the 57th Annual Meeting of the Association for Computational Linguistics",month = jul,year = "2019",address = "Florence, Italy",publisher = acl,url = anth # {P19-1452/},doi = "10.18653/v1/P19-1452",pages = "4593--4601"}

@inproceedings{voita-titov-2020-information,title = "Information-Theoretic Probing with Minimum Description Length",author = "Voita, Elena and Titov, Ivan",editor = "Webber, Bonnie and Cohn, Trevor and He, Yulan and Liu, Yang",booktitle = "Proceedings of the 2020 Conference on Empirical Methods in Natural Language Processing (EMNLP)",month = nov,year = "2020",address = "Online",publisher = acl,url = anth # {2020.emnlp-main.14/},doi = "10.18653/v1/2020.emnlp-main.14",pages = "183--196"}

@inproceedings{vulic-etal-2020-probing,title = "Probing Pretrained Language Models for Lexical Semantics",author = "Vuli{\'c}, Ivan and Ponti, Edoardo Maria and Litschko, Robert and Glava{\v{s}}, Goran and Korhonen, Anna",editor = "Webber, Bonnie and Cohn, Trevor and He, Yulan and Liu, Yang",booktitle = "Proceedings of the 2020 Conference on Empirical Methods in Natural Language Processing (EMNLP)",month = nov,year = "2020",address = "Online",publisher = acl,url = anth # {2020.emnlp-main.586/},doi = "10.18653/v1/2020.emnlp-main.586",pages = "7222--7240"}

@inproceedings{xue-etal-2021-mt5,title = "m{T}5: A Massively Multilingual Pre-trained Text-to-Text Transformer",author = "Xue, Linting and Constant, Noah and Roberts, Adam and Kale, Mihir and Al-Rfou, Rami and Siddhant, Aditya and Barua, Aditya and Raffel, Colin",editor = "Toutanova, Kristina and Rumshisky, Anna and Zettlemoyer, Luke and Hakkani-Tur, Dilek and Beltagy, Iz and Bethard, Steven and Cotterell, Ryan and Chakraborty, Tanmoy and Zhou, Yichao",booktitle = "Proceedings of the 2021 Conference of the North American Chapter of the Association for Computational Linguistics: Human Language Technologies",month = jun,year = "2021",address = "Online",publisher = acl,url = anth # {2021.naacl-main.41/},doi = "10.18653/v1/2021.naacl-main.41",pages = "483--498"}
\clearpage
\appendix

\section{Additional Related Work}
\label{sec:additional_related_work}

\subsection{Advanced Probing Methodologies}
Beyond standard linear probes, sophisticated approaches have emerged to understand model representations. Amnesic probing \citep{elazar-etal-2021-amnesic} removes specific information from representations to test whether it's necessary for downstream tasks. Information-theoretic probing frameworks formalize probing as estimating how much information representations contain about linguistic structure \citep{pimentel-etal-2020-information}, while minimum description length probes \citep{voita-titov-2020-information} balance probe complexity with performance to avoid overfitting. Causal abstraction \citep{geiger2021causal} aims to establish causal rather than merely correlational relationships between representations and linguistic properties.

Complementary work has revisited what layerwise probing results imply: \citet{niu-etal-2022-bert} re-examine prior evidence for pipeline-like separation of linguistic knowledge across BERT layers, arguing that the pattern is more nuanced than depth alone explains. Recently, \citet{subramani-etal-2025-mice} and \citet{liu2025llm} find that decoding from or probing intermediate activations can yield reliable confidence and correctness estimators for LLMs~\citep{nostalgebraist2020logitlens}.

For morphology specifically, \citet{Acs_Hamerlik_Schwartz_Smith_Kornai_2024} introduced an extensive multilingual probing dataset (247 tasks across 42 languages), finding that \texttt{mBERT} and \texttt{XLM-RoBERTa} encode morphosyntactic features strongly, with preceding context more informative than following context for morphological prediction. Using minimal pairs, \citet{he-etal-2024-decoding} found that GPT-2 captures syntactic structure in its first third of layers, with morphological and semantics-syntax interface features proving harder to decode than pure syntax. \citet{dang-etal-2025-tokenization} compared mT5 and ByT5 on morphological probing, finding that linear classifiers match MLP performance, suggesting morphological features are encoded in linearly separable subspaces; tokenization strategies significantly impact morphological representation quality, particularly for morphologically rich languages.

\subsection{Model Manipulation and Steering}
Steering vectors demonstrate that specific directions in activation space correspond to high-level behavioral changes~\citep{NEURIPS2019_48c8c396, DBLP:journals/corr/abs-2008-09049, subramani-etal-2022-extracting}. Building on this, \citet{panickssery2024steeringllama2contrastive} achieves behavioral control by adding activation differences between contrasting examples. \citet{li2023inferencetime} introduce inference-time intervention, shifting model activations during inference across limited attention heads to control behavior. While these methods operate in activation space, task vectors enable arithmetic operations on model capabilities by manipulating weight space~\citep{ilharco2023editing}. Sparse autoencoders provide another avenue for feature discovery~\citep{bricken2023monosemanticity}, while causal mediation analysis~\citep{vig2020investigating} helps identify which components mediate specific behaviors.

\section{Probe Training Details}

\label{sec:probe_details}

We stratify each dataset into train, validation, and test splits. Probes are trained on the training split, hyperparameters are selected using the validation split, and we report accuracy and macro F1 on the held-out test split. For the linear regression probe we apply ridge regularization with $\lambda = 0.01$ and solve \cref{eq:ridge} in closed form. For the MLP probe we use a hidden dimension of 64, a learning rate of 0.001, weight decay of 0.01, and train for up to 100 epochs with early stopping based on validation loss, optimizing cross-entropy with AdamW. Both probes share the same data splits to enable fair comparison.

\begin{figure*}[htbp]
  \centering
  \includegraphics[width=\textwidth]{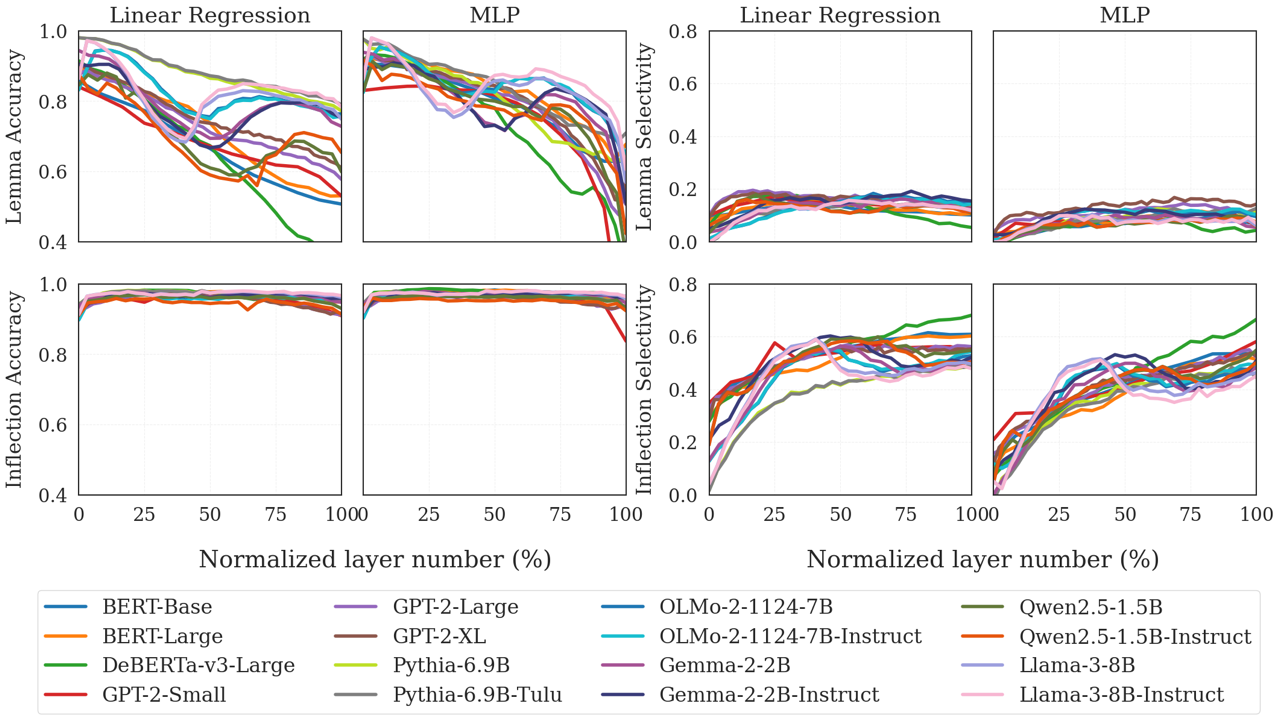}
  \caption{Full lemma and inflection probing results for English, showing individual curves for every model. Columns show prediction accuracy (Linear vs.\ MLP probes) and selectivity scores (linguistic minus control accuracy).}
  \label{fig:appendix_linguistic_and_selectivity}
\end{figure*}

\begin{figure*}[htbp]
  \centering
  \includegraphics[width=\textwidth]{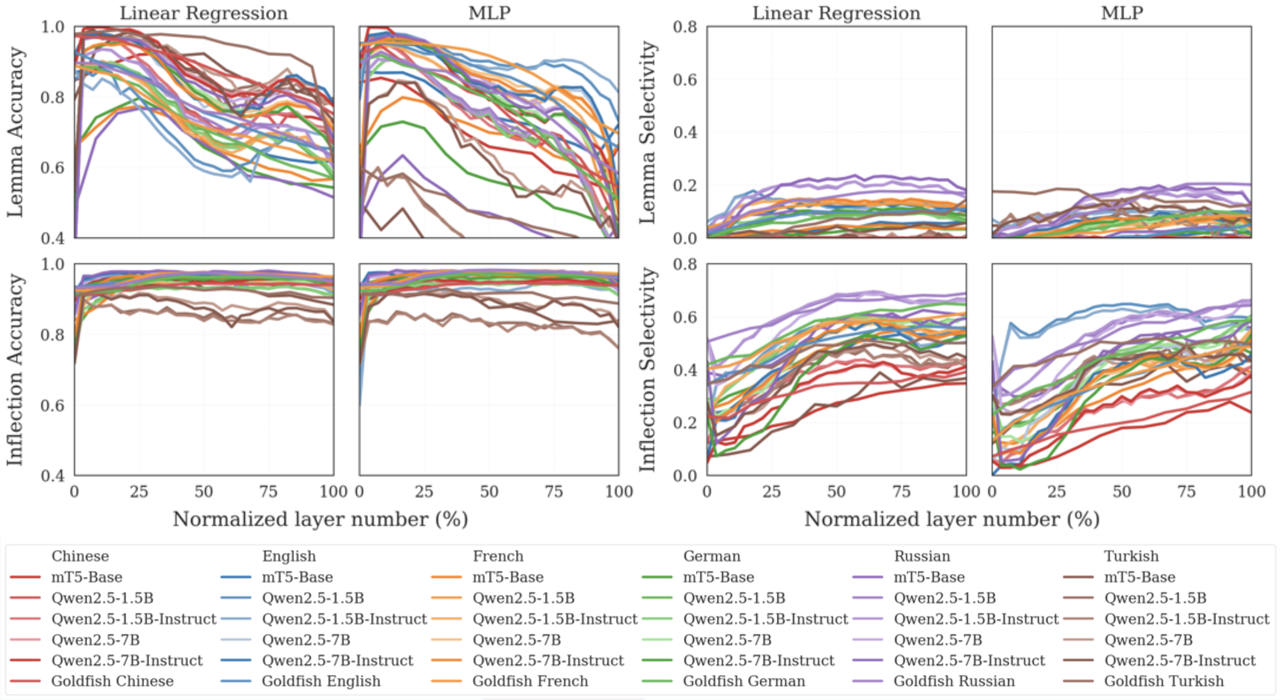}
  \caption{Full cross-linguistic probing results showing individual curves for every model within each language. Columns show lemma and inflection accuracy (Linear vs.\ MLP) followed by selectivity scores.}
  \label{fig:appendix_all_languages_combined}
\end{figure*}

\section{Full Lemma and Inflection Probe Results}
\label{sec:full_linguistic_results}

We provide the full, non-averaged results for the linguistic probing tasks (lemma identity and inflectional features) for every individual model. \Cref{fig:appendix_linguistic_and_selectivity} shows the detailed breakdown for English models, and \Cref{fig:appendix_all_languages_combined} presents the results for all six languages.

\section{Layer-wise Tables for Lemma and Inflection Results}
\label{sec:detailed_layerwise_tables}

This section contains detailed tables for layer-wise accuracy and selectivity across all models and languages. 

For English, we provide separate tables for each probe type and metric:
\begin{itemize}[itemsep=1pt]
    \item \Cref{tab:english_linear_accuracy}: Accuracy using linear probes
    \item \Cref{tab:english_linear_selectivity}: Selectivity using linear probes
    \item \Cref{tab:english_mlp_accuracy}: Accuracy using MLP probes
    \item \Cref{tab:english_mlp_selectivity}: Selectivity using MLP probes
\end{itemize}

For the cross-linguistic experiments, we provide combined tables showing both linear and MLP results for accuracy and selectivity:
\begin{itemize}[itemsep=1pt]
    \item \Cref{tab:probes_chinese}: Probing results for Chinese
    \item \Cref{tab:probes_french}: Probing results for French
    \item \Cref{tab:probes_german}: Probing results for German
    \item \Cref{tab:probes_russian}: Probing results for Russian
    \item \Cref{tab:probes_turkish}: Probing results for Turkish
\end{itemize}

\begin{table*}[h!]\centering
\caption{English Accuracy (Linear Probes) Across Model Depth}
\label{tab:english_linear_accuracy}
\resizebox{0.8\textwidth}{!}{%
\begin{tabular}{llccccc}
\toprule
& & \multicolumn{5}{c}{\textbf{Relative Depth (\%)}} \\
\cmidrule(lr){3-7}
\textbf{Model} & \textbf{Task} & \textbf{0\%} & \textbf{25\%} & \textbf{50\%} & \textbf{75\%} & \textbf{100\%} \\
\midrule
\multirow{2}{*}{BERT-Base} & Inflection & 0.934 & 0.971 & 0.977 & 0.966 & 0.951 \\
 & Lexeme & 0.858 & 0.773 & 0.665 & 0.564 & 0.507 \\
\addlinespace
\multirow{2}{*}{BERT-Large} & Inflection & 0.938 & 0.972 & 0.978 & 0.966 & 0.949 \\
 & Lexeme & 0.896 & 0.820 & 0.737 & 0.585 & 0.531 \\
\addlinespace
\multirow{2}{*}{DeBERTa-v3-Large} & Inflection & 0.939 & 0.982 & 0.972 & 0.960 & 0.955 \\
 & Lexeme & 0.914 & 0.805 & 0.673 & 0.482 & 0.309 \\
\addlinespace
\multirow{2}{*}{Gemma-2-2B} & Inflection & 0.936 & 0.974 & 0.963 & 0.972 & 0.948 \\
 & Lexeme & 0.944 & 0.817 & 0.694 & 0.792 & 0.728 \\
\addlinespace
\multirow{2}{*}{Gemma-2-2B-Instruct} & Inflection & 0.917 & 0.971 & 0.960 & 0.972 & 0.960 \\
 & Lexeme & 0.904 & 0.802 & 0.667 & 0.791 & 0.752 \\
\addlinespace
\multirow{2}{*}{Goldfish English} & Inflection & 0.934 & 0.956 & 0.972 & 0.964 & 0.949 \\
 & Lexeme & 0.926 & 0.859 & 0.755 & 0.690 & 0.652 \\
\addlinespace
\multirow{2}{*}{GPT-2-Large} & Inflection & 0.927 & 0.970 & 0.971 & 0.946 & 0.912 \\
 & Lexeme & 0.874 & 0.818 & 0.711 & 0.665 & 0.577 \\
\addlinespace
\multirow{2}{*}{GPT-2-Small} & Inflection & 0.939 & 0.948 & 0.971 & 0.956 & 0.909 \\
 & Lexeme & 0.840 & 0.737 & 0.671 & 0.618 & 0.530 \\
\addlinespace
\multirow{2}{*}{GPT-2-XL} & Inflection & 0.929 & 0.974 & 0.973 & 0.946 & 0.915 \\
 & Lexeme & 0.906 & 0.827 & 0.737 & 0.690 & 0.609 \\
\addlinespace
\multirow{2}{*}{Llama-3-8B} & Inflection & 0.912 & 0.971 & 0.974 & 0.976 & 0.962 \\
 & Lexeme & 0.864 & 0.794 & 0.796 & 0.816 & 0.749 \\
\addlinespace
\multirow{2}{*}{Llama-3-8B-Instruct} & Inflection & 0.913 & 0.972 & 0.974 & 0.977 & 0.967 \\
 & Lexeme & 0.864 & 0.803 & 0.812 & 0.839 & 0.783 \\
\addlinespace
\multirow{2}{*}{mT5-Base} & Inflection & 0.920 & 0.966 & 0.969 & 0.966 & 0.958 \\
 & Lexeme & 0.868 & 0.862 & 0.731 & 0.634 & 0.619 \\
\addlinespace
\multirow{2}{*}{OLMo-2-1124-7B} & Inflection & 0.897 & 0.970 & 0.959 & 0.961 & 0.965 \\
 & Lexeme & 0.832 & 0.883 & 0.755 & 0.808 & 0.763 \\
\addlinespace
\multirow{2}{*}{OLMo-2-1124-7B-Instruct} & Inflection & 0.897 & 0.970 & 0.958 & 0.962 & 0.961 \\
 & Lexeme & 0.832 & 0.881 & 0.749 & 0.806 & 0.757 \\
\addlinespace
\multirow{2}{*}{Pythia-6.9B} & Inflection & 0.942 & 0.982 & 0.974 & 0.966 & 0.953 \\
 & Lexeme & 0.980 & 0.928 & 0.865 & 0.829 & 0.772 \\
\addlinespace
\multirow{2}{*}{Pythia-6.9B-Tulu} & Inflection & 0.941 & 0.980 & 0.975 & 0.968 & 0.954 \\
 & Lexeme & 0.980 & 0.928 & 0.872 & 0.841 & 0.789 \\
\addlinespace
\multirow{2}{*}{Qwen2.5-1.5B} & Inflection & 0.913 & 0.969 & 0.959 & 0.961 & 0.930 \\
 & Lexeme & 0.845 & 0.799 & 0.610 & 0.654 & 0.599 \\
\addlinespace
\multirow{2}{*}{Qwen2.5-1.5B-Instruct} & Inflection & 0.910 & 0.957 & 0.944 & 0.949 & 0.914 \\
 & Lexeme & 0.876 & 0.768 & 0.590 & 0.647 & 0.654 \\
\addlinespace
\multirow{2}{*}{Qwen2.5-7B} & Inflection & 0.915 & 0.977 & 0.966 & 0.973 & 0.958 \\
 & Lexeme & 0.916 & 0.952 & 0.769 & 0.808 & 0.781 \\
\addlinespace
\multirow{2}{*}{Qwen2.5-7B-Instruct} & Inflection & 0.915 & 0.977 & 0.964 & 0.973 & 0.957 \\
 & Lexeme & 0.916 & 0.950 & 0.791 & 0.810 & 0.781 \\
\bottomrule
\end{tabular}%
}
\end{table*}

\begin{table*}[h!]
\centering
\caption{English Selectivity (Linear Probes) Across Model Depth}
\label{tab:english_linear_selectivity}
\resizebox{0.8\textwidth}{!}{%
\begin{tabular}{llccccc}
\toprule
& & \multicolumn{5}{c}{\textbf{Relative Depth (\%)}} \\
\cmidrule(lr){3-7}
\textbf{Model} & \textbf{Task} & \textbf{0\%} & \textbf{25\%} & \textbf{50\%} & \textbf{75\%} & \textbf{100\%} \\
\midrule
\multirow{2}{*}{BERT-Base} & Inflection & 0.348 & 0.486 & 0.556 & 0.596 & 0.609 \\
 & Lexeme & 0.094 & 0.132 & 0.134 & 0.114 & 0.101 \\
\addlinespace
\multirow{2}{*}{BERT-Large} & Inflection & 0.292 & 0.459 & 0.520 & 0.598 & 0.603 \\
 & Lexeme & 0.083 & 0.145 & 0.148 & 0.119 & 0.107 \\
\addlinespace
\multirow{2}{*}{DeBERTa-v3-Large} & Inflection & 0.279 & 0.471 & 0.545 & 0.644 & 0.681 \\
 & Lexeme & 0.062 & 0.183 & 0.158 & 0.097 & 0.054 \\
\addlinespace
\multirow{2}{*}{Gemma-2-2B} & Inflection & 0.131 & 0.466 & 0.565 & 0.464 & 0.512 \\
 & Lexeme & 0.036 & 0.132 & 0.136 & 0.161 & 0.128 \\
\addlinespace
\multirow{2}{*}{Gemma-2-2B-Instruct} & Inflection & 0.210 & 0.517 & 0.594 & 0.487 & 0.522 \\
 & Lexeme & 0.052 & 0.157 & 0.144 & 0.184 & 0.154 \\
\addlinespace
\multirow{2}{*}{Goldfish English} & Inflection & 0.341 & 0.437 & 0.522 & 0.549 & 0.556 \\
 & Lexeme & 0.017 & 0.073 & 0.096 & 0.100 & 0.110 \\
\addlinespace
\multirow{2}{*}{GPT-2-Large} & Inflection & 0.324 & 0.489 & 0.563 & 0.550 & 0.562 \\
 & Lexeme & 0.099 & 0.183 & 0.162 & 0.161 & 0.121 \\
\addlinespace
\multirow{2}{*}{GPT-2-Small} & Inflection & 0.347 & 0.576 & 0.541 & 0.558 & 0.547 \\
 & Lexeme & 0.100 & 0.171 & 0.152 & 0.134 & 0.121 \\
\addlinespace
\multirow{2}{*}{GPT-2-XL} & Inflection & 0.289 & 0.488 & 0.549 & 0.547 & 0.553 \\
 & Lexeme & 0.092 & 0.172 & 0.172 & 0.162 & 0.135 \\
\addlinespace
\multirow{2}{*}{Llama-3-8B} & Inflection & 0.041 & 0.516 & 0.482 & 0.474 & 0.497 \\
 & Lexeme & -0.003 & 0.132 & 0.146 & 0.140 & 0.117 \\
\addlinespace
\multirow{2}{*}{Llama-3-8B-Instruct} & Inflection & 0.042 & 0.508 & 0.473 & 0.453 & 0.478 \\
 & Lexeme & -0.003 & 0.131 & 0.146 & 0.144 & 0.119 \\
\addlinespace
\multirow{2}{*}{mT5-Base} & Inflection & 0.229 & 0.348 & 0.462 & 0.533 & 0.530 \\
 & Lexeme & 0.003 & 0.005 & 0.047 & 0.053 & 0.063 \\
\addlinespace
\multirow{2}{*}{OLMo-2-1124-7B} & Inflection & 0.128 & 0.440 & 0.538 & 0.476 & 0.528 \\
 & Lexeme & 0.011 & 0.104 & 0.152 & 0.170 & 0.144 \\
\addlinespace
\multirow{2}{*}{OLMo-2-1124-7B-Instruct} & Inflection & 0.127 & 0.443 & 0.545 & 0.483 & 0.543 \\
 & Lexeme & 0.011 & 0.104 & 0.149 & 0.164 & 0.144 \\
\addlinespace
\multirow{2}{*}{Pythia-6.9B} & Inflection & 0.017 & 0.347 & 0.416 & 0.457 & 0.484 \\
 & Lexeme & -0.002 & 0.117 & 0.134 & 0.140 & 0.141 \\
\addlinespace
\multirow{2}{*}{Pythia-6.9B-Tulu} & Inflection & 0.016 & 0.348 & 0.419 & 0.454 & 0.492 \\
 & Lexeme & -0.002 & 0.118 & 0.137 & 0.143 & 0.150 \\
\addlinespace
\multirow{2}{*}{Qwen2.5-1.5B} & Inflection & 0.199 & 0.483 & 0.589 & 0.566 & 0.548 \\
 & Lexeme & 0.034 & 0.146 & 0.117 & 0.129 & 0.108 \\
\addlinespace
\multirow{2}{*}{Qwen2.5-1.5B-Instruct} & Inflection & 0.189 & 0.474 & 0.581 & 0.541 & 0.503 \\
 & Lexeme & 0.061 & 0.143 & 0.113 & 0.129 & 0.112 \\
\addlinespace
\multirow{2}{*}{Qwen2.5-7B} & Inflection & 0.059 & 0.383 & 0.538 & 0.533 & 0.523 \\
 & Lexeme & -0.006 & 0.098 & 0.121 & 0.131 & 0.099 \\
\addlinespace
\multirow{2}{*}{Qwen2.5-7B-Instruct} & Inflection & 0.059 & 0.392 & 0.542 & 0.531 & 0.528 \\
 & Lexeme & -0.006 & 0.098 & 0.116 & 0.129 & 0.100 \\
\bottomrule
\end{tabular}%
}
\end{table*}

\begin{table*}[h!]
\centering
\caption{English Accuracy (MLP Probes) Across Model Depth}
\label{tab:english_mlp_accuracy}
\resizebox{0.8\textwidth}{!}{%
\begin{tabular}{llccccc}
\toprule
& & \multicolumn{5}{c}{\textbf{Relative Depth (\%)}} \\
\cmidrule(lr){3-7}
\textbf{Model} & \textbf{Task} & \textbf{0\%} & \textbf{25\%} & \textbf{50\%} & \textbf{75\%} & \textbf{100\%} \\
\midrule
\multirow{2}{*}{BERT-Base} & Inflection & 0.941 & 0.978 & 0.983 & 0.975 & 0.960 \\
 & Lexeme & 0.906 & 0.888 & 0.808 & 0.719 & 0.629 \\
\addlinespace
\multirow{2}{*}{BERT-Large} & Inflection & 0.943 & 0.977 & 0.981 & 0.973 & 0.958 \\
 & Lexeme & 0.920 & 0.904 & 0.843 & 0.762 & 0.677 \\
\addlinespace
\multirow{2}{*}{DeBERTa-v3-Large} & Inflection & 0.944 & 0.986 & 0.976 & 0.971 & 0.957 \\
 & Lexeme & 0.920 & 0.885 & 0.781 & 0.574 & 0.318 \\
\addlinespace
\multirow{2}{*}{Gemma-2-2B} & Inflection & 0.940 & 0.977 & 0.970 & 0.975 & 0.946 \\
 & Lexeme & 0.939 & 0.868 & 0.732 & 0.812 & 0.506 \\
\addlinespace
\multirow{2}{*}{Gemma-2-2B-Instruct} & Inflection & 0.919 & 0.973 & 0.965 & 0.973 & 0.960 \\
 & Lexeme & 0.890 & 0.874 & 0.731 & 0.831 & 0.508 \\
\addlinespace
\multirow{2}{*}{Goldfish English} & Inflection & 0.937 & 0.962 & 0.976 & 0.971 & 0.964 \\
 & Lexeme & 0.952 & 0.922 & 0.871 & 0.790 & 0.656 \\
\addlinespace
\multirow{2}{*}{GPT-2-Large} & Inflection & 0.930 & 0.964 & 0.964 & 0.952 & 0.937 \\
 & Lexeme & 0.892 & 0.878 & 0.825 & 0.719 & 0.552 \\
\addlinespace
\multirow{2}{*}{GPT-2-Small} & Inflection & 0.943 & 0.963 & 0.965 & 0.957 & 0.837 \\
 & Lexeme & 0.830 & 0.843 & 0.817 & 0.705 & 0.075 \\
\addlinespace
\multirow{2}{*}{GPT-2-XL} & Inflection & 0.929 & 0.965 & 0.963 & 0.948 & 0.937 \\
 & Lexeme & 0.906 & 0.884 & 0.844 & 0.734 & 0.571 \\
\addlinespace
\multirow{2}{*}{Llama-3-8B} & Inflection & 0.920 & 0.972 & 0.977 & 0.976 & 0.958 \\
 & Lexeme & 0.863 & 0.808 & 0.857 & 0.840 & 0.568 \\
\addlinespace
\multirow{2}{*}{Llama-3-8B-Instruct} & Inflection & 0.920 & 0.972 & 0.977 & 0.977 & 0.964 \\
 & Lexeme & 0.863 & 0.821 & 0.873 & 0.870 & 0.605 \\
\addlinespace
\multirow{2}{*}{mT5-Base} & Inflection & NaN & NaN & NaN & NaN & NaN \\
 & Lexeme & 0.871 & 0.845 & 0.744 & 0.686 & 0.722 \\
\addlinespace
\multirow{2}{*}{OLMo-2-1124-7B} & Inflection & 0.903 & 0.974 & 0.970 & 0.975 & 0.964 \\
 & Lexeme & 0.825 & 0.877 & 0.833 & 0.845 & 0.621 \\
\addlinespace
\multirow{2}{*}{OLMo-2-1124-7B-Instruct} & Inflection & 0.903 & 0.975 & 0.968 & 0.973 & 0.964 \\
 & Lexeme & 0.825 & 0.880 & 0.825 & 0.847 & 0.650 \\
\addlinespace
\multirow{2}{*}{Pythia-6.9B} & Inflection & 0.940 & 0.973 & 0.971 & 0.959 & 0.959 \\
 & Lexeme & 0.976 & 0.891 & 0.823 & 0.683 & 0.655 \\
\addlinespace
\multirow{2}{*}{Pythia-6.9B-Tulu} & Inflection & 0.944 & 0.973 & 0.970 & 0.963 & 0.961 \\
 & Lexeme & 0.976 & 0.904 & 0.858 & 0.752 & 0.709 \\
\addlinespace
\multirow{2}{*}{Qwen2.5-1.5B} & Inflection & 0.666 & 0.956 & 0.956 & 0.961 & 0.929 \\
 & Lexeme & 0.792 & 0.959 & 0.901 & 0.886 & 0.731 \\
\addlinespace
\multirow{2}{*}{Qwen2.5-1.5B-Instruct} & Inflection & 0.598 & 0.922 & 0.928 & 0.942 & 0.913 \\
 & Lexeme & 0.852 & 0.939 & 0.880 & 0.900 & 0.812 \\
\addlinespace
\multirow{2}{*}{Qwen2.5-7B} & Inflection & 0.919 & 0.970 & 0.963 & 0.970 & 0.953 \\
 & Lexeme & 0.913 & 0.935 & 0.831 & 0.818 & 0.506 \\
\addlinespace
\multirow{2}{*}{Qwen2.5-7B-Instruct} & Inflection & 0.930 & 0.976 & 0.970 & 0.976 & 0.951 \\
 & Lexeme & 0.913 & 0.933 & 0.824 & 0.818 & 0.521 \\
\bottomrule
\end{tabular}%
}
\end{table*}

\begin{table*}[h!]
\centering
\caption{English Selectivity (MLP Probes) Across Model Depth}
\label{tab:english_mlp_selectivity}
\resizebox{0.8\textwidth}{!}{%
\begin{tabular}{llccccc}
\toprule
& & \multicolumn{5}{c}{\textbf{Relative Depth (\%)}} \\
\cmidrule(lr){3-7}
\textbf{Model} & \textbf{Task} & \textbf{0\%} & \textbf{25\%} & \textbf{50\%} & \textbf{75\%} & \textbf{100\%} \\
\midrule
\multirow{2}{*}{BERT-Base} & Inflection & 0.139 & 0.307 & 0.416 & 0.505 & 0.536 \\
 & Lexeme & 0.025 & 0.057 & 0.069 & 0.084 & 0.068 \\
\addlinespace
\multirow{2}{*}{BERT-Large} & Inflection & 0.104 & 0.294 & 0.386 & 0.496 & 0.514 \\
 & Lexeme & 0.028 & 0.084 & 0.088 & 0.106 & 0.074 \\
\addlinespace
\multirow{2}{*}{DeBERTa-v3-Large} & Inflection & 0.071 & 0.313 & 0.428 & 0.582 & 0.666 \\
 & Lexeme & 0.005 & 0.061 & 0.096 & 0.065 & 0.043 \\
\addlinespace
\multirow{2}{*}{Gemma-2-2B} & Inflection & 0.002 & 0.360 & 0.499 & 0.395 & 0.483 \\
 & Lexeme & -0.007 & 0.067 & 0.090 & 0.088 & 0.054 \\
\addlinespace
\multirow{2}{*}{Gemma-2-2B-Instruct} & Inflection & 0.089 & 0.393 & 0.523 & 0.404 & 0.509 \\
 & Lexeme & 0.026 & 0.102 & 0.111 & 0.100 & 0.058 \\
\addlinespace
\multirow{2}{*}{Goldfish English} & Inflection & 0.131 & 0.272 & 0.371 & 0.438 & 0.493 \\
 & Lexeme & 0.006 & -0.021 & 0.005 & 0.031 & 0.050 \\
\addlinespace
\multirow{2}{*}{GPT-2-Large} & Inflection & 0.157 & 0.323 & 0.446 & 0.487 & 0.521 \\
 & Lexeme & 0.040 & 0.093 & 0.123 & 0.138 & 0.115 \\
\addlinespace
\multirow{2}{*}{GPT-2-Small} & Inflection & 0.210 & 0.342 & 0.410 & 0.470 & 0.581 \\
 & Lexeme & 0.012 & 0.077 & 0.094 & 0.107 & 0.061 \\
\addlinespace
\multirow{2}{*}{GPT-2-XL} & Inflection & 0.118 & 0.334 & 0.433 & 0.491 & 0.527 \\
 & Lexeme & 0.027 & 0.110 & 0.145 & 0.163 & 0.142 \\
\addlinespace
\multirow{2}{*}{Llama-3-8B} & Inflection & 0.050 & 0.454 & 0.414 & 0.402 & 0.463 \\
 & Lexeme & -0.001 & 0.098 & 0.079 & 0.084 & 0.060 \\
\addlinespace
\multirow{2}{*}{Llama-3-8B-Instruct} & Inflection & 0.050 & 0.442 & 0.400 & 0.364 & 0.451 \\
 & Lexeme & -0.002 & 0.098 & 0.074 & 0.084 & 0.069 \\
\addlinespace
\multirow{2}{*}{mT5-Base} & Inflection & NaN & NaN & NaN & NaN & NaN \\
 & Lexeme & -0.012 & -0.023 & 0.003 & 0.011 & 0.026 \\
\addlinespace
\multirow{2}{*}{OLMo-2-1124-7B} & Inflection & 0.118 & 0.363 & 0.470 & 0.426 & 0.483 \\
 & Lexeme & 0.007 & 0.088 & 0.121 & 0.109 & 0.095 \\
\addlinespace
\multirow{2}{*}{OLMo-2-1124-7B-Instruct} & Inflection & 0.118 & 0.371 & 0.476 & 0.434 & 0.493 \\
 & Lexeme & 0.007 & 0.094 & 0.122 & 0.119 & 0.103 \\
\addlinespace
\multirow{2}{*}{Pythia-6.9B} & Inflection & -0.025 & 0.301 & 0.404 & 0.444 & 0.451 \\
 & Lexeme & -0.005 & 0.064 & 0.096 & 0.110 & 0.113 \\
\addlinespace
\multirow{2}{*}{Pythia-6.9B-Tulu} & Inflection & -0.021 & 0.294 & 0.394 & 0.438 & 0.445 \\
 & Lexeme & -0.005 & 0.057 & 0.092 & 0.112 & 0.126 \\
\addlinespace
\multirow{2}{*}{Qwen2.5-1.5B} & Inflection & 0.248 & 0.588 & 0.647 & 0.625 & 0.595 \\
 & Lexeme & 0.019 & 0.085 & 0.096 & 0.101 & 0.090 \\
\addlinespace
\multirow{2}{*}{Qwen2.5-1.5B-Instruct} & Inflection & 0.187 & 0.571 & 0.629 & 0.608 & 0.549 \\
 & Lexeme & 0.069 & 0.090 & 0.097 & 0.100 & 0.087 \\
\addlinespace
\multirow{2}{*}{Qwen2.5-7B} & Inflection & 0.051 & 0.282 & 0.456 & 0.457 & 0.507 \\
 & Lexeme & -0.008 & 0.039 & 0.070 & 0.073 & 0.061 \\
\addlinespace
\multirow{2}{*}{Qwen2.5-7B-Instruct} & Inflection & 0.001 & 0.229 & 0.424 & 0.416 & 0.457 \\
 & Lexeme & -0.008 & 0.042 & 0.072 & 0.067 & 0.068 \\
\bottomrule
\end{tabular}%
}
\end{table*}

\begin{table*}[h!]
\centering
\caption{Probing Results for Chinese}
\label{tab:probes_chinese}
\begin{subtable}{0.48\textwidth}
\centering
\caption{Accuracy (Linear Probes)}
\resizebox{\textwidth}{!}{%
\begin{tabular}{llccccc}
\toprule
& & \multicolumn{5}{c}{\textbf{Relative Depth (\%)}} \\
\cmidrule(lr){3-7}
\textbf{Model} & \textbf{Task} & \textbf{0\%} & \textbf{25\%} & \textbf{50\%} & \textbf{75\%} & \textbf{100\%} \\
\midrule
\multirow{2}{*}{Goldfish Chinese (Chinese)} & Inflection & 0.911 & 0.928 & 0.944 & 0.942 & 0.941 \\
 & Lexeme & 0.972 & 0.941 & 0.887 & 0.824 & 0.751 \\
\addlinespace
\multirow{2}{*}{Qwen2.5-1.5B (Chinese)} & Inflection & 0.898 & 0.948 & 0.949 & 0.950 & 0.946 \\
 & Lexeme & 0.883 & 0.905 & 0.735 & 0.743 & 0.667 \\
\addlinespace
\multirow{2}{*}{Qwen2.5-1.5B-Instruct (Chinese)} & Inflection & 0.897 & 0.948 & 0.949 & 0.950 & 0.948 \\
 & Lexeme & 0.883 & 0.907 & 0.729 & 0.748 & 0.678 \\
\addlinespace
\multirow{2}{*}{Qwen2.5-7B (Chinese)} & Inflection & 0.893 & 0.957 & 0.951 & 0.956 & 0.950 \\
 & Lexeme & 0.883 & 0.983 & 0.844 & 0.828 & 0.776 \\
\addlinespace
\multirow{2}{*}{Qwen2.5-7B-Instruct (Chinese)} & Inflection & 0.893 & 0.957 & 0.950 & 0.956 & 0.949 \\
 & Lexeme & 0.883 & 0.981 & 0.839 & 0.823 & 0.773 \\
\addlinespace
\multirow{2}{*}{mT5-Base (Chinese)} & Inflection & 0.901 & 0.933 & 0.945 & 0.941 & 0.943 \\
 & Lexeme & 0.846 & 0.919 & 0.863 & 0.757 & 0.727 \\
\bottomrule
\end{tabular}%
}
\end{subtable}
\hfill
\begin{subtable}{0.48\textwidth}
\centering
\caption{Selectivity (Linear Probes)}
\resizebox{\textwidth}{!}{%
\begin{tabular}{llccccc}
\toprule
& & \multicolumn{5}{c}{\textbf{Relative Depth (\%)}} \\
\cmidrule(lr){3-7}
\textbf{Model} & \textbf{Task} & \textbf{0\%} & \textbf{25\%} & \textbf{50\%} & \textbf{75\%} & \textbf{100\%} \\
\midrule
\multirow{2}{*}{Goldfish Chinese (Chinese)} & Inflection & 0.223 & 0.292 & 0.346 & 0.356 & 0.391 \\
 & Lexeme & -0.000 & -0.001 & -0.002 & -0.001 & -0.003 \\
\addlinespace
\multirow{2}{*}{Qwen2.5-1.5B (Chinese)} & Inflection & 0.122 & 0.345 & 0.429 & 0.412 & 0.441 \\
 & Lexeme & -0.000 & -0.000 & -0.002 & -0.002 & -0.003 \\
\addlinespace
\multirow{2}{*}{Qwen2.5-1.5B-Instruct (Chinese)} & Inflection & 0.122 & 0.345 & 0.430 & 0.412 & 0.437 \\
 & Lexeme & -0.000 & -0.001 & -0.003 & -0.001 & -0.002 \\
\addlinespace
\multirow{2}{*}{Qwen2.5-7B (Chinese)} & Inflection & 0.047 & 0.250 & 0.387 & 0.386 & 0.408 \\
 & Lexeme & -0.000 & 0.000 & -0.001 & 0.000 & -0.000 \\
\addlinespace
\multirow{2}{*}{Qwen2.5-7B-Instruct (Chinese)} & Inflection & 0.048 & 0.250 & 0.392 & 0.387 & 0.411 \\
 & Lexeme & -0.000 & -0.000 & -0.001 & 0.000 & -0.001 \\
\addlinespace
\multirow{2}{*}{mT5-Base (Chinese)} & Inflection & 0.123 & 0.186 & 0.274 & 0.321 & 0.348 \\
 & Lexeme & 0.001 & 0.000 & -0.001 & 0.000 & -0.003 \\
\bottomrule
\end{tabular}%
}
\end{subtable}

\vspace{0.75em}

\begin{subtable}{0.48\textwidth}
\centering
\caption{Accuracy (MLP Probes)}
\resizebox{\textwidth}{!}{%
\begin{tabular}{llccccc}
\toprule
& & \multicolumn{5}{c}{\textbf{Relative Depth (\%)}} \\
\cmidrule(lr){3-7}
\textbf{Model} & \textbf{Task} & \textbf{0\%} & \textbf{25\%} & \textbf{50\%} & \textbf{75\%} & \textbf{100\%} \\
\midrule
\multirow{2}{*}{Goldfish Chinese (Chinese)} & Inflection & 0.913 & 0.930 & 0.946 & 0.944 & 0.939 \\
 & Lexeme & 0.922 & 0.874 & 0.797 & 0.652 & 0.543 \\
\addlinespace
\multirow{2}{*}{Qwen2.5-1.5B (Chinese)} & Inflection & 0.896 & 0.947 & 0.943 & 0.950 & 0.942 \\
 & Lexeme & 0.882 & 0.869 & 0.738 & 0.695 & 0.449 \\
\addlinespace
\multirow{2}{*}{Qwen2.5-1.5B-Instruct (Chinese)} & Inflection & 0.896 & 0.941 & 0.942 & 0.950 & 0.943 \\
 & Lexeme & 0.883 & 0.864 & 0.719 & 0.691 & 0.383 \\
\addlinespace
\multirow{2}{*}{Qwen2.5-7B (Chinese)} & Inflection & 0.899 & 0.952 & 0.947 & 0.955 & 0.946 \\
 & Lexeme & 0.881 & 0.951 & 0.795 & 0.749 & 0.471 \\
\addlinespace
\multirow{2}{*}{Qwen2.5-7B-Instruct (Chinese)} & Inflection & 0.900 & 0.952 & 0.948 & 0.955 & 0.943 \\
 & Lexeme & 0.881 & 0.950 & 0.791 & 0.750 & 0.475 \\
\addlinespace
\multirow{2}{*}{mT5-Base (Chinese)} & Inflection & 0.907 & 0.938 & 0.947 & 0.942 & 0.948 \\
 & Lexeme & 0.841 & 0.796 & 0.658 & 0.587 & 0.661 \\
\bottomrule
\end{tabular}%
}
\end{subtable}
\hfill
\begin{subtable}{0.48\textwidth}
\centering
\caption{Selectivity (MLP Probes)}
\resizebox{\textwidth}{!}{%
\begin{tabular}{llccccc}
\toprule
& & \multicolumn{5}{c}{\textbf{Relative Depth (\%)}} \\
\cmidrule(lr){3-7}
\textbf{Model} & \textbf{Task} & \textbf{0\%} & \textbf{25\%} & \textbf{50\%} & \textbf{75\%} & \textbf{100\%} \\
\midrule
\multirow{2}{*}{Goldfish Chinese (Chinese)} & Inflection & 0.072 & 0.153 & 0.198 & 0.252 & 0.315 \\
 & Lexeme & -0.001 & -0.039 & -0.049 & -0.064 & -0.047 \\
\addlinespace
\multirow{2}{*}{Qwen2.5-1.5B (Chinese)} & Inflection & 0.058 & 0.168 & 0.280 & 0.305 & 0.411 \\
 & Lexeme & 0.000 & -0.001 & -0.006 & -0.001 & 0.002 \\
\addlinespace
\multirow{2}{*}{Qwen2.5-1.5B-Instruct (Chinese)} & Inflection & 0.059 & 0.164 & 0.290 & 0.301 & 0.411 \\
 & Lexeme & 0.000 & 0.002 & -0.012 & -0.000 & -0.002 \\
\addlinespace
\multirow{2}{*}{Qwen2.5-7B (Chinese)} & Inflection & 0.061 & 0.136 & 0.287 & 0.312 & 0.388 \\
 & Lexeme & -0.001 & -0.001 & -0.007 & -0.008 & 0.000 \\
\addlinespace
\multirow{2}{*}{Qwen2.5-7B-Instruct (Chinese)} & Inflection & 0.055 & 0.135 & 0.298 & 0.314 & 0.379 \\
 & Lexeme & -0.001 & -0.003 & -0.006 & -0.004 & -0.000 \\
\addlinespace
\multirow{2}{*}{mT5-Base (Chinese)} & Inflection & 0.055 & 0.070 & 0.179 & 0.238 & 0.238 \\
 & Lexeme & 0.003 & -0.008 & -0.013 & -0.006 & -0.011 \\
\bottomrule
\end{tabular}%
}
\end{subtable}
\end{table*}

\begin{table*}[h!]
\centering
\caption{Probing Results for French}
\label{tab:probes_french}
\begin{subtable}{0.48\textwidth}
\centering
\caption{Accuracy (Linear Probes)}
\resizebox{\textwidth}{!}{%
\begin{tabular}{llccccc}
\toprule
& & \multicolumn{5}{c}{\textbf{Relative Depth (\%)}} \\
\cmidrule(lr){3-7}
\textbf{Model} & \textbf{Task} & \textbf{0\%} & \textbf{25\%} & \textbf{50\%} & \textbf{75\%} & \textbf{100\%} \\
\midrule
\multirow{2}{*}{Goldfish French (French)} & Inflection & 0.924 & 0.959 & 0.976 & 0.970 & 0.963 \\
 & Lexeme & 0.888 & 0.813 & 0.714 & 0.665 & 0.619 \\
\addlinespace
\multirow{2}{*}{Qwen2.5-1.5B (French)} & Inflection & 0.792 & 0.947 & 0.954 & 0.952 & 0.928 \\
 & Lexeme & 0.541 & 0.850 & 0.696 & 0.696 & 0.602 \\
\addlinespace
\multirow{2}{*}{Qwen2.5-1.5B-Instruct (French)} & Inflection & 0.792 & 0.945 & 0.951 & 0.949 & 0.925 \\
 & Lexeme & 0.541 & 0.845 & 0.687 & 0.690 & 0.611 \\
\addlinespace
\multirow{2}{*}{Qwen2.5-7B (French)} & Inflection & 0.793 & 0.966 & 0.965 & 0.964 & 0.945 \\
 & Lexeme & 0.541 & 0.943 & 0.801 & 0.769 & 0.714 \\
\addlinespace
\multirow{2}{*}{Qwen2.5-7B-Instruct (French)} & Inflection & 0.793 & 0.963 & 0.962 & 0.961 & 0.941 \\
 & Lexeme & 0.541 & 0.942 & 0.790 & 0.760 & 0.706 \\
\addlinespace
\multirow{2}{*}{mT5-Base (French)} & Inflection & 0.840 & 0.943 & 0.967 & 0.961 & 0.944 \\
 & Lexeme & 0.656 & 0.773 & 0.674 & 0.596 & 0.567 \\
\bottomrule
\end{tabular}%
}
\end{subtable}
\hfill
\begin{subtable}{0.48\textwidth}
\centering
\caption{Selectivity (Linear Probes)}
\resizebox{\textwidth}{!}{%
\begin{tabular}{llccccc}
\toprule
& & \multicolumn{5}{c}{\textbf{Relative Depth (\%)}} \\
\cmidrule(lr){3-7}
\textbf{Model} & \textbf{Task} & \textbf{0\%} & \textbf{25\%} & \textbf{50\%} & \textbf{75\%} & \textbf{100\%} \\
\midrule
\multirow{2}{*}{Goldfish French (French)} & Inflection & 0.403 & 0.497 & 0.581 & 0.585 & 0.613 \\
 & Lexeme & 0.039 & 0.109 & 0.132 & 0.131 & 0.122 \\
\addlinespace
\multirow{2}{*}{Qwen2.5-1.5B (French)} & Inflection & 0.244 & 0.463 & 0.576 & 0.559 & 0.561 \\
 & Lexeme & 0.002 & 0.137 & 0.126 & 0.132 & 0.088 \\
\addlinespace
\multirow{2}{*}{Qwen2.5-1.5B-Instruct (French)} & Inflection & 0.244 & 0.467 & 0.582 & 0.565 & 0.562 \\
 & Lexeme & 0.002 & 0.137 & 0.126 & 0.132 & 0.092 \\
\addlinespace
\multirow{2}{*}{Qwen2.5-7B (French)} & Inflection & 0.228 & 0.370 & 0.538 & 0.545 & 0.540 \\
 & Lexeme & 0.003 & 0.116 & 0.145 & 0.139 & 0.112 \\
\addlinespace
\multirow{2}{*}{Qwen2.5-7B-Instruct (French)} & Inflection & 0.228 & 0.375 & 0.545 & 0.552 & 0.544 \\
 & Lexeme & 0.002 & 0.118 & 0.143 & 0.138 & 0.112 \\
\addlinespace
\multirow{2}{*}{mT5-Base (French)} & Inflection & 0.248 & 0.386 & 0.495 & 0.548 & 0.530 \\
 & Lexeme & -0.006 & 0.027 & 0.045 & 0.046 & 0.037 \\
\bottomrule
\end{tabular}%
}
\end{subtable}

\vspace{0.75em}

\begin{subtable}{0.48\textwidth}
\centering
\caption{Accuracy (MLP Probes)}
\resizebox{\textwidth}{!}{%
\begin{tabular}{llccccc}
\toprule
& & \multicolumn{5}{c}{\textbf{Relative Depth (\%)}} \\
\cmidrule(lr){3-7}
\textbf{Model} & \textbf{Task} & \textbf{0\%} & \textbf{25\%} & \textbf{50\%} & \textbf{75\%} & \textbf{100\%} \\
\midrule
\multirow{2}{*}{Goldfish French (French)} & Inflection & 0.932 & 0.972 & 0.980 & 0.979 & 0.971 \\
 & Lexeme & 0.947 & 0.949 & 0.916 & 0.829 & 0.696 \\
\addlinespace
\multirow{2}{*}{Qwen2.5-1.5B (French)} & Inflection & 0.789 & 0.956 & 0.965 & 0.960 & 0.929 \\
 & Lexeme & 0.536 & 0.910 & 0.845 & 0.788 & 0.360 \\
\addlinespace
\multirow{2}{*}{Qwen2.5-1.5B-Instruct (French)} & Inflection & 0.791 & 0.954 & 0.962 & 0.959 & 0.929 \\
 & Lexeme & 0.537 & 0.911 & 0.835 & 0.798 & 0.535 \\
\addlinespace
\multirow{2}{*}{Qwen2.5-7B (French)} & Inflection & 0.791 & 0.971 & 0.971 & 0.968 & 0.943 \\
 & Lexeme & 0.533 & 0.953 & 0.862 & 0.830 & 0.461 \\
\addlinespace
\multirow{2}{*}{Qwen2.5-7B-Instruct (French)} & Inflection & 0.791 & 0.968 & 0.969 & 0.965 & 0.939 \\
 & Lexeme & 0.534 & 0.949 & 0.851 & 0.823 & 0.467 \\
\addlinespace
\multirow{2}{*}{mT5-Base (French)} & Inflection & 0.851 & 0.962 & 0.975 & 0.967 & 0.969 \\
 & Lexeme & 0.654 & 0.785 & 0.698 & 0.633 & 0.665 \\
\bottomrule
\end{tabular}%
}
\end{subtable}
\hfill
\begin{subtable}{0.48\textwidth}
\centering
\caption{Selectivity (MLP Probes)}
\resizebox{\textwidth}{!}{%
\begin{tabular}{llccccc}
\toprule
& & \multicolumn{5}{c}{\textbf{Relative Depth (\%)}} \\
\cmidrule(lr){3-7}
\textbf{Model} & \textbf{Task} & \textbf{0\%} & \textbf{25\%} & \textbf{50\%} & \textbf{75\%} & \textbf{100\%} \\
\midrule
\multirow{2}{*}{Goldfish French (French)} & Inflection & 0.131 & 0.289 & 0.372 & 0.443 & 0.486 \\
 & Lexeme & -0.016 & 0.006 & 0.032 & 0.072 & 0.084 \\
\addlinespace
\multirow{2}{*}{Qwen2.5-1.5B (French)} & Inflection & 0.236 & 0.260 & 0.402 & 0.405 & 0.548 \\
 & Lexeme & 0.006 & 0.025 & 0.046 & 0.056 & 0.043 \\
\addlinespace
\multirow{2}{*}{Qwen2.5-1.5B-Instruct (French)} & Inflection & 0.235 & 0.263 & 0.411 & 0.410 & 0.552 \\
 & Lexeme & 0.002 & 0.030 & 0.048 & 0.055 & 0.064 \\
\addlinespace
\multirow{2}{*}{Qwen2.5-7B (French)} & Inflection & 0.227 & 0.198 & 0.394 & 0.415 & 0.516 \\
 & Lexeme & -0.001 & 0.025 & 0.065 & 0.070 & 0.027 \\
\addlinespace
\multirow{2}{*}{Qwen2.5-7B-Instruct (French)} & Inflection & 0.229 & 0.205 & 0.407 & 0.426 & 0.514 \\
 & Lexeme & 0.000 & 0.028 & 0.072 & 0.070 & 0.037 \\
\addlinespace
\multirow{2}{*}{mT5-Base (French)} & Inflection & 0.127 & 0.190 & 0.321 & 0.404 & 0.388 \\
 & Lexeme & -0.020 & -0.036 & -0.015 & -0.000 & -0.000 \\
\bottomrule
\end{tabular}%
}
\end{subtable}
\end{table*}

\begin{table*}[h!]
\centering
\caption{Probing Results for German}
\label{tab:probes_german}
\begin{subtable}{0.48\textwidth}
\centering
\caption{Accuracy (Linear Probes)}
\resizebox{\textwidth}{!}{%
\begin{tabular}{llccccc}
\toprule
& & \multicolumn{5}{c}{\textbf{Relative Depth (\%)}} \\
\cmidrule(lr){3-7}
\textbf{Model} & \textbf{Task} & \textbf{0\%} & \textbf{25\%} & \textbf{50\%} & \textbf{75\%} & \textbf{100\%} \\
\midrule
\multirow{2}{*}{Goldfish German (German)} & Inflection & 0.911 & 0.946 & 0.961 & 0.961 & 0.952 \\
 & Lexeme & 0.886 & 0.831 & 0.707 & 0.627 & 0.569 \\
\addlinespace
\multirow{2}{*}{Qwen2.5-1.5B (German)} & Inflection & 0.744 & 0.929 & 0.931 & 0.930 & 0.911 \\
 & Lexeme & 0.479 & 0.865 & 0.707 & 0.690 & 0.569 \\
\addlinespace
\multirow{2}{*}{Qwen2.5-1.5B-Instruct (German)} & Inflection & 0.745 & 0.928 & 0.929 & 0.929 & 0.912 \\
 & Lexeme & 0.479 & 0.862 & 0.694 & 0.686 & 0.582 \\
\addlinespace
\multirow{2}{*}{Qwen2.5-7B (German)} & Inflection & 0.744 & 0.949 & 0.946 & 0.950 & 0.935 \\
 & Lexeme & 0.480 & 0.942 & 0.811 & 0.764 & 0.651 \\
\addlinespace
\multirow{2}{*}{Qwen2.5-7B-Instruct (German)} & Inflection & 0.760 & 0.958 & 0.954 & 0.958 & 0.938 \\
 & Lexeme & 0.480 & 0.943 & 0.801 & 0.757 & 0.646 \\
\addlinespace
\multirow{2}{*}{mT5-Base (German)} & Inflection & 0.811 & 0.942 & 0.954 & 0.956 & 0.916 \\
 & Lexeme & 0.650 & 0.796 & 0.656 & 0.574 & 0.543 \\
\bottomrule
\end{tabular}%
}
\end{subtable}
\hfill
\begin{subtable}{0.48\textwidth}
\centering
\caption{Selectivity (Linear Probes)}
\resizebox{\textwidth}{!}{%
\begin{tabular}{llccccc}
\toprule
& & \multicolumn{5}{c}{\textbf{Relative Depth (\%)}} \\
\cmidrule(lr){3-7}
\textbf{Model} & \textbf{Task} & \textbf{0\%} & \textbf{25\%} & \textbf{50\%} & \textbf{75\%} & \textbf{100\%} \\
\midrule
\multirow{2}{*}{Goldfish German (German)} & Inflection & 0.418 & 0.503 & 0.593 & 0.622 & 0.645 \\
 & Lexeme & -0.001 & 0.057 & 0.088 & 0.088 & 0.069 \\
\addlinespace
\multirow{2}{*}{Qwen2.5-1.5B (German)} & Inflection & 0.289 & 0.446 & 0.577 & 0.582 & 0.599 \\
 & Lexeme & 0.009 & 0.084 & 0.082 & 0.096 & 0.065 \\
\addlinespace
\multirow{2}{*}{Qwen2.5-1.5B-Instruct (German)} & Inflection & 0.289 & 0.450 & 0.577 & 0.583 & 0.602 \\
 & Lexeme & 0.009 & 0.082 & 0.082 & 0.098 & 0.069 \\
\addlinespace
\multirow{2}{*}{Qwen2.5-7B (German)} & Inflection & 0.286 & 0.357 & 0.546 & 0.574 & 0.597 \\
 & Lexeme & 0.009 & 0.065 & 0.104 & 0.113 & 0.085 \\
\addlinespace
\multirow{2}{*}{Qwen2.5-7B-Instruct (German)} & Inflection & 0.228 & 0.210 & 0.468 & 0.509 & 0.530 \\
 & Lexeme & 0.009 & 0.067 & 0.105 & 0.111 & 0.084 \\
\addlinespace
\multirow{2}{*}{mT5-Base (German)} & Inflection & 0.251 & 0.371 & 0.494 & 0.560 & 0.529 \\
 & Lexeme & 0.000 & 0.012 & 0.023 & 0.043 & 0.033 \\
\bottomrule
\end{tabular}%
}
\end{subtable}

\vspace{0.75em}

\begin{subtable}{0.48\textwidth}
\centering
\caption{Accuracy (MLP Probes)}
\resizebox{\textwidth}{!}{%
\begin{tabular}{llccccc}
\toprule
& & \multicolumn{5}{c}{\textbf{Relative Depth (\%)}} \\
\cmidrule(lr){3-7}
\textbf{Model} & \textbf{Task} & \textbf{0\%} & \textbf{25\%} & \textbf{50\%} & \textbf{75\%} & \textbf{100\%} \\
\midrule
\multirow{2}{*}{Goldfish German (German)} & Inflection & 0.923 & 0.955 & 0.969 & 0.969 & 0.960 \\
 & Lexeme & 0.902 & 0.876 & 0.794 & 0.657 & 0.511 \\
\addlinespace
\multirow{2}{*}{Qwen2.5-1.5B (German)} & Inflection & 0.741 & 0.943 & 0.944 & 0.942 & 0.910 \\
 & Lexeme & 0.473 & 0.869 & 0.763 & 0.682 & 0.292 \\
\addlinespace
\multirow{2}{*}{Qwen2.5-1.5B-Instruct (German)} & Inflection & 0.741 & 0.943 & 0.943 & 0.941 & 0.910 \\
 & Lexeme & 0.474 & 0.870 & 0.753 & 0.688 & 0.300 \\
\addlinespace
\multirow{2}{*}{Qwen2.5-7B (German)} & Inflection & 0.740 & 0.956 & 0.954 & 0.956 & 0.935 \\
 & Lexeme & 0.471 & 0.943 & 0.820 & 0.746 & 0.383 \\
\addlinespace
\multirow{2}{*}{Qwen2.5-7B-Instruct (German)} & Inflection & 0.758 & 0.962 & 0.959 & 0.961 & 0.935 \\
 & Lexeme & 0.471 & 0.943 & 0.810 & 0.749 & 0.397 \\
\addlinespace
\multirow{2}{*}{mT5-Base (German)} & Inflection & 0.820 & 0.956 & 0.954 & 0.952 & 0.939 \\
 & Lexeme & 0.641 & 0.710 & 0.563 & 0.486 & 0.530 \\
\bottomrule
\end{tabular}%
}
\end{subtable}
\hfill
\begin{subtable}{0.48\textwidth}
\centering
\caption{Selectivity (MLP Probes)}
\resizebox{\textwidth}{!}{%
\begin{tabular}{llccccc}
\toprule
& & \multicolumn{5}{c}{\textbf{Relative Depth (\%)}} \\
\cmidrule(lr){3-7}
\textbf{Model} & \textbf{Task} & \textbf{0\%} & \textbf{25\%} & \textbf{50\%} & \textbf{75\%} & \textbf{100\%} \\
\midrule
\multirow{2}{*}{Goldfish German (German)} & Inflection & 0.229 & 0.350 & 0.457 & 0.523 & 0.585 \\
 & Lexeme & -0.021 & 0.017 & 0.051 & 0.084 & 0.082 \\
\addlinespace
\multirow{2}{*}{Qwen2.5-1.5B (German)} & Inflection & 0.292 & 0.292 & 0.457 & 0.492 & 0.604 \\
 & Lexeme & 0.008 & 0.022 & 0.054 & 0.060 & 0.035 \\
\addlinespace
\multirow{2}{*}{Qwen2.5-1.5B-Instruct (German)} & Inflection & 0.291 & 0.295 & 0.472 & 0.488 & 0.600 \\
 & Lexeme & 0.009 & 0.023 & 0.056 & 0.063 & 0.032 \\
\addlinespace
\multirow{2}{*}{Qwen2.5-7B (German)} & Inflection & 0.288 & 0.214 & 0.455 & 0.493 & 0.596 \\
 & Lexeme & 0.007 & 0.011 & 0.066 & 0.070 & 0.046 \\
\addlinespace
\multirow{2}{*}{Qwen2.5-7B-Instruct (German)} & Inflection & 0.227 & 0.111 & 0.382 & 0.424 & 0.532 \\
 & Lexeme & 0.008 & 0.015 & 0.073 & 0.077 & 0.049 \\
\addlinespace
\multirow{2}{*}{mT5-Base (German)} & Inflection & 0.160 & 0.253 & 0.405 & 0.475 & 0.462 \\
 & Lexeme & -0.012 & -0.037 & 0.000 & 0.023 & 0.012 \\
\bottomrule
\end{tabular}%
}
\end{subtable}
\end{table*}

\begin{table*}[h!]
\centering
\caption{Probing Results for Russian}
\label{tab:probes_russian}
\begin{subtable}{0.48\textwidth}
\centering
\caption{Accuracy (Linear Probes)}
\resizebox{\textwidth}{!}{%
\begin{tabular}{llccccc}
\toprule
& & \multicolumn{5}{c}{\textbf{Relative Depth (\%)}} \\
\cmidrule(lr){3-7}
\textbf{Model} & \textbf{Task} & \textbf{0\%} & \textbf{25\%} & \textbf{50\%} & \textbf{75\%} & \textbf{100\%} \\
\midrule
\multirow{2}{*}{Goldfish Russian (Russian)} & Inflection & 0.932 & 0.952 & 0.975 & 0.968 & 0.950 \\
 & Lexeme & 0.896 & 0.854 & 0.758 & 0.710 & 0.631 \\
\addlinespace
\multirow{2}{*}{Qwen2.5-1.5B (Russian)} & Inflection & 0.850 & 0.966 & 0.966 & 0.962 & 0.933 \\
 & Lexeme & 0.315 & 0.896 & 0.739 & 0.720 & 0.598 \\
\addlinespace
\multirow{2}{*}{Qwen2.5-1.5B-Instruct (Russian)} & Inflection & 0.850 & 0.965 & 0.964 & 0.960 & 0.932 \\
 & Lexeme & 0.315 & 0.893 & 0.725 & 0.714 & 0.600 \\
\addlinespace
\multirow{2}{*}{Qwen2.5-7B (Russian)} & Inflection & 0.850 & 0.977 & 0.976 & 0.974 & 0.954 \\
 & Lexeme & 0.315 & 0.960 & 0.834 & 0.798 & 0.696 \\
\addlinespace
\multirow{2}{*}{Qwen2.5-7B-Instruct (Russian)} & Inflection & 0.858 & 0.977 & 0.974 & 0.980 & 0.953 \\
 & Lexeme & 0.315 & 0.959 & 0.821 & 0.785 & 0.680 \\
\addlinespace
\multirow{2}{*}{mT5-Base (Russian)} & Inflection & 0.882 & 0.944 & 0.974 & 0.971 & 0.952 \\
 & Lexeme & 0.480 & 0.766 & 0.666 & 0.570 & 0.515 \\
\bottomrule
\end{tabular}%
}
\end{subtable}
\hfill
\begin{subtable}{0.48\textwidth}
\centering
\caption{Selectivity (Linear Probes)}
\resizebox{\textwidth}{!}{%
\begin{tabular}{llccccc}
\toprule
& & \multicolumn{5}{c}{\textbf{Relative Depth (\%)}} \\
\cmidrule(lr){3-7}
\textbf{Model} & \textbf{Task} & \textbf{0\%} & \textbf{25\%} & \textbf{50\%} & \textbf{75\%} & \textbf{100\%} \\
\midrule
\multirow{2}{*}{Goldfish Russian (Russian)} & Inflection & 0.505 & 0.568 & 0.663 & 0.675 & 0.688 \\
 & Lexeme & 0.024 & 0.115 & 0.157 & 0.175 & 0.170 \\
\addlinespace
\multirow{2}{*}{Qwen2.5-1.5B (Russian)} & Inflection & 0.518 & 0.576 & 0.679 & 0.661 & 0.665 \\
 & Lexeme & 0.012 & 0.190 & 0.187 & 0.202 & 0.153 \\
\addlinespace
\multirow{2}{*}{Qwen2.5-1.5B-Instruct (Russian)} & Inflection & 0.518 & 0.582 & 0.685 & 0.664 & 0.666 \\
 & Lexeme & 0.012 & 0.191 & 0.190 & 0.205 & 0.152 \\
\addlinespace
\multirow{2}{*}{Qwen2.5-7B (Russian)} & Inflection & 0.517 & 0.504 & 0.670 & 0.671 & 0.658 \\
 & Lexeme & 0.011 & 0.165 & 0.218 & 0.222 & 0.183 \\
\addlinespace
\multirow{2}{*}{Qwen2.5-7B-Instruct (Russian)} & Inflection & 0.431 & 0.332 & 0.581 & 0.594 & 0.593 \\
 & Lexeme & 0.011 & 0.167 & 0.221 & 0.222 & 0.181 \\
\addlinespace
\multirow{2}{*}{mT5-Base (Russian)} & Inflection & 0.388 & 0.418 & 0.548 & 0.595 & 0.605 \\
 & Lexeme & 0.004 & 0.088 & 0.092 & 0.099 & 0.092 \\
\bottomrule
\end{tabular}%
}
\end{subtable}

\vspace{0.75em}

\begin{subtable}{0.48\textwidth}
\centering
\caption{Accuracy (MLP Probes)}
\resizebox{\textwidth}{!}{%
\begin{tabular}{llccccc}
\toprule
& & \multicolumn{5}{c}{\textbf{Relative Depth (\%)}} \\
\cmidrule(lr){3-7}
\textbf{Model} & \textbf{Task} & \textbf{0\%} & \textbf{25\%} & \textbf{50\%} & \textbf{75\%} & \textbf{100\%} \\
\midrule
\multirow{2}{*}{Goldfish Russian (Russian)} & Inflection & 0.944 & 0.966 & 0.982 & 0.977 & 0.959 \\
 & Lexeme & 0.896 & 0.878 & 0.814 & 0.732 & 0.582 \\
\addlinespace
\multirow{2}{*}{Qwen2.5-1.5B (Russian)} & Inflection & 0.848 & 0.972 & 0.971 & 0.969 & 0.924 \\
 & Lexeme & 0.302 & 0.874 & 0.768 & 0.690 & 0.281 \\
\addlinespace
\multirow{2}{*}{Qwen2.5-1.5B-Instruct (Russian)} & Inflection & 0.848 & 0.971 & 0.970 & 0.967 & 0.927 \\
 & Lexeme & 0.302 & 0.870 & 0.760 & 0.692 & 0.282 \\
\addlinespace
\multirow{2}{*}{Qwen2.5-7B (Russian)} & Inflection & 0.850 & 0.981 & 0.981 & 0.977 & 0.940 \\
 & Lexeme & 0.312 & 0.960 & 0.838 & 0.770 & 0.361 \\
\addlinespace
\multirow{2}{*}{Qwen2.5-7B-Instruct (Russian)} & Inflection & 0.857 & 0.976 & 0.976 & 0.978 & 0.935 \\
 & Lexeme & 0.308 & 0.958 & 0.828 & 0.765 & 0.371 \\
\addlinespace
\multirow{2}{*}{mT5-Base (Russian)} & Inflection & 0.883 & 0.956 & 0.978 & 0.974 & 0.971 \\
 & Lexeme & 0.448 & 0.571 & 0.470 & 0.397 & 0.437 \\
\bottomrule
\end{tabular}%
}
\end{subtable}
\hfill
\begin{subtable}{0.48\textwidth}
\centering
\caption{Selectivity (MLP Probes)}
\resizebox{\textwidth}{!}{%
\begin{tabular}{llccccc}
\toprule
& & \multicolumn{5}{c}{\textbf{Relative Depth (\%)}} \\
\cmidrule(lr){3-7}
\textbf{Model} & \textbf{Task} & \textbf{0\%} & \textbf{25\%} & \textbf{50\%} & \textbf{75\%} & \textbf{100\%} \\
\midrule
\multirow{2}{*}{Goldfish Russian (Russian)} & Inflection & 0.322 & 0.466 & 0.557 & 0.604 & 0.642 \\
 & Lexeme & -0.007 & 0.049 & 0.120 & 0.203 & 0.201 \\
\addlinespace
\multirow{2}{*}{Qwen2.5-1.5B (Russian)} & Inflection & 0.524 & 0.451 & 0.575 & 0.594 & 0.662 \\
 & Lexeme & 0.004 & 0.080 & 0.122 & 0.162 & 0.108 \\
\addlinespace
\multirow{2}{*}{Qwen2.5-1.5B-Instruct (Russian)} & Inflection & 0.523 & 0.457 & 0.589 & 0.594 & 0.660 \\
 & Lexeme & 0.004 & 0.085 & 0.133 & 0.166 & 0.101 \\
\addlinespace
\multirow{2}{*}{Qwen2.5-7B (Russian)} & Inflection & 0.526 & 0.376 & 0.591 & 0.604 & 0.652 \\
 & Lexeme & 0.011 & 0.062 & 0.175 & 0.183 & 0.123 \\
\addlinespace
\multirow{2}{*}{Qwen2.5-7B-Instruct (Russian)} & Inflection & 0.432 & 0.211 & 0.493 & 0.512 & 0.583 \\
 & Lexeme & 0.008 & 0.065 & 0.184 & 0.193 & 0.125 \\
\addlinespace
\multirow{2}{*}{mT5-Base (Russian)} & Inflection & 0.351 & 0.335 & 0.497 & 0.547 & 0.557 \\
 & Lexeme & -0.016 & -0.016 & 0.023 & 0.035 & 0.042 \\
\bottomrule
\end{tabular}%
}
\end{subtable}
\end{table*}

\begin{table*}[h!]
\centering
\caption{Probing Results for Turkish}
\label{tab:probes_turkish}
\begin{subtable}{0.48\textwidth}
\centering
\caption{Accuracy (Linear Probes)}
\resizebox{\textwidth}{!}{%
\begin{tabular}{llccccc}
\toprule
& & \multicolumn{5}{c}{\textbf{Relative Depth (\%)}} \\
\cmidrule(lr){3-7}
\textbf{Model} & \textbf{Task} & \textbf{0\%} & \textbf{25\%} & \textbf{50\%} & \textbf{75\%} & \textbf{100\%} \\
\midrule
\multirow{2}{*}{Goldfish Turkish (Turkish)} & Inflection & 0.907 & 0.930 & 0.925 & 0.913 & 0.903 \\
 & Lexeme & 0.978 & 0.973 & 0.968 & 0.921 & 0.614 \\
\addlinespace
\multirow{2}{*}{Qwen2.5-1.5B (Turkish)} & Inflection & 0.719 & 0.869 & 0.847 & 0.849 & 0.831 \\
 & Lexeme & 0.530 & 0.959 & 0.868 & 0.815 & 0.796 \\
\addlinespace
\multirow{2}{*}{Qwen2.5-1.5B-Instruct (Turkish)} & Inflection & 0.719 & 0.869 & 0.838 & 0.846 & 0.827 \\
 & Lexeme & 0.530 & 0.961 & 0.852 & 0.804 & 0.786 \\
\addlinespace
\multirow{2}{*}{Qwen2.5-7B (Turkish)} & Inflection & 0.718 & 0.917 & 0.889 & 0.879 & 0.839 \\
 & Lexeme & 0.531 & 0.974 & 0.878 & 0.839 & 0.777 \\
\addlinespace
\multirow{2}{*}{Qwen2.5-7B-Instruct (Turkish)} & Inflection & 0.718 & 0.911 & 0.875 & 0.874 & 0.836 \\
 & Lexeme & 0.531 & 0.975 & 0.854 & 0.803 & 0.731 \\
\addlinespace
\multirow{2}{*}{mT5-Base (Turkish)} & Inflection & 0.913 & 0.972 & 0.931 & 0.908 & 0.884 \\
 & Lexeme & 0.792 & 0.952 & 0.922 & 0.819 & 0.785 \\
\bottomrule
\end{tabular}%
}
\end{subtable}
\hfill
\begin{subtable}{0.48\textwidth}
\centering
\caption{Selectivity (Linear Probes)}
\resizebox{\textwidth}{!}{%
\begin{tabular}{llccccc}
\toprule
& & \multicolumn{5}{c}{\textbf{Relative Depth (\%)}} \\
\cmidrule(lr){3-7}
\textbf{Model} & \textbf{Task} & \textbf{0\%} & \textbf{25\%} & \textbf{50\%} & \textbf{75\%} & \textbf{100\%} \\
\midrule
\multirow{2}{*}{Goldfish Turkish (Turkish)} & Inflection & 0.345 & 0.407 & 0.491 & 0.492 & 0.500 \\
 & Lexeme & -0.002 & 0.008 & 0.074 & 0.089 & 0.144 \\
\addlinespace
\multirow{2}{*}{Qwen2.5-1.5B (Turkish)} & Inflection & 0.277 & 0.323 & 0.452 & 0.455 & 0.411 \\
 & Lexeme & 0.013 & 0.001 & 0.000 & -0.014 & 0.008 \\
\addlinespace
\multirow{2}{*}{Qwen2.5-1.5B-Instruct (Turkish)} & Inflection & 0.277 & 0.319 & 0.447 & 0.452 & 0.416 \\
 & Lexeme & 0.013 & 0.003 & 0.004 & -0.009 & 0.009 \\
\addlinespace
\multirow{2}{*}{Qwen2.5-7B (Turkish)} & Inflection & 0.275 & 0.293 & 0.472 & 0.462 & 0.417 \\
 & Lexeme & 0.014 & -0.002 & 0.003 & -0.010 & -0.021 \\
\addlinespace
\multirow{2}{*}{Qwen2.5-7B-Instruct (Turkish)} & Inflection & 0.276 & 0.291 & 0.463 & 0.497 & 0.445 \\
 & Lexeme & 0.014 & 0.003 & -0.003 & -0.017 & -0.037 \\
\addlinespace
\multirow{2}{*}{mT5-Base (Turkish)} & Inflection & 0.071 & 0.165 & 0.261 & 0.331 & 0.366 \\
 & Lexeme & 0.008 & 0.013 & 0.042 & 0.035 & 0.057 \\
\bottomrule
\end{tabular}%
}
\end{subtable}

\vspace{0.75em}

\begin{subtable}{0.48\textwidth}
\centering
\caption{Accuracy (MLP Probes)}
\resizebox{\textwidth}{!}{%
\begin{tabular}{llccccc}
\toprule
& & \multicolumn{5}{c}{\textbf{Relative Depth (\%)}} \\
\cmidrule(lr){3-7}
\textbf{Model} & \textbf{Task} & \textbf{0\%} & \textbf{25\%} & \textbf{50\%} & \textbf{75\%} & \textbf{100\%} \\
\midrule
\multirow{2}{*}{Goldfish Turkish (Turkish)} & Inflection & 0.911 & 0.918 & 0.916 & 0.899 & 0.887 \\
 & Lexeme & 0.601 & 0.536 & 0.459 & 0.418 & 0.360 \\
\addlinespace
\multirow{2}{*}{Qwen2.5-1.5B (Turkish)} & Inflection & 0.713 & 0.854 & 0.823 & 0.829 & 0.760 \\
 & Lexeme & 0.459 & 0.500 & 0.369 & 0.325 & 0.232 \\
\addlinespace
\multirow{2}{*}{Qwen2.5-1.5B-Instruct (Turkish)} & Inflection & 0.712 & 0.857 & 0.831 & 0.817 & 0.760 \\
 & Lexeme & 0.462 & 0.491 & 0.367 & 0.333 & 0.233 \\
\addlinespace
\multirow{2}{*}{Qwen2.5-7B (Turkish)} & Inflection & 0.713 & 0.923 & 0.902 & 0.900 & 0.828 \\
 & Lexeme & 0.519 & 0.805 & 0.639 & 0.525 & 0.441 \\
\addlinespace
\multirow{2}{*}{Qwen2.5-7B-Instruct (Turkish)} & Inflection & 0.717 & 0.914 & 0.900 & 0.895 & 0.820 \\
 & Lexeme & 0.521 & 0.797 & 0.615 & 0.523 & 0.383 \\
\addlinespace
\multirow{2}{*}{mT5-Base (Turkish)} & Inflection & 0.884 & 0.915 & 0.875 & 0.836 & 0.839 \\
 & Lexeme & 0.503 & 0.395 & 0.300 & 0.257 & 0.312 \\
\bottomrule
\end{tabular}%
}
\end{subtable}
\hfill
\begin{subtable}{0.48\textwidth}
\centering
\caption{Selectivity (MLP Probes)}
\resizebox{\textwidth}{!}{%
\begin{tabular}{llccccc}
\toprule
& & \multicolumn{5}{c}{\textbf{Relative Depth (\%)}} \\
\cmidrule(lr){3-7}
\textbf{Model} & \textbf{Task} & \textbf{0\%} & \textbf{25\%} & \textbf{50\%} & \textbf{75\%} & \textbf{100\%} \\
\midrule
\multirow{2}{*}{Goldfish Turkish (Turkish)} & Inflection & 0.333 & 0.449 & 0.499 & 0.488 & 0.496 \\
 & Lexeme & 0.175 & 0.185 & 0.142 & 0.138 & 0.125 \\
\addlinespace
\multirow{2}{*}{Qwen2.5-1.5B (Turkish)} & Inflection & 0.299 & 0.355 & 0.438 & 0.500 & 0.432 \\
 & Lexeme & -0.001 & 0.045 & 0.093 & 0.049 & 0.045 \\
\addlinespace
\multirow{2}{*}{Qwen2.5-1.5B-Instruct (Turkish)} & Inflection & 0.303 & 0.350 & 0.447 & 0.447 & 0.432 \\
 & Lexeme & 0.001 & 0.042 & 0.104 & 0.065 & -0.078 \\
\addlinespace
\multirow{2}{*}{Qwen2.5-7B (Turkish)} & Inflection & 0.297 & 0.299 & 0.422 & 0.473 & 0.442 \\
 & Lexeme & 0.007 & 0.084 & 0.128 & 0.105 & 0.065 \\
\addlinespace
\multirow{2}{*}{Qwen2.5-7B-Instruct (Turkish)} & Inflection & 0.303 & 0.297 & 0.437 & 0.492 & 0.432 \\
 & Lexeme & 0.012 & 0.087 & 0.147 & 0.103 & 0.061 \\
\addlinespace
\multirow{2}{*}{mT5-Base (Turkish)} & Inflection & 0.120 & 0.301 & 0.343 & 0.355 & 0.368 \\
 & Lexeme & 0.047 & 0.094 & 0.109 & 0.097 & 0.106 \\
\bottomrule
\end{tabular}%
}
\end{subtable}
\end{table*}


\section{Dataset Statistics}
\label{sec:dataset_appendix}
This section provides statistics and visualizations for the datasets and models used in our experiments across all six languages. Only words containing alphabetic characters and apostrophes were considered.

\begin{table*}[htbp]
\resizebox{\linewidth}{!}{%
    \centering
    \begin{tabular}{lrrrrrr}
        \hline
        \textbf{Language} & \textbf{Total Words} & \textbf{Unique Lemmas} & \textbf{Unique Forms} & \textbf{Inflection Types} & \textbf{Sentences} & \textbf{Avg. Length} \\
        \hline
        English & 54,816 & 7,848 & 11,720 & 8 & 8,415 & 6.5 \\
        Chinese & 44,166 & 11,184 & 11,237 & 4 & 7,892 & 5.8 \\
        German & 84,710 & 24,140 & 31,890 & 9 & 9,234 & 7.3 \\
        French & 115,847 & 13,804 & 24,485 & 6 & 8,765 & 6.6 \\
        Russian & 193,320 & 20,943 & 59,830 & 8 & 10,234 & 7.1 \\
        Turkish & 20,881 & 3,776 & 11,680 & 7 & 6,789 & 6.4 \\
        \hline
    \end{tabular}
}%
\caption{Dataset statistics across all six languages. Russian has the largest dataset and the highest number of unique forms, reflecting its rich inflectional morphology. Turkish has the fewest total words and lemmas, while Chinese has the fewest inflection types.}
\label{tab:multilingual_dataset_stats}
\end{table*}

\subsection{English Dataset Details}

For the English GUM corpus specifically, the data covers three main syntactic categories: nouns (49.5\%), verbs (31.2\%), and adjectives (19.4\%). 

\cref{tab:category_distribution_english} shows the distribution of word categories in the English dataset, and \cref{tab:inflection_distribution_english} presents the distribution of inflection categories.

\begin{table*}[htbp]
  \small
  \centering
  \begin{subtable}[t]{0.30\textwidth}
    \centering
    \begin{tabular}{lrr}
      \toprule
      \textbf{Category} & \textbf{Count} & \textbf{\%} \\
      \midrule
      Noun & 27111 & 49.5 \\
      Verb & 17093 & 31.2 \\
      Adjective & 10612 & 19.4 \\
      \bottomrule
    \end{tabular}
    \caption{Word categories}
    \label{tab:category_distribution_english}
  \end{subtable}
  \hfill
  \begin{subtable}[t]{0.35\textwidth}
    \centering
    \begin{tabular}{lrr}
      \toprule
      \textbf{Inflection} & \textbf{Count} & \textbf{\%} \\
      \midrule
      Singular & 19830 & 36.2 \\
      Base & 10076 & 18.4 \\
      Positive & 9926 & 18.1 \\
      Plural & 7281 & 13.3 \\
      Past & 5604 & 10.2 \\
      3rd Person & 1413 & 2.6 \\
      Comparative & 403 & 0.7 \\
      Superlative & 283 & 0.5 \\
      \bottomrule
    \end{tabular}
    \caption{Inflection categories}
    \label{tab:inflection_distribution_english}
  \end{subtable}
  \hfill
  \begin{subtable}[t]{0.30\textwidth}
    \centering
    \begin{tabular}{lr}
       \toprule
       \textbf{Metric} & \textbf{Value} \\
       \midrule
       Avg. Words & 6.5 \\
       Median Words & 5 \\
       Min. Words & 1 \\
       Max. Words & 40 \\
       \bottomrule
     \end{tabular}
     \caption{Sentence length stats}
     \label{tab:sentence_stats_english}
  \end{subtable}
  
  \caption{Distribution statistics for the English dataset. Table (a) shows syntactic categories, (b) details inflection types, and (c) provides sentence length heuristics.}
  \label{fig:english_distributions}
\end{table*}

\subsection{Tokenization Statistics}

\begin{table*}[htbp]
  \centering
  \begin{tabular}{ll}
    \toprule
    \textbf{Model} & \textbf{Tokenizer Type} \\
    \midrule
    \texttt{BERT Base/Large} & WordPiece \\
    \texttt{DeBERTa V3 Large} & SentencePiece \\
    \texttt{GPT-2 variants} & BPE \\
    \texttt{Pythia variants} & BPE \\
    \texttt{OLMo 2 variants} & BPE (\texttt{tiktoken}) \\
    \texttt{Gemma 2 variants} & SentencePiece \\
    \texttt{Qwen 2.5 variants} & Byte-level BPE \\
    \texttt{Llama 3.1 variants} & BPE (\texttt{tiktoken}) \\
    \bottomrule
  \end{tabular}
  \caption{Tokenization strategies used by different model families. BPE means byte-pair encoding.}
  \label{tab:tokenizers}
\end{table*}

An important consideration for our analysis is how different models tokenize the words in our dataset.~\cref{tab:tokenization_stats} shows tokenization statistics across the models we analyze. Encoder-only models like \texttt{BERT} and \texttt{DeBERTa} tend to split words into more tokens than decoder-only models like \texttt{GPT-2} and \texttt{Qwen2}, which may affect how information is encoded across layers.

\begin{table*}[htbp]
    \small
    \centering
    \begin{tabular}{lcccc}
        \hline
        \multirow{2}{*}{\textbf{Model}}  & Avg.\ tokens &Med.\ tokens & Max\ tokens & Percent multitoken \\
         & per word & per word & per word & \\
        \hline
        \texttt{BERT variants} & 1.11 & 1.0 & 6.0 & 6.95 \\
        \texttt{DeBERTa-v3-large} & 1.03 & 1.0 & 4.0 & 2.2 \\
        \texttt{GPT-2 variants} & 1.52 & 1.0 & 5.0 & 42.25 \\
        \texttt{Pythia-6.9B variants} & 1.48 & 1.0 & 5.0 & 39.1 \\
        \texttt{OLMo2-7B variants} & 1.43 & 1.0 & 4.0 & 35.9 \\
        \texttt{Gemma2-2B variants} & 1.19 & 1.0 & 4.0 & 16.55 \\
        \texttt{Qwen2.5-1.5B variants} & 1.43 & 1.0 & 4.0 & 35.9 \\
        \texttt{Llama-3.1-8B variants} & 1.43 & 1.0 & 4.0 & 35.85 \\
        \hline
    \end{tabular}
    \caption{Tokenization statistics across different models (English only). Most models have an average of 1.0-1.5 tokens per word and a median of 1, indicating that most words are tokenized as a single unit. However, there is variation in the proportion of words split into multiple tokens. Decoder-only models (\eg \texttt{GPT-2}, \texttt{Pythia}, \texttt{Qwen2}, \texttt{LLaMA}) split 35-42\% of words, while BERT and DeBERTa variants split fewer words (2-7\%). Maximum tokens per word range from 4 to 6 across all models.}
    \label{tab:tokenization_stats}
\end{table*}

\subsection{Effects of Tokenization}
\label{sec:tokenization_appendix}

\begin{figure}[htbp]
  \centering
    \includegraphics[width=\linewidth]{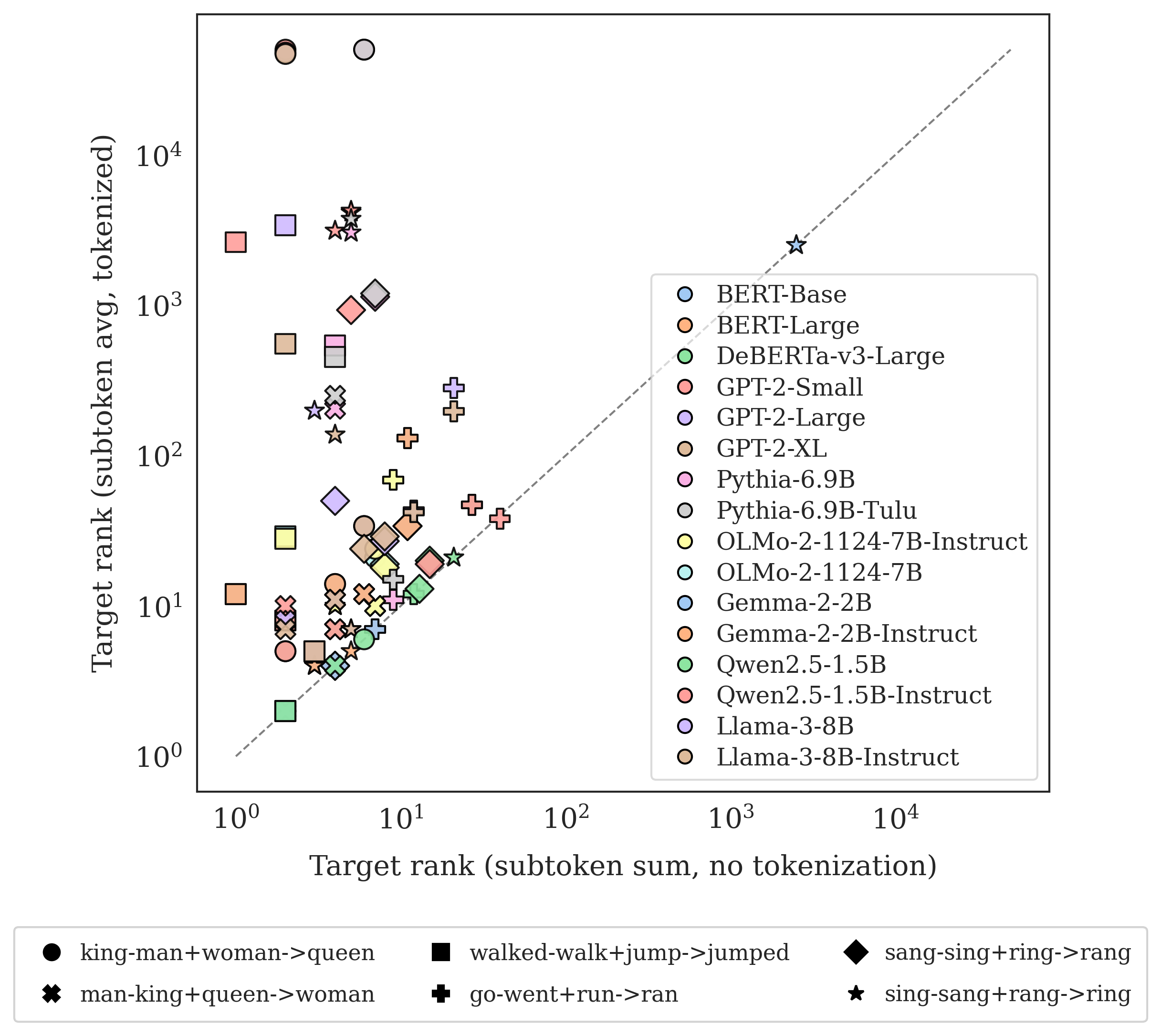}
    \caption{Effect of tokenization strategy on analogy completion rank. Each point corresponds to a model (color) and analogy (shape). The x-axis is the rank using whole-word representations. The y-axis is the rank using tokenized representations. Here, rank means the position of the expected word when all vocabulary words are sorted by similarity to the resulting embedding from vector arithmetic; lower is better. Points above the gray $y{=}x$ line mean tokenization hurts performance.}
  \label{fig:tokenization_scatter}
\end{figure}

Tokenization is an essential component of language modeling. To test how tokenization influences our findings, we examine whether tokenizing versus not affects the encoding of linguistic information. We measure this with analogy completion tasks (\eg \textit{man:king::woman:?}) using the embedding layer. When a word is split into subtokens, we compare two approaches to building representations - averaging versus summing those subtoken embeddings.

For each approach, we perform vector arithmetic on word representations (\eg \textit{king - man + woman}). We measure performance by ranking all vocabulary words by cosine similarity to the resulting representation, and observe how highly the expected word (\eg \textit{queen}) ranks, with a lower rank indicating better performance.

\Cref{fig:tokenization_scatter} shows that whole-word representations yield markedly better analogy performance than averaged subtokens across all models. This implies that linguistic regularities are primarily stored in whole-word embeddings rather than compositionally across subtokens. Despite these tokenization effects, our classifier results show consistent patterns across models using different tokenizers (see Table 6), indicating that the separation of lexical and morphological information is robust.

\begin{table*}[htbp]
    \centering
    \resizebox{0.6\textwidth}{!}{
    \begin{tabular}{ll}
        \hline
        \textbf{Model} & \textbf{HuggingFace ID} \\
        \hline
        \texttt{BERT-Base} & \texttt{bert-base-uncased} \\
        \texttt{BERT-Large} & \texttt{bert-large-uncased} \\
        \texttt{DeBERTa-v3-Large} & \texttt{microsoft/deberta-v3-large} \\
        \texttt{mT5-base} & \texttt{google/mt5-base} \\
        \texttt{GPT-2-Small} & \texttt{openai-community/gpt2} \\
        \texttt{GPT-2-Large} & \texttt{openai-community/gpt2-large} \\
        \texttt{GPT-2-XL} & \texttt{openai-community/gpt2-xl} \\
        \texttt{Pythia-6.9B} & \texttt{EleutherAI/pythia-6.9b} \\
        \texttt{Pythia-6.9B-Tulu} & \texttt{allenai/open-instruct-pythia-6.9b-tulu} \\
        \texttt{OLMo-2-1124-7B} & \texttt{allenai/OLMo-2-1124-7B} \\
        \texttt{OLMo-2-1124-7B-Instruct} & \texttt{allenai/OLMo-2-1124-7B-Instruct} \\
        \texttt{Gemma-2-2B} & \texttt{google/gemma-2-2b} \\
        \texttt{Gemma-2-2B-Instruct} & \texttt{google/gemma-2-2b-it} \\
        \texttt{Qwen2.5-1.5B} & \texttt{Qwen/Qwen2.5-1.5B} \\
        \texttt{Qwen2.5-1.5B-Instruct} & \texttt{Qwen/Qwen2.5-1.5B-Instruct} \\
        \texttt{Qwen2.5-7B} & \texttt{Qwen/Qwen2.5-7B} \\
        \texttt{Qwen2.5-7B-Instruct} & \texttt{Qwen/Qwen2.5-7B-Instruct} \\
        \texttt{Llama-3.1-8B} & \texttt{meta-llama/Llama-3.1-8B} \\
        \texttt{Llama-3.1-8B-Instruct} & \texttt{meta-llama/Llama-3.1-8B-Instruct} \\
        \texttt{Goldfish English} & \texttt{goldfish-models/goldfish\_eng\_latn\_1000mb} \\
        \texttt{Goldfish Chinese} & \texttt{goldfish-models/goldfish\_zho\_hans\_1000mb} \\
        \texttt{Goldfish German} & \texttt{goldfish-models/goldfish\_deu\_latn\_1000mb} \\
        \texttt{Goldfish French} & \texttt{goldfish-models/goldfish\_fra\_latn\_1000mb} \\
        \texttt{Goldfish Russian} & \texttt{goldfish-models/goldfish\_rus\_cyrl\_1000mb} \\
        \texttt{Goldfish Turkish} & \texttt{goldfish-models/goldfish\_tur\_latn\_1000mb} \\
        \hline
    \end{tabular} }
    \caption{Canonical HuggingFace model IDs used to load models in our study.}
    \label{tab:hf_id_mapping}
\end{table*}

\section{Additional Analysis}
\label{sec:additional_tables}

\subsection{Linear Effective Dimensionality Results}
\label{sec:intrinsic_dim_appendix}

This section presents detailed linear effective dimensionality analyses showing how representation compression varies across layers and between models.

\Cref{fig:intrinsic_dim_table} provides a numerical summary, reporting the number of principal components required to reach 50\%, 70\%, and 90\% explained variance at the first, middle, and final layers of each model. 

\begin{figure*}[htbp]
  \centering
  \includegraphics[width=\textwidth]{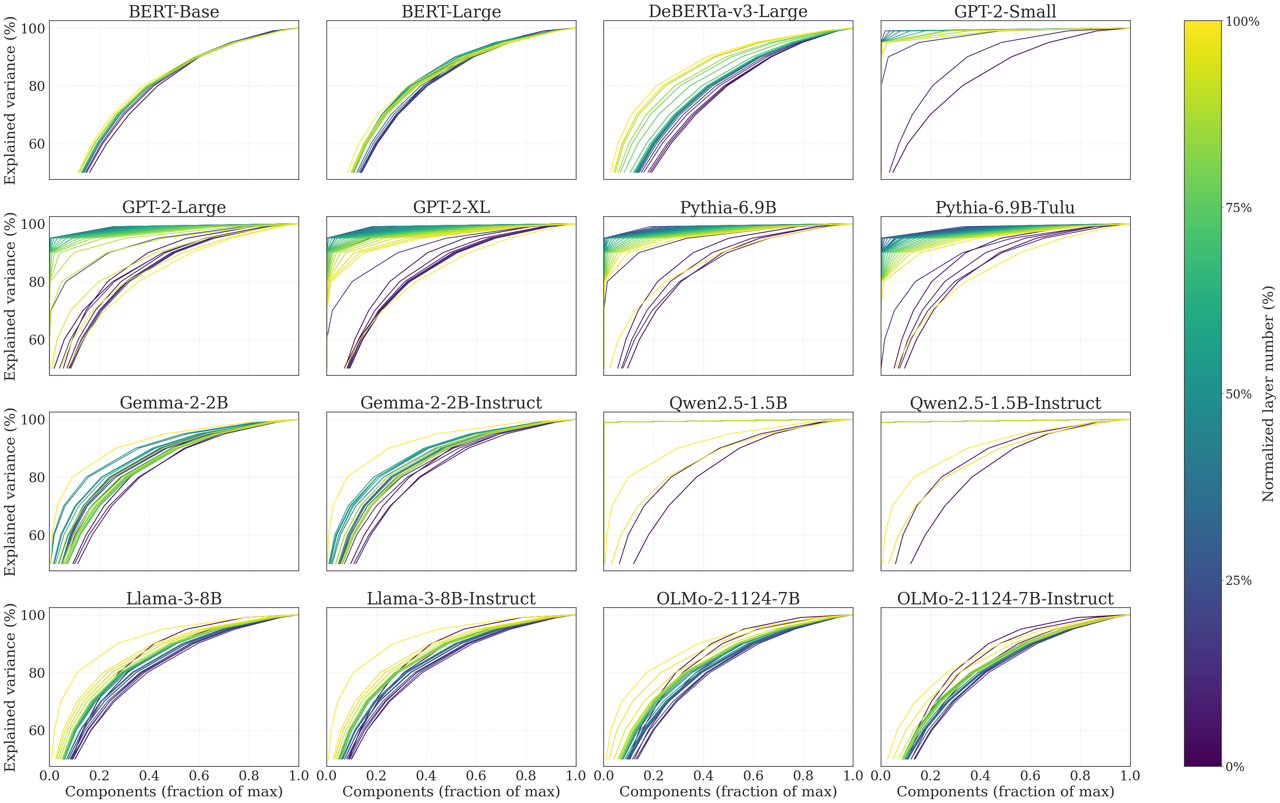}
  \caption{Linear effective dimensionality curves for all models for English. Each subplot shows the relationship between the percentage of maximum PCA components (x-axis) and the percentage of explained variance (y-axis) across different layers. The color gradient from purple (early layers, 0\%) to yellow (late layers, 100\%) indicates the relative layer depth within each model. Models like \texttt{BERT}, \texttt{Gemma}, and \texttt{Llama} show similar compression patterns, while \texttt{GPT-2} variants, \texttt{Qwen} and \texttt{Pythia} exhibit opposite trends in their middle layers.}
  \label{fig:intrinsic_dim_complete}
\end{figure*}

\Cref{fig:intrinsic_dim_complete} visualizes these relationships as curves for all models, with each subplot showing how explained variance accumulates as a function of the percentage of maximum PCA components, color-coded by relative layer depth.

\begin{table*}[htbp]
\small
\centering
\renewcommand\arraystretch{1.2}
\setlength{\tabcolsep}{9pt}
\resizebox{\linewidth}{!}{%
\begin{tabular}{@{}l c ccc ccc ccc@{}}
    \toprule
    \multirow{2}{*}{Model} & \multirow{2}{*}{$d_\text{model}$} &
      \multicolumn{3}{c}{ID$_{50}$} &
      \multicolumn{3}{c}{ID$_{70}$} &
      \multicolumn{3}{c}{ID$_{90}$} \\
    \cmidrule(lr){3-5}\cmidrule(lr){6-8}\cmidrule(lr){9-11}
      & & First & Mid & Final & First & Mid & Final & First & Mid & Final \\
    \midrule
    \texttt{BERT-Base}                & 768 & 123 & 100 & 88 & 244 & 212 & 192 & 461 & 451 & 446 \\
    \texttt{BERT-Large}               & 1024 & 138 & 105 & 85 & 286 & 226 & 208 & 567 & 527 & 554 \\
    \texttt{DeBERTa-v3-Large}         & 1024 & 196 & 133 & 29 & 377 & 299 & 113 & 688 & 635 & 423 \\
    \texttt{GPT-2-Small}              & 768 & 37 & 1 & 1 & 152 & 1 & 1 & 402 & 1 & 3 \\
    \texttt{GPT-2-Large}              & 1280 & 24 & 1 & 95 & 172 & 1 & 284 & 583 & 1 & 726 \\
    \texttt{GPT-2-XL}                 & 1600 & 113 & 1 & 118 & 340 & 1 & 356 & 838 & 1 & 914 \\
    \texttt{Pythia-6.9B}              & 4096 & 391 & 1 & 96 & 865 & 1 & 517 & 1952 & 1 & 1925 \\
    \texttt{Pythia-6.9B-Tulu}         & 4096 & 390 & 1 & 244 & 862 & 1 & 832 & 1949 & 1 & 2292 \\
    \texttt{OLMo-2-7B}                & 4096 & 404 & 310 & 41 & 833 & 896 & 299 & 1772 & 2279 & 1550 \\
    \texttt{OLMo-2-7B-Instruct}       & 4096 & 404 & 358 & 111 & 833 & 974 & 567 & 1772 & 2361 & 1964 \\
    \texttt{Gemma-2-2B}               & 2304 & 216 & 8 & 11 & 505 & 130 & 70 & 1129 & 794 & 611 \\
    \texttt{Gemma-2-2B-Instruct}      & 2304 & 222 & 22 & 8 & 520 & 198 & 57 & 1153 & 899 & 572 \\
    \texttt{Qwen-2.5-1.5B}            & 1536 & 184 & 1 & 9 & 399 & 1 & 50 & 835 & 1 & 452 \\
    \texttt{Qwen-2.5-1.5B-Instruct}   & 1536 & 184 & 1 & 11 & 394 & 1 & 70 & 820 & 1 & 533 \\
    \texttt{Llama-3.1-8B}             & 4096 & 373 & 240 & 35 & 789 & 727 & 187 & 1722 & 2051 & 1119 \\
    \texttt{Llama-3.1-8B-Instruct}    & 4096 & 372 & 215 & 31 & 788 & 664 & 181 & 1722 & 1957 & 1093 \\
    \bottomrule
\end{tabular}
} %
\caption{Number of principal-component axes required to reach 50\% (ID$_{50}$), 70\% (ID$_{70}$) and 90\% (ID$_{90}$) explained variance in the first, middle and last layers of each model.}
\label{fig:intrinsic_dim_table}
\end{table*}

\subsection{Massive Activations and Outlier Dimensions}
\label{sec:massive_activations_appendix}

We computed the maximum absolute activation, maximum mean (absolute value) per dimension, and maximum standard deviation per dimension across all layers for representative models to understand the low linear effective dimensionality observed in Table~\cref{fig:intrinsic_dim_table}.

\begin{figure*}[htbp]
  \centering
  \includegraphics[width=0.85\textwidth]{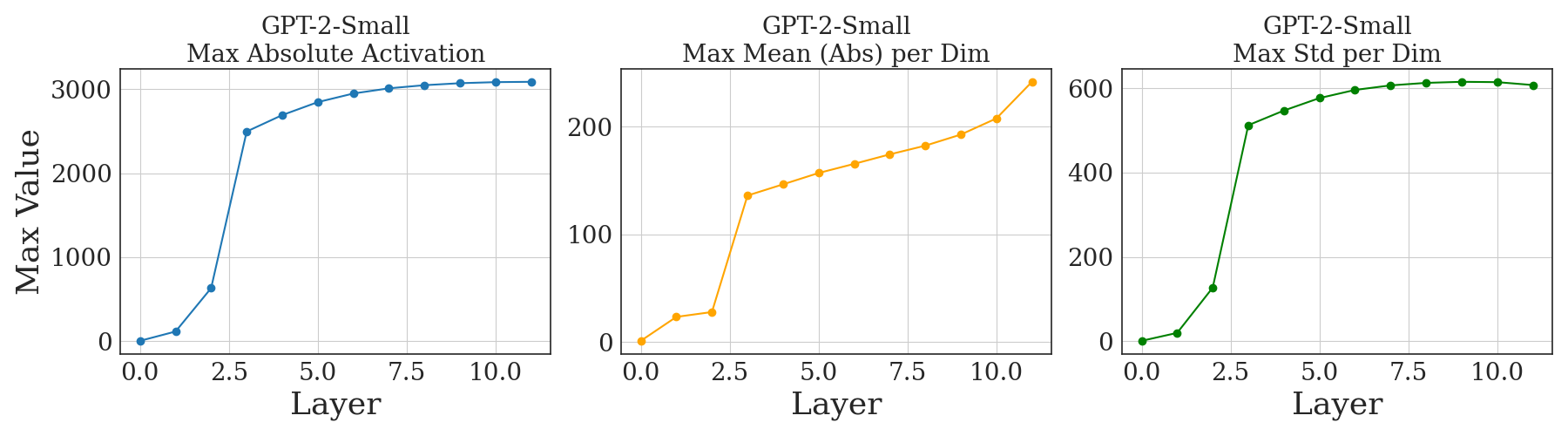}
  \caption{Activation statistics across layers for \texttt{GPT-2-Small}.}
  \label{fig:massive_activations_gpt2}
\end{figure*}

\begin{figure*}[htbp]
  \centering
  \includegraphics[width=0.85\textwidth]{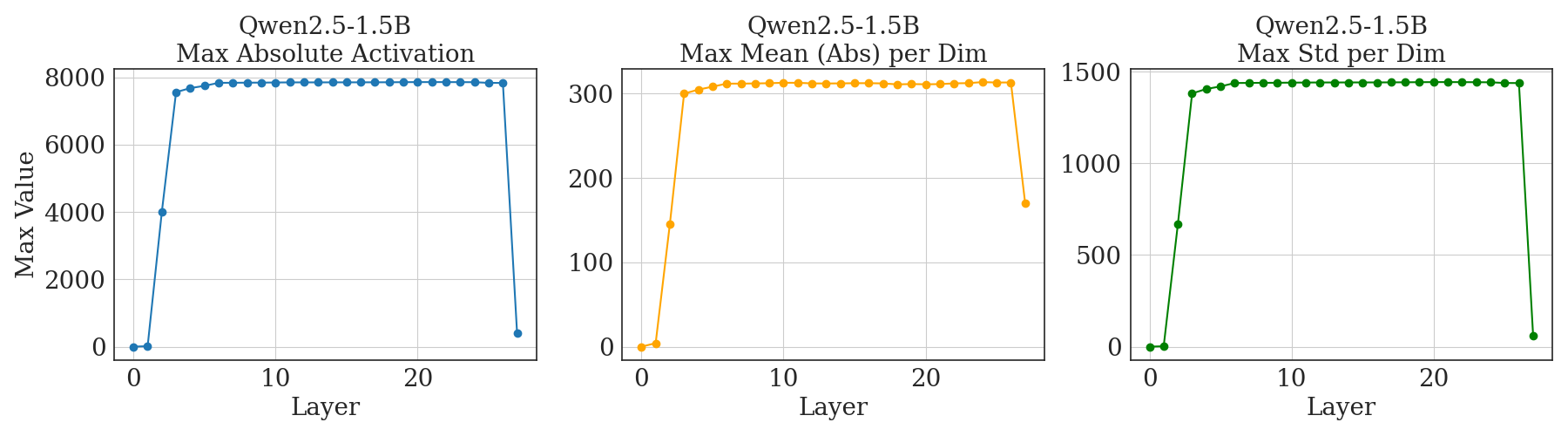}
  \caption{Activation statistics across layers for \texttt{Qwen2.5-1.5B}.}
  \label{fig:massive_activations_qwen2}
\end{figure*}

\begin{figure*}[htbp]
  \centering
  \includegraphics[width=0.85\textwidth]{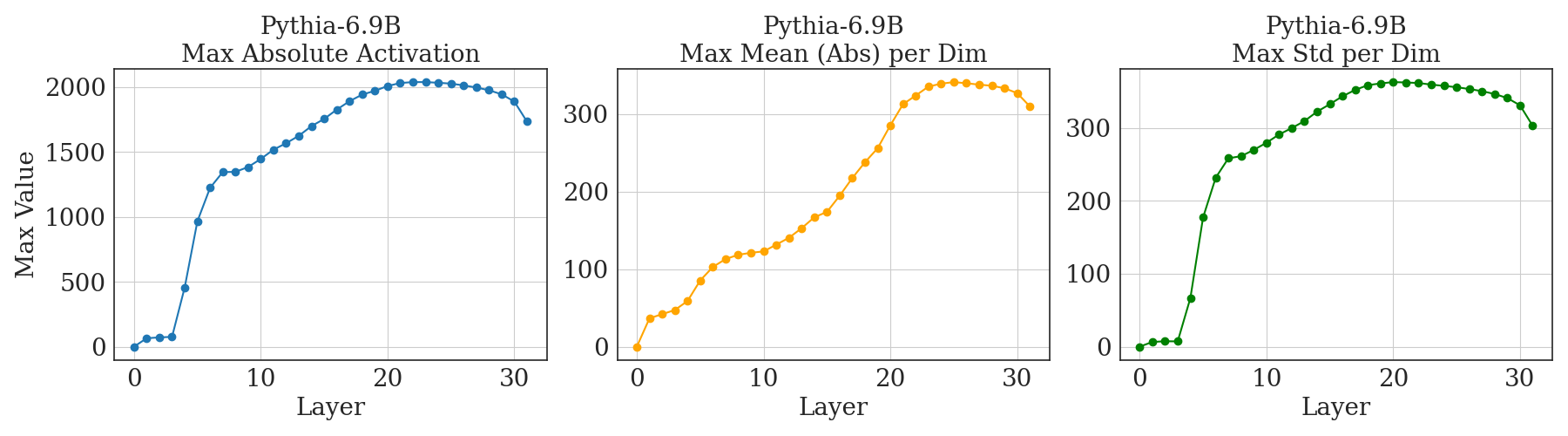}
  \caption{Activation statistics across layers for \texttt{Pythia-6.9B}.}
  \label{fig:massive_activations_pythia}
\end{figure*}

\begin{figure*}[htbp]
  \centering
  \includegraphics[width=0.85\textwidth]{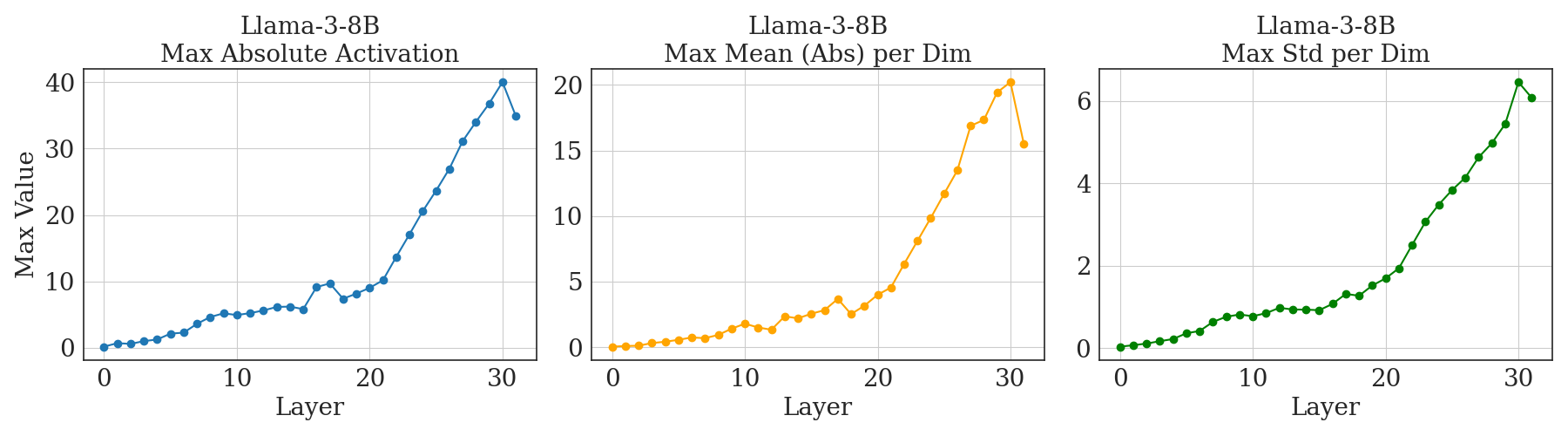}
  \caption{Activation statistics across layers for \texttt{Llama-3-8B}.}
  \label{fig:massive_activations_llama3}
\end{figure*}

\begin{figure*}[htbp]
  \centering
  \includegraphics[width=0.85\textwidth]{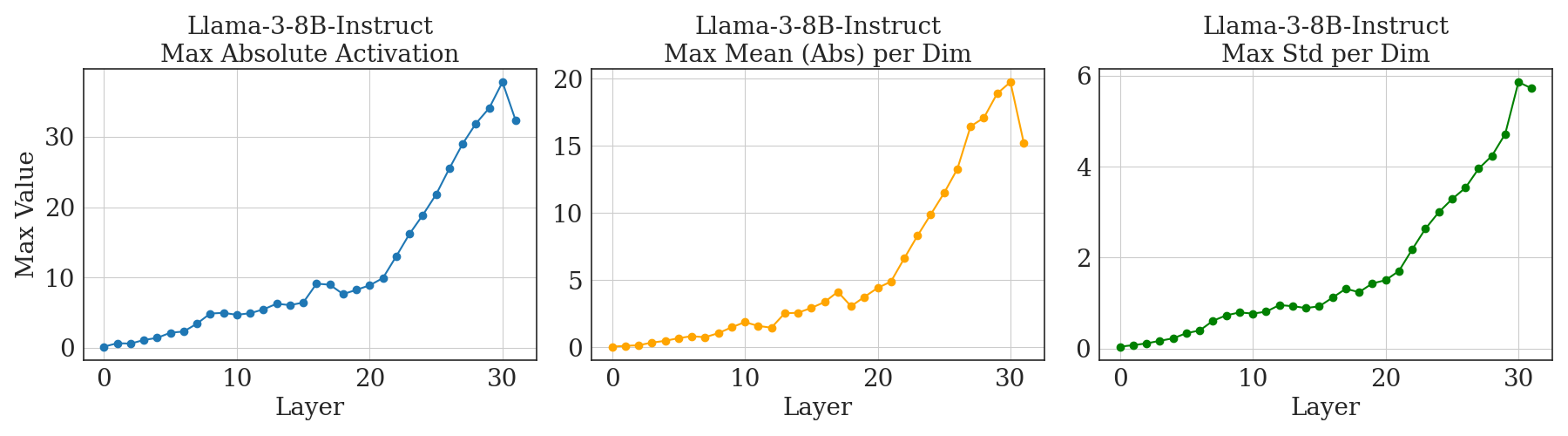}
  \caption{Activation statistics across layers for \texttt{Llama-3-8B-Instruct}.}
  \label{fig:massive_activations_llama3_instruct}
\end{figure*}

\begin{figure*}[htbp]
  \centering
  \includegraphics[width=0.85\textwidth]{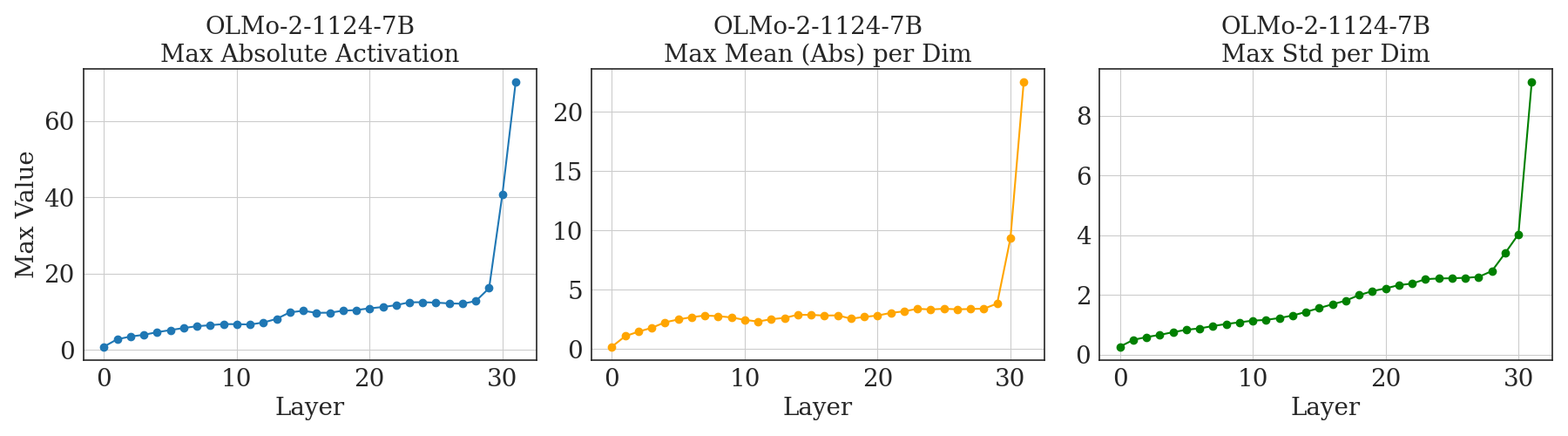}
  \caption{Activation statistics across layers for \texttt{OLMo-2-1124-7B}.}
  \label{fig:massive_activations_olmo2}
\end{figure*}

\begin{figure*}[htbp]
  \centering
  \includegraphics[width=0.85\textwidth]{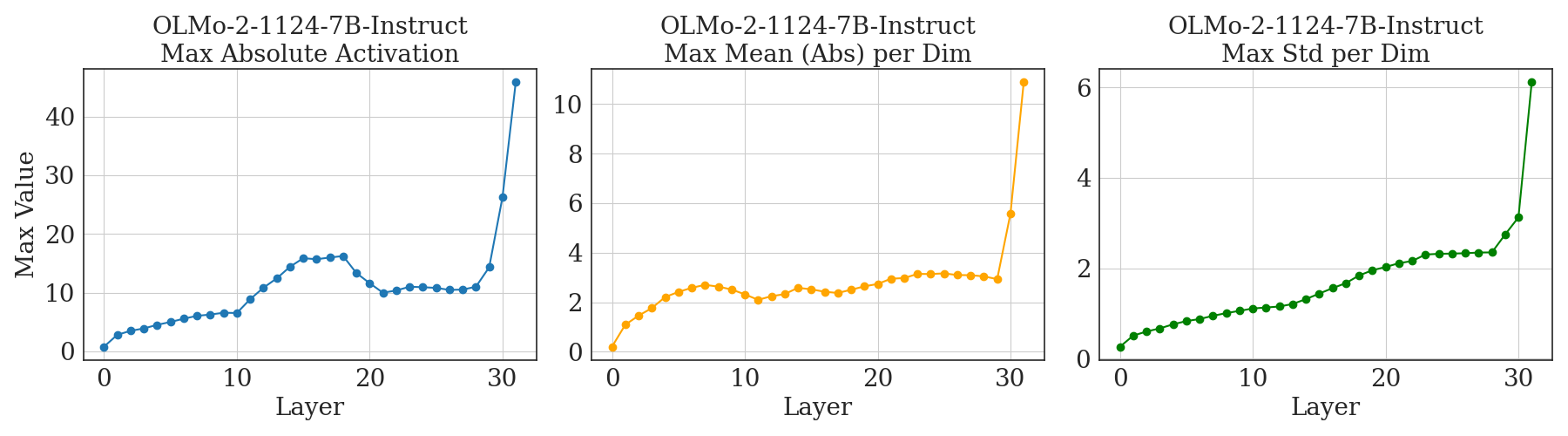}
  \caption{Activation statistics across layers for \texttt{OLMo-2-1124-7B-Instruct}.}
  \label{fig:massive_activations_olmo2_instruct}
\end{figure*}

Figures~\cref{fig:massive_activations_gpt2,fig:massive_activations_qwen2,fig:massive_activations_pythia,fig:massive_activations_llama3,fig:massive_activations_llama3_instruct,fig:massive_activations_olmo2,fig:massive_activations_olmo2_instruct} show the results. Models like \texttt{Qwen2.5-1.5B} and \texttt{GPT-2} variants show large maximum activation values. For example, \texttt{Qwen2.5-1.5B} reaches maximum absolute activations around 8000, while models like \texttt{Llama-3-8B} and \texttt{OLMo-1124-7B} show gradual increases across layers, with maximum values only reaching 30-40 in final layers.

This corresponds with the linear effective dimensionality measurements in Table~\cref{fig:intrinsic_dim_table}. Models with large activations in middle layers correspond to those requiring only 1-2 components to reach 50-90\% explained variance at those depths. Models with gradual activation increases correspond to those requiring hundreds of components at all depths. The presence of outlier dimensions with large activations makes the representation anisotropic, with variance concentrated along a small number of directions.

\subsection{Linear Separability Gap}
\label{sec:linear_separability_gap}

To determine whether the extracted linguistic signal benefits from non-linear decision boundaries, we compute the linear separability gap (defined in \cref{eq:linear_separability_gap}) as the difference in selectivity between the MLP and linear probes. A positive gap means the MLP extracts more \emph{selective} signal than the linear probe, while a negative gap indicates that additional probe capacity decreases selectivity, consistent with the MLP exploiting spurious correlations rather than improving extraction of genuine linguistic structure.

\begin{figure*}[htbp]
  \centering
  {\textbf{Linear Separability Gap}\par\medskip}
  \begin{minipage}[b]{0.9\textwidth}
    \centering
    \includegraphics[width=\textwidth]{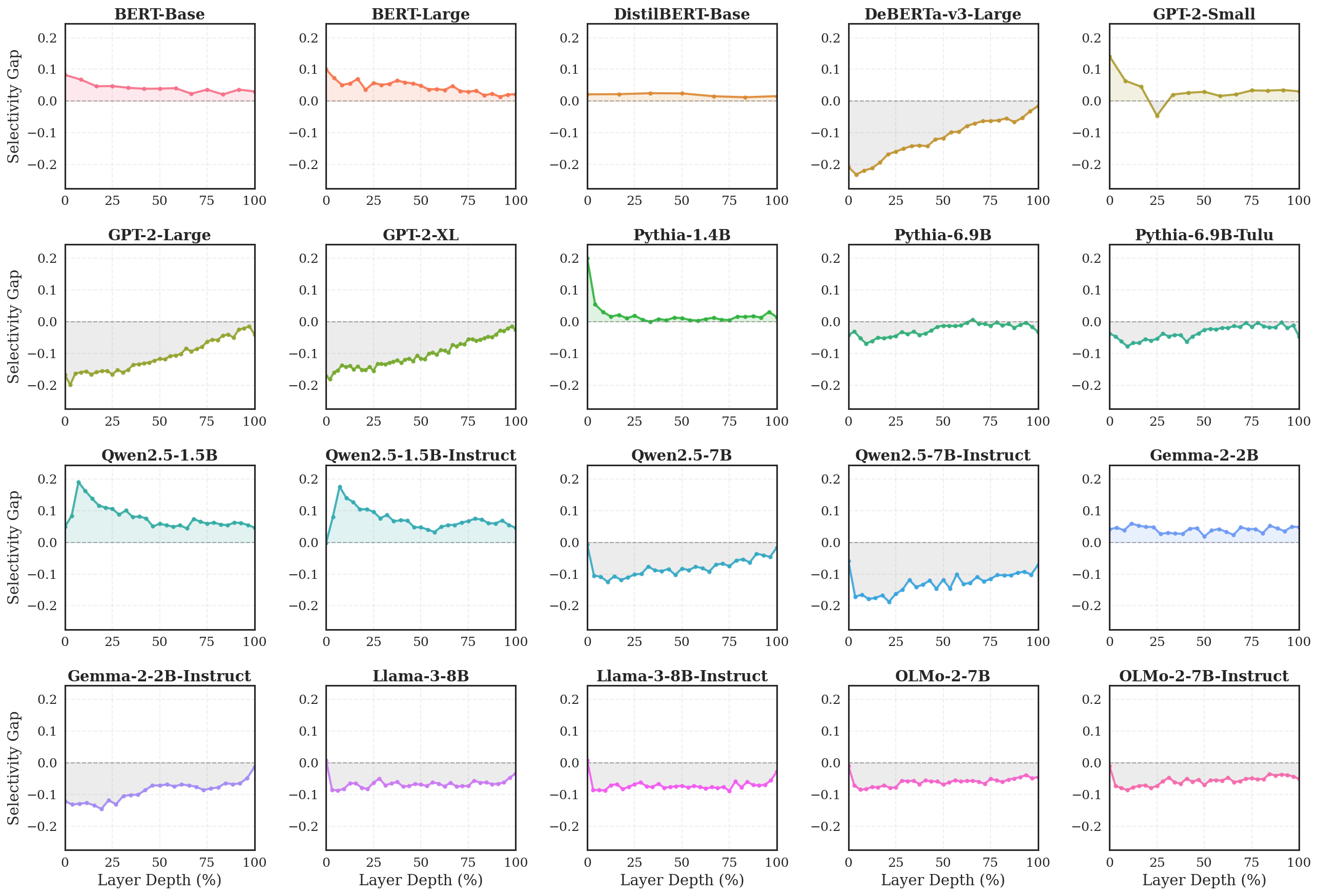}
    \par\smallskip
    \textbf{(a)} Linear separability gap for inflection prediction\\
    \label{fig:mlp_advantage_inflection}
  \end{minipage}
  
  \vspace{0.2cm}
  
  \begin{minipage}[b]{0.9\textwidth}
    \centering
    \includegraphics[width=\textwidth]{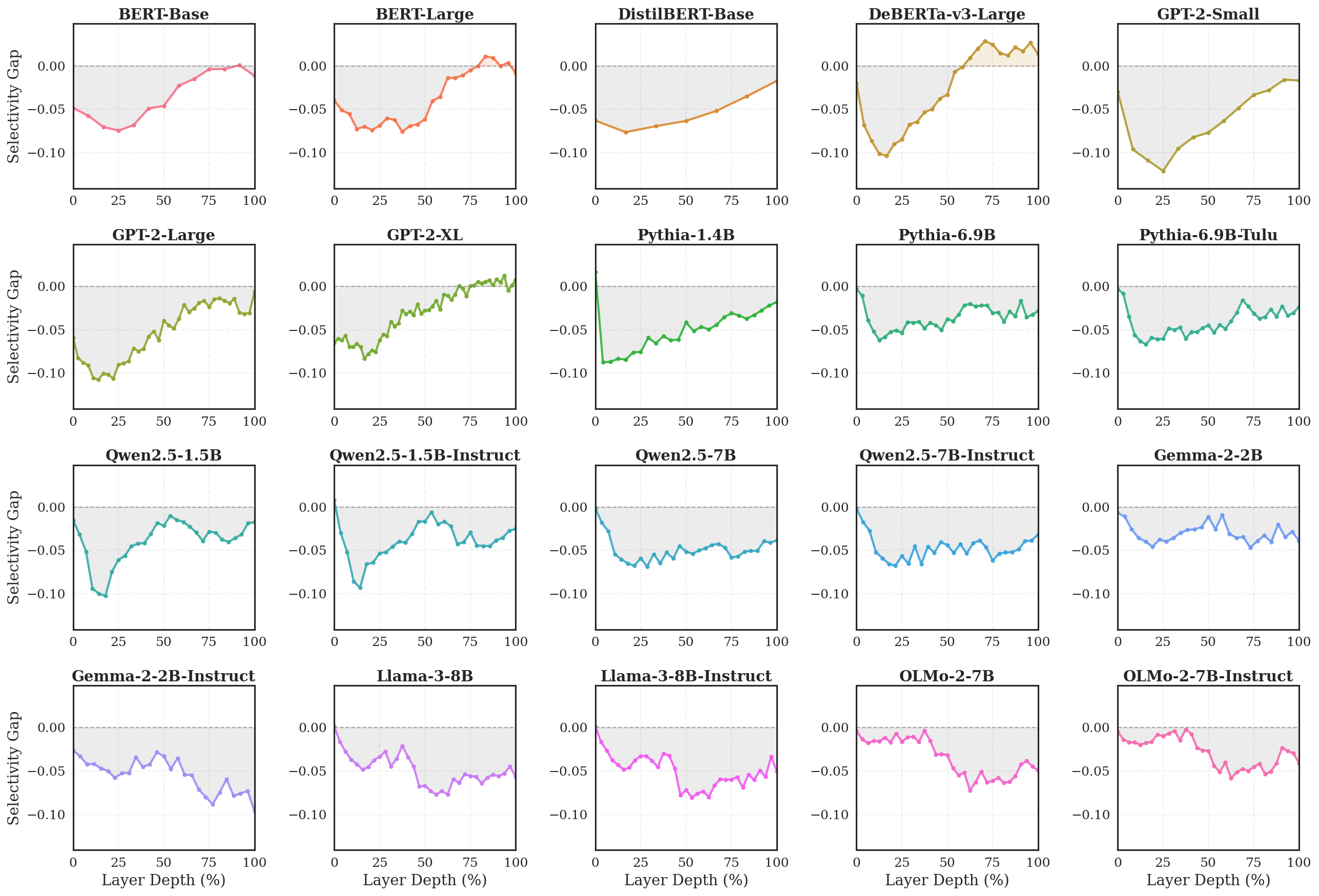}
    \par\smallskip
    \textbf{(b)} Linear separability gap for lemma prediction\\
    \label{fig:mlp_advantage_lemma}
  \end{minipage}
  \caption{Linear separability gap (difference in selectivity between MLP and linear probes) across model layers for English. The gap measures how much a non-linear transformation improves the extraction of genuine linguistic signal compared to a simple linear mapping.}
  \label{fig:mlp_advantage}
\end{figure*}

\Cref{fig:mlp_advantage} shows these results for English. For both inflection (\Cref{fig:mlp_advantage_inflection}) and lemma (\Cref{fig:mlp_advantage_lemma}), gaps are typically near zero or negative across much of the depth range, indicating that linear probes are at least as selective as MLP probes in most models and layers. This is particularly salient for lemma: despite occasional accuracy improvements from the MLP, the selectivity gap often remains negative, suggesting that the additional accuracy is frequently driven by non-linguistic memorization effects rather than more structured lexical encoding.

\begin{figure*}[htbp]
  \centering
  \begin{minipage}[b]{0.48\linewidth}
    \centering
    \includegraphics[width=\textwidth]{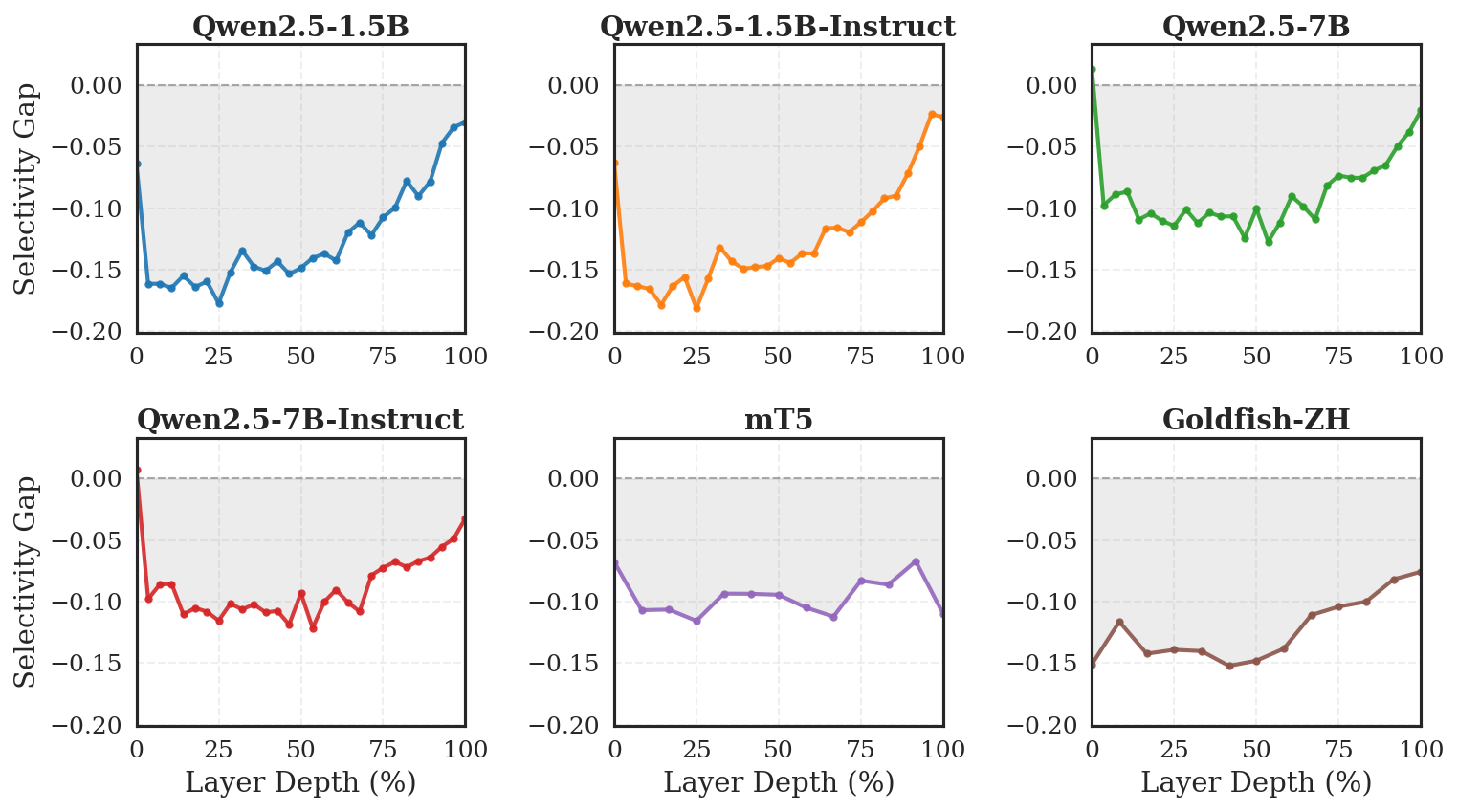}
    \par\smallskip
    \small\textbf{(a)} Inflection
  \end{minipage}
  \hfill
  \begin{minipage}[b]{0.48\linewidth}
    \centering
    \includegraphics[width=\textwidth]{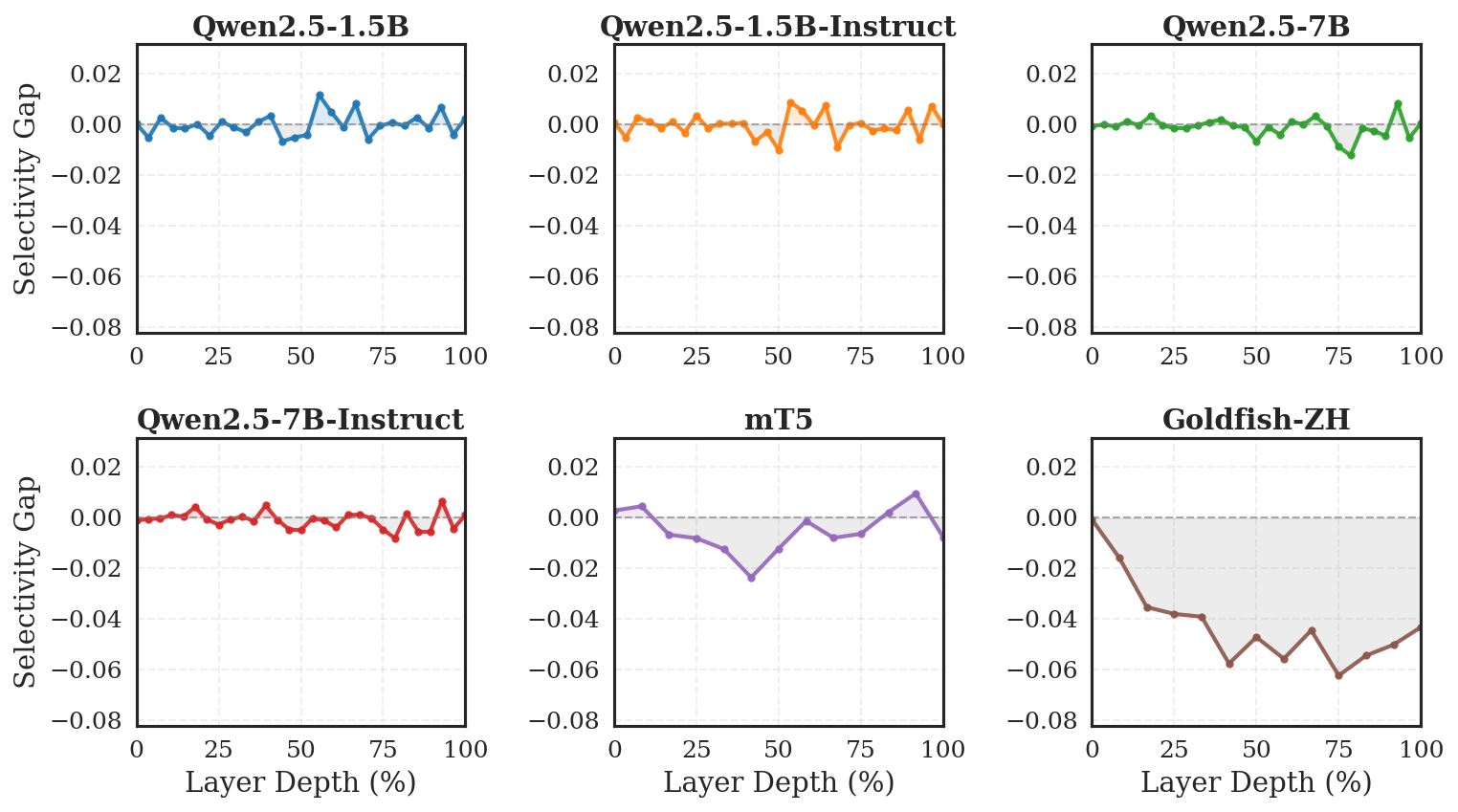}
    \par\smallskip
    \small\textbf{(b)} Lemma
  \end{minipage}
  \caption{Linear separability gap for Chinese.}
  \label{fig:mlp_advantage_chinese}
\end{figure*}

\begin{figure*}[htbp]
  \centering
  \begin{minipage}[b]{0.48\linewidth}
    \centering
    \includegraphics[width=\textwidth]{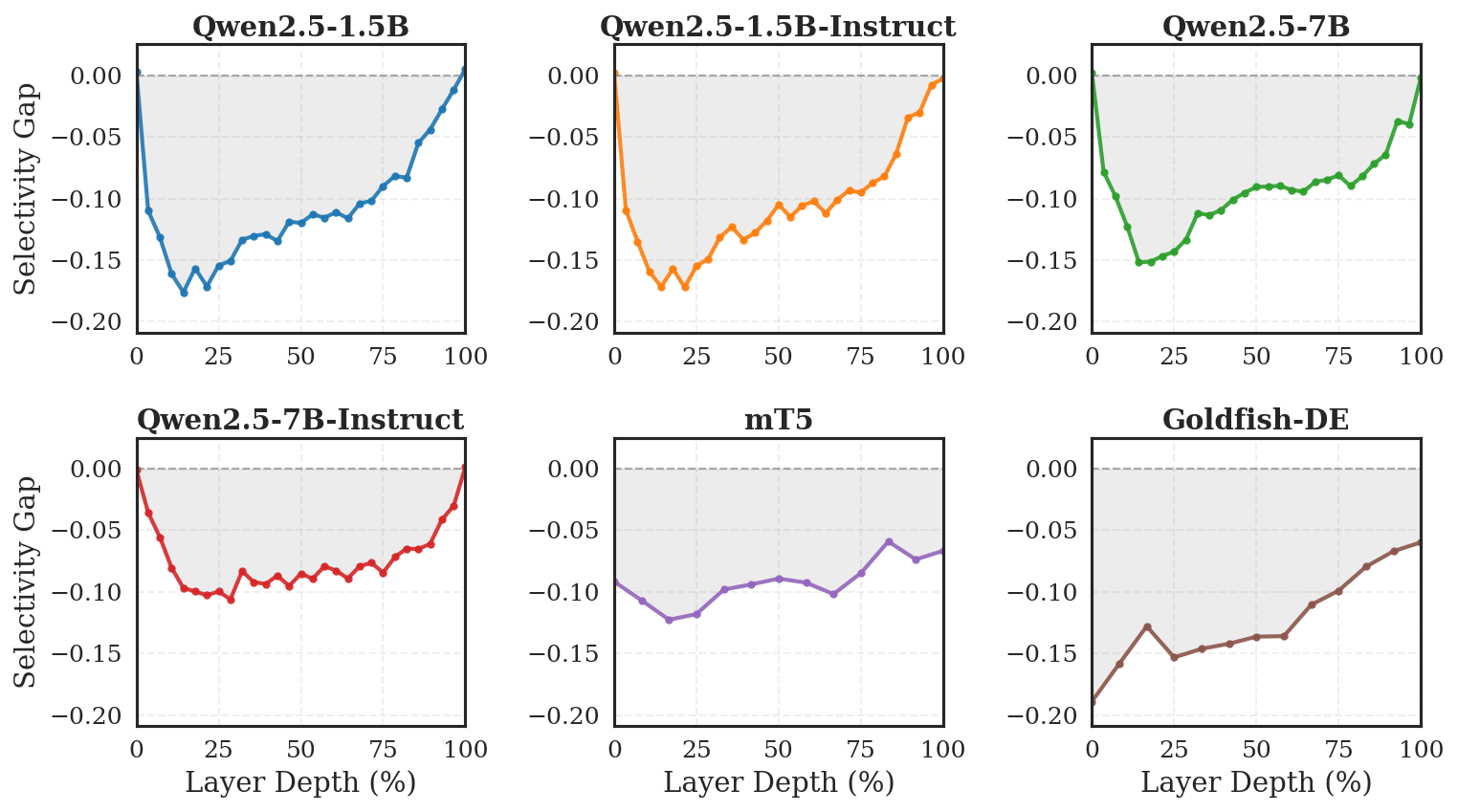}
    \par\smallskip
    \small\textbf{(a)} Inflection
  \end{minipage}
  \hfill
  \begin{minipage}[b]{0.48\linewidth}
    \centering
    \includegraphics[width=\textwidth]{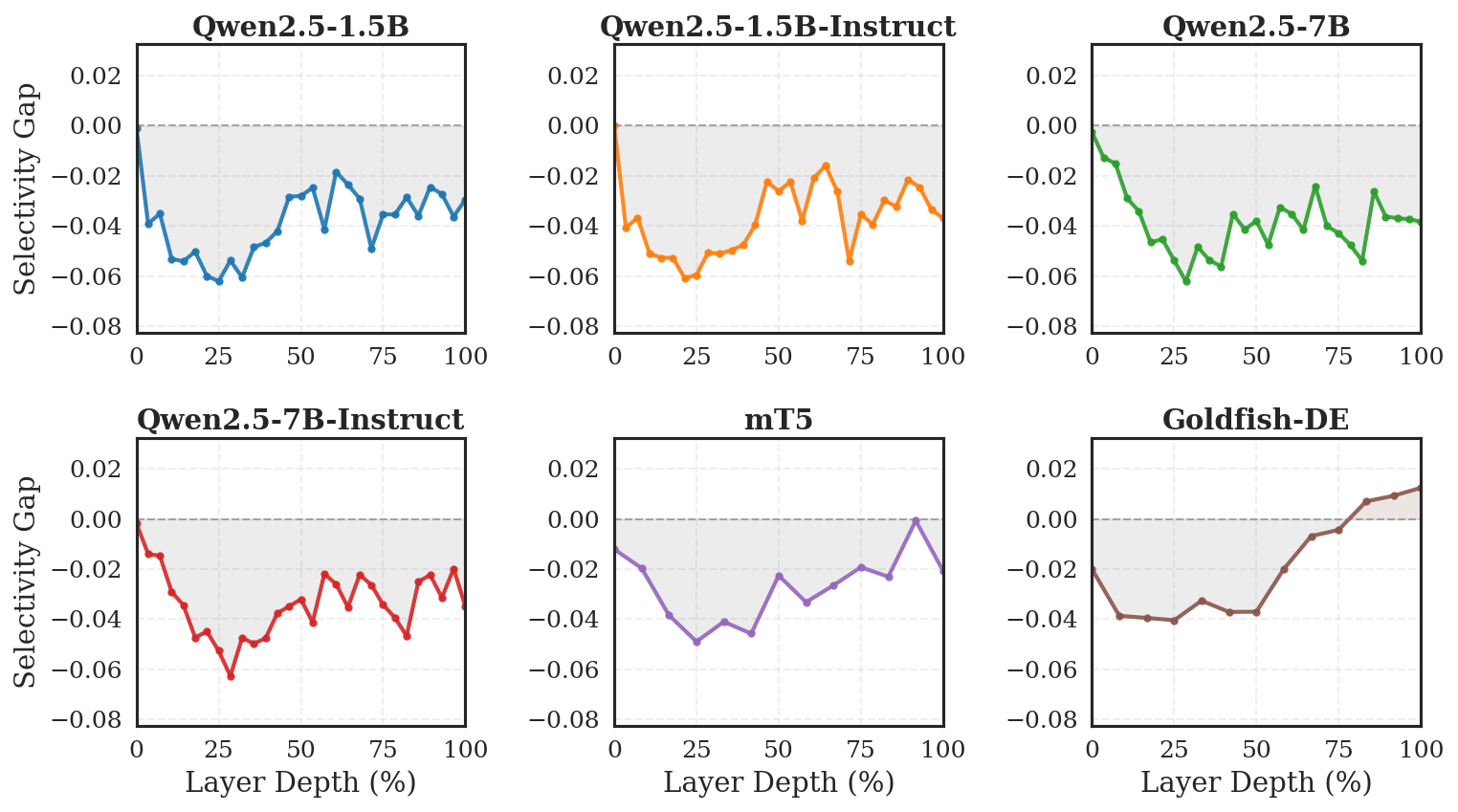}
    \par\smallskip
    \small\textbf{(b)} Lemma
  \end{minipage}
  \caption{Linear separability gap for German.}
  \label{fig:mlp_advantage_german}
\end{figure*}

\begin{figure*}[htbp]
  \centering
  \begin{minipage}[b]{0.48\linewidth}
    \centering
    \includegraphics[width=\textwidth]{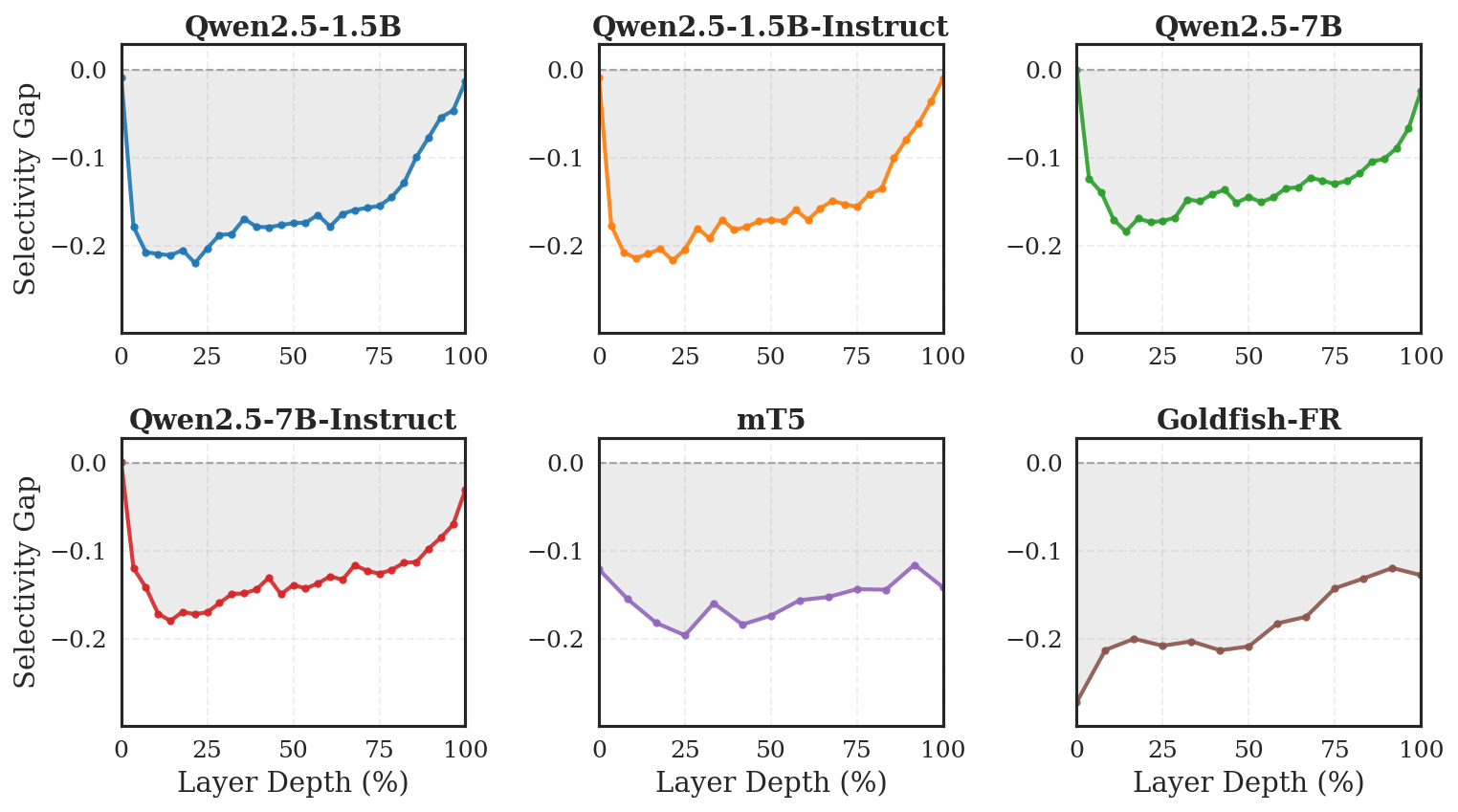}
    \par\smallskip
    \small\textbf{(a)} Inflection
  \end{minipage}
  \hfill
  \begin{minipage}[b]{0.48\linewidth}
    \centering
    \includegraphics[width=\textwidth]{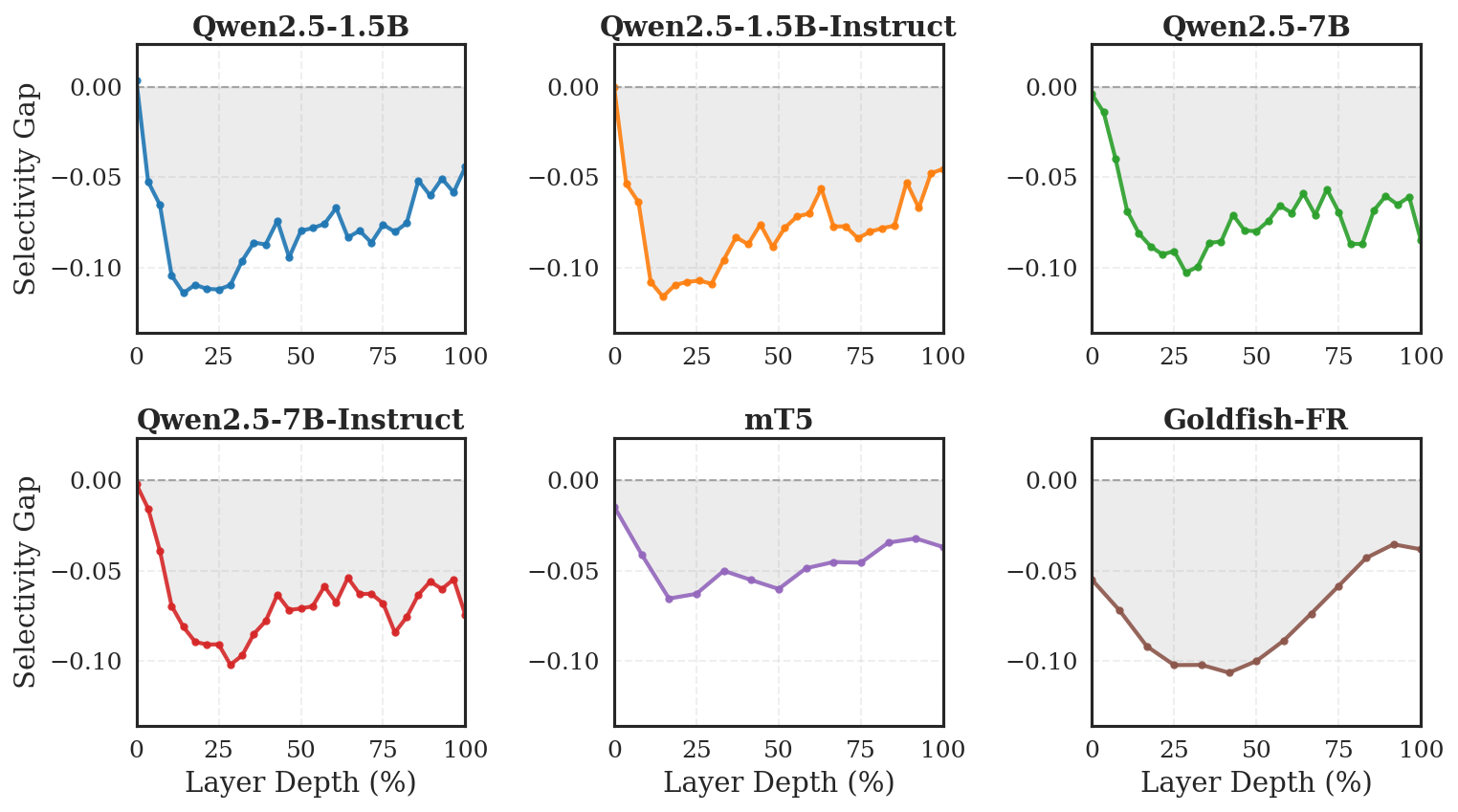}
    \par\smallskip
    \small\textbf{(b)} Lemma
  \end{minipage}
  \caption{Linear separability gap for French.}
  \label{fig:mlp_advantage_french}
\end{figure*}

\begin{figure*}[htbp]
  \centering
  \begin{minipage}[b]{0.48\linewidth}
    \centering
    \includegraphics[width=\textwidth]{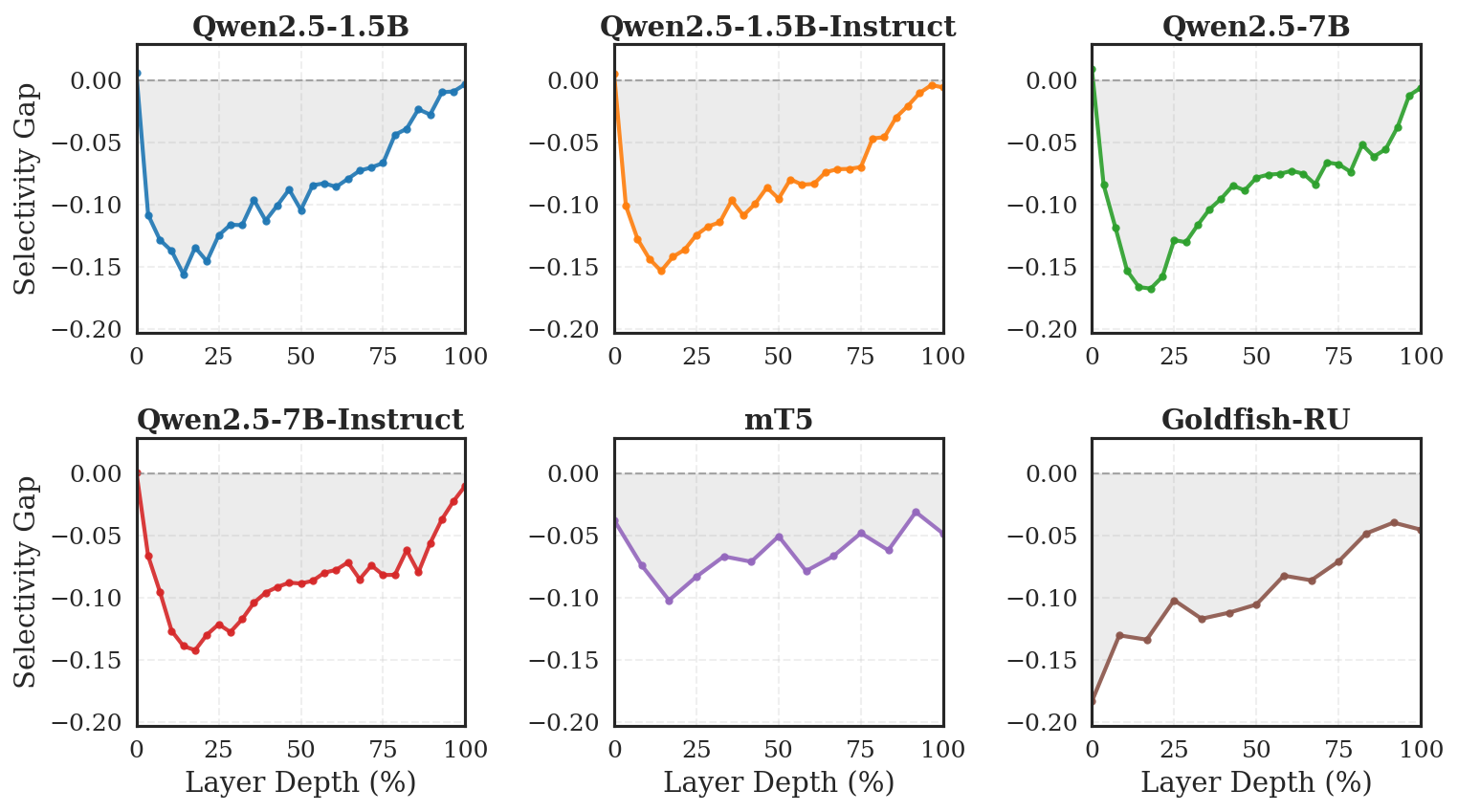}
    \par\smallskip
    \small\textbf{(a)} Inflection
  \end{minipage}
  \hfill
  \begin{minipage}[b]{0.48\linewidth}
    \centering
    \includegraphics[width=\textwidth]{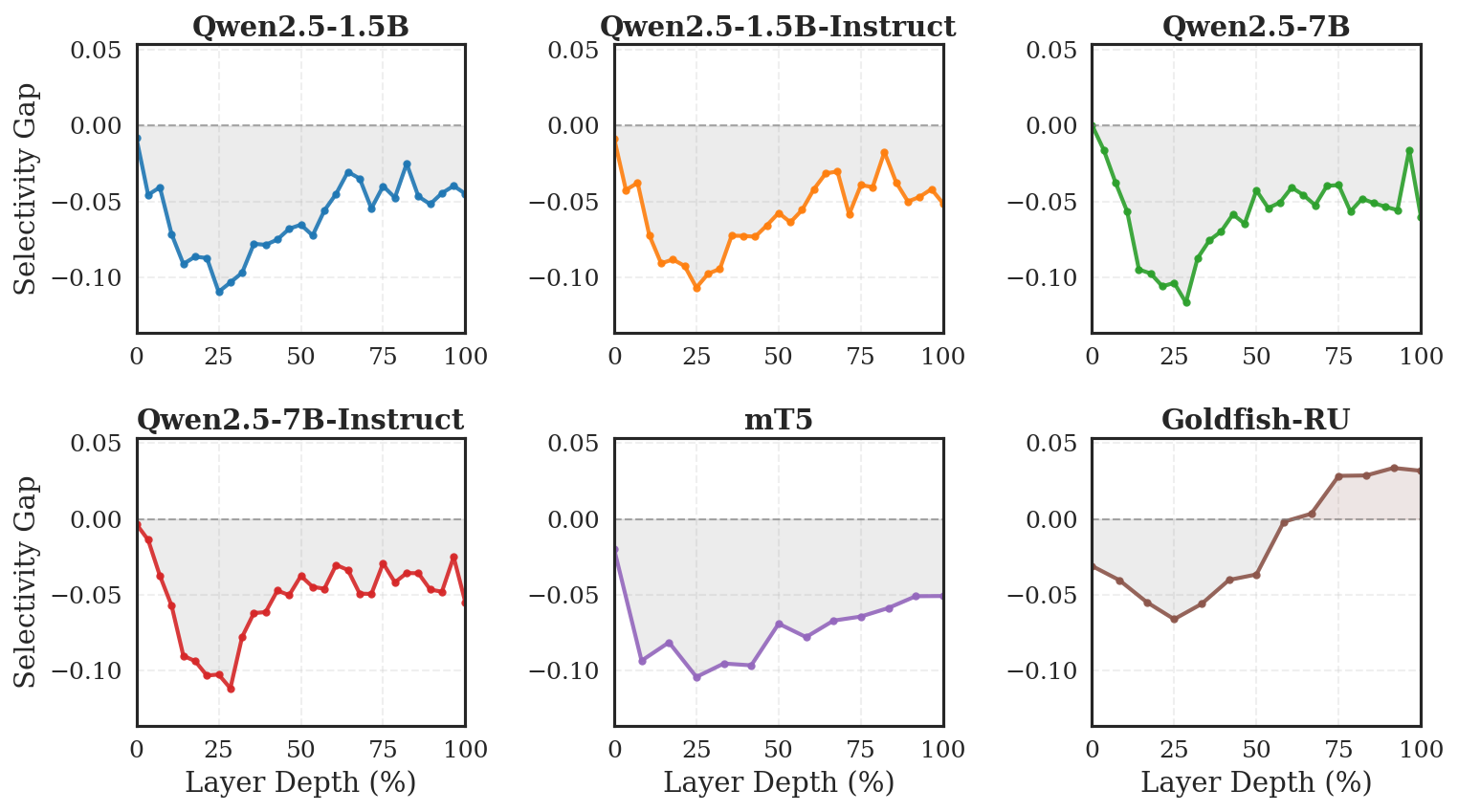}
    \par\smallskip
    \small\textbf{(b)} Lemma
  \end{minipage}
  \caption{Linear separability gap for Russian.}
  \label{fig:mlp_advantage_russian}
\end{figure*}

\begin{figure*}[htbp]
  \centering
  \begin{minipage}[b]{0.48\linewidth}
    \centering
    \includegraphics[width=\textwidth]{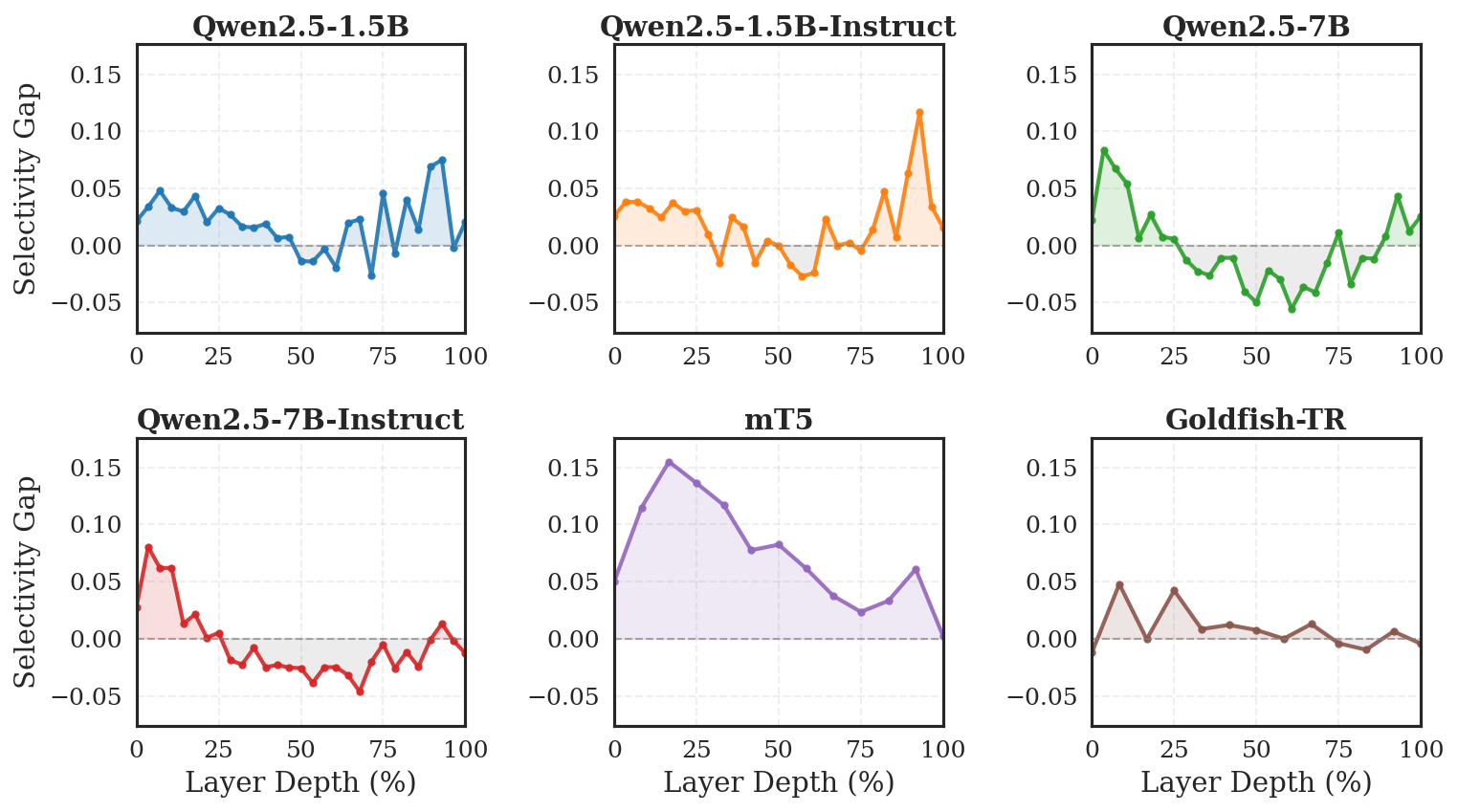}
    \par\smallskip
    \small\textbf{(a)} Inflection
  \end{minipage}
  \hfill
  \begin{minipage}[b]{0.48\linewidth}
    \centering
    \includegraphics[width=\textwidth]{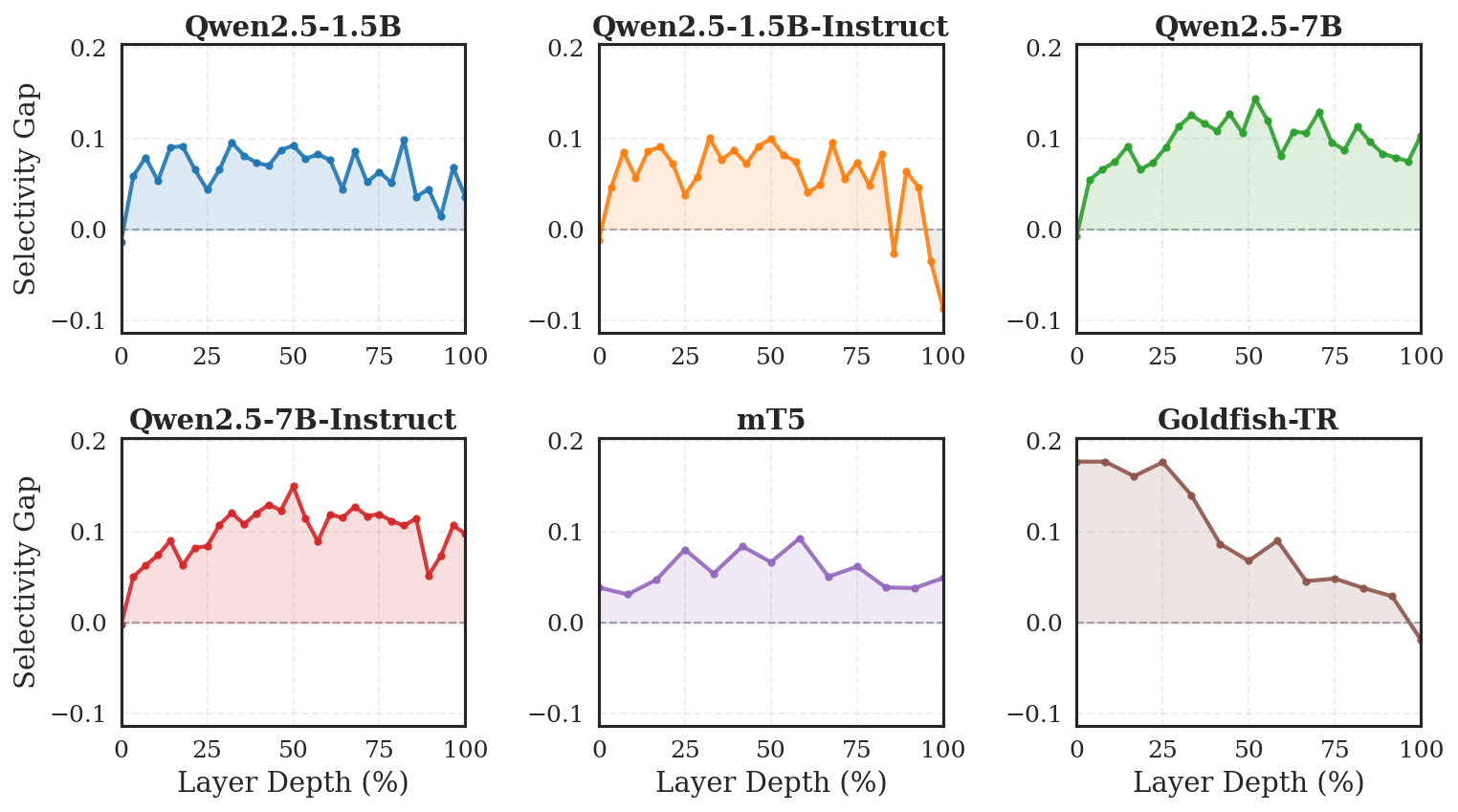}
    \par\smallskip
    \small\textbf{(b)} Lemma
  \end{minipage}
  \caption{Linear separability gap for Turkish.}
  \label{fig:mlp_advantage_turkish}
\end{figure*}

\Cref{fig:mlp_advantage_chinese,fig:mlp_advantage_german,fig:mlp_advantage_french,fig:mlp_advantage_russian,fig:mlp_advantage_turkish} extend this analysis cross-linguistically. Across Chinese, German, French, and Russian, the gap is predominantly negative for inflection and generally slightly negative (or near zero) for lemma, again implying that increased non-linear capacity rarely improves selectivity. Turkish is a notable exception: for several model families the gap is positive (especially for lemma), indicating cases where an MLP can extract additional selective signal beyond a linear readout. Overall, these trends suggest that by the selectivity criterion, most of the recoverable lexical and morphological signal is already accessible to a regularized linear probe, and extra probe capacity more often harms than helps.

\subsection{Training Dynamics}
\label{sec:training_dynamics_appendix}

\begin{figure*}[htbp]
  \centering
  \includegraphics[width=\textwidth]{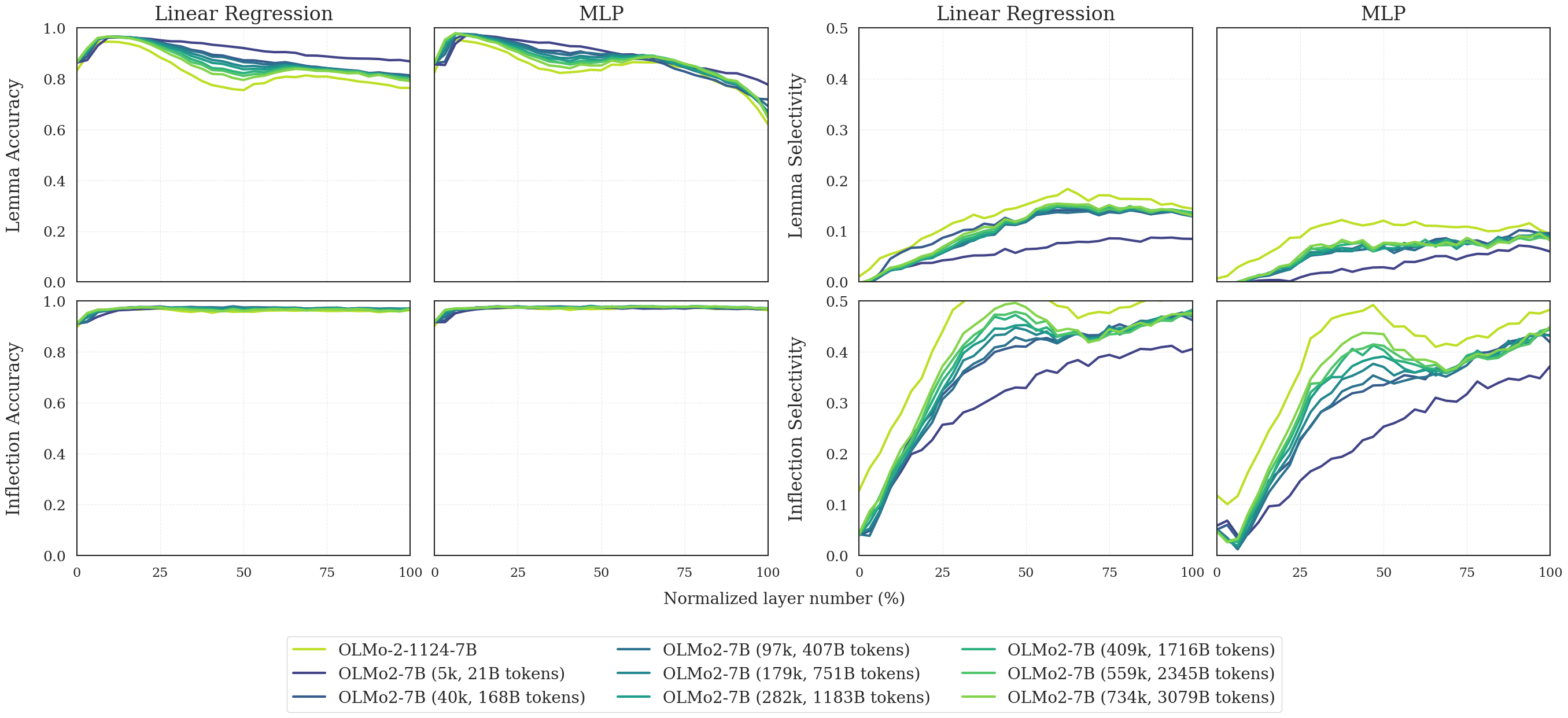}
  \caption{\textbf{\texttt{OLMo-2-7B} Training Dynamics.} Performance across pretraining checkpoints (5k--734k steps) for English. The full model is 928k steps. Checkpoints are colored from brightest (earliest) to darkest (latest). \textbf{Left:} Prediction accuracy for Lemma (top) and Inflection (bottom). Early checkpoints exhibit higher lemma accuracy than later ones, while inflectional accuracy remains flat. \textbf{Right:} Selectivity scores for the same tasks. Selectivity generally increases with model depth and training steps, particularly for inflection.}
  \label{fig:olmo2_combined}
\end{figure*}

\begin{figure*}[htbp]
  \centering
  \includegraphics[width=\textwidth]{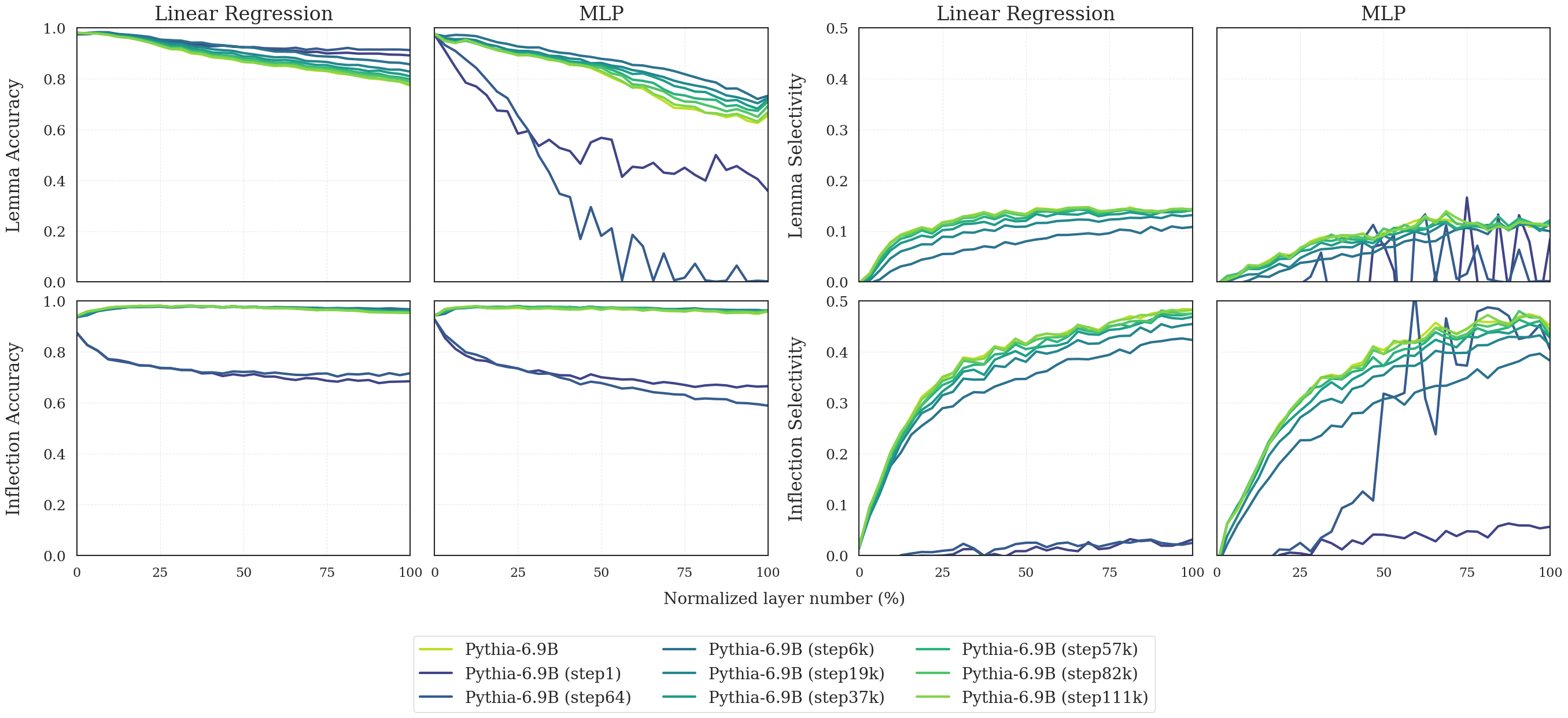}
  \caption{\textbf{\texttt{Pythia-6.9B} Training Dynamics.} Performance across pretraining checkpoints (step 1--111k) for English. The full model is 143k steps. Checkpoints are colored from brightest (earliest) to darkest (latest). \textbf{Left:} Prediction accuracy for Lemma (top) and Inflection (bottom). Lemma accuracy declines both with deeper layers and with more training, whereas inflectional accuracy stays uniformly high. \textbf{Right:} Selectivity scores for the same tasks, showing distinct separation between early and late checkpoints in the inflection task.}
  \label{fig:pythia_combined}
\end{figure*}

We analyze how linguistic information encoding evolves during the pretraining process. \Cref{fig:olmo2_combined} and \Cref{fig:pythia_combined} visualize the probing performance across training checkpoints for \texttt{OLMo-2-7B} and \texttt{Pythia-6.9B}, respectively.

For \texttt{OLMo-2-7B} (\Cref{fig:olmo2_combined}), we observe a counter-intuitive trend in lemma accuracy: early checkpoints (lighter curves) actually exhibit higher accuracy than the fully trained model (darker curves). This suggests that as the model matures, it abstracts away from rigid surface-level lemma identities, making them harder to probe linearly. However, the selectivity scores (right) consistently increase with training steps, indicating that while raw accuracy drops, the model becomes better at distinguishing linguistic features from control tasks.

A similar pattern is observed for \texttt{Pythia-6.9B} (\Cref{fig:pythia_combined}). Lemma accuracy significantly declines both with increasing layer depth and with more training steps. Conversely, inflectional accuracy remains robust and high throughout training. The selectivity plots demonstrate a clear separation between early and late checkpoints, particularly for inflection tasks. This confirms that pretraining helps the model refine its internal representations, suppressing irrelevant correlations (the control task) while maintaining necessary information.

\section{Attention Head Analysis}
\label{sec:attention_analysis_appendix}

We conducted additional experiments analyzing attention head outputs alongside residual stream representations to understand how different components of transformer models contribute to linguistic encoding.

\subsection{Methodology}

We averaged activations across all attention heads at each layer for \texttt{Qwen2.5-1.5B} and \texttt{Qwen2.5-1.5B-Instruct} models using the English dataset. 
We then trained linear regression and MLP classifiers on both attention head outputs and residual stream representations to compare their encoding patterns.

\subsection{Results}

\begin{figure*}[htbp]
  \centering
  \includegraphics[width=\textwidth]{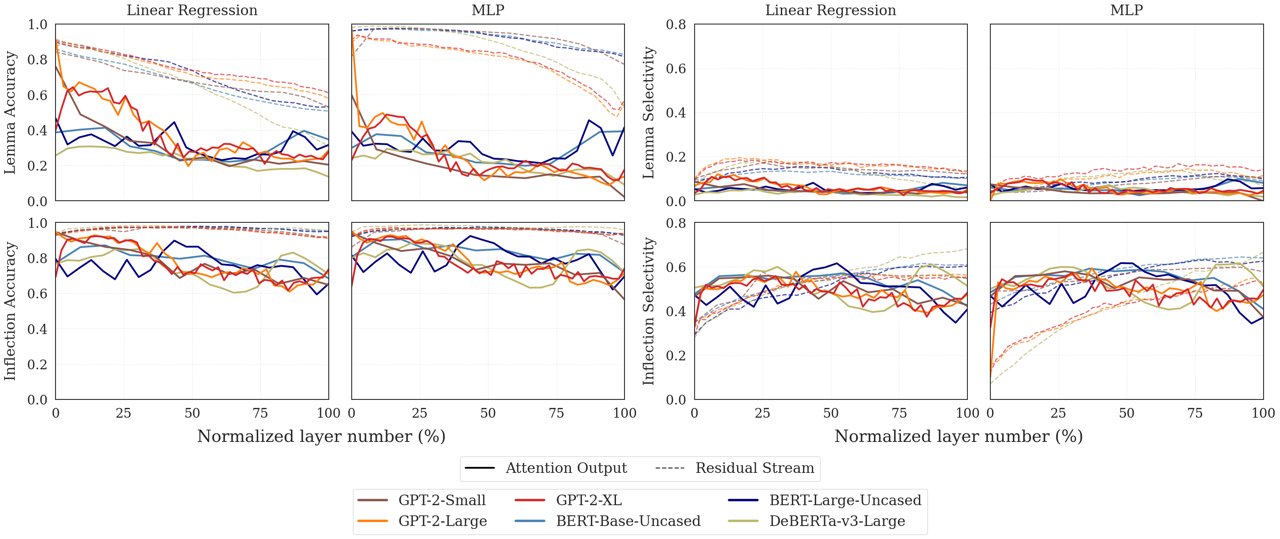}
  \caption{Linguistic task accuracy (left two columns) and classifier selectivity (right two columns) for attention head outputs (solid lines) versus residual stream representations (dashed lines) across BERT and GPT-2 model families. The top row corresponds to Lemma tasks, and the bottom row to Inflection tasks.}
  \label{fig:attention_bert_combined}
\end{figure*}

\begin{figure*}[htbp]
  \centering
  \includegraphics[width=\textwidth]{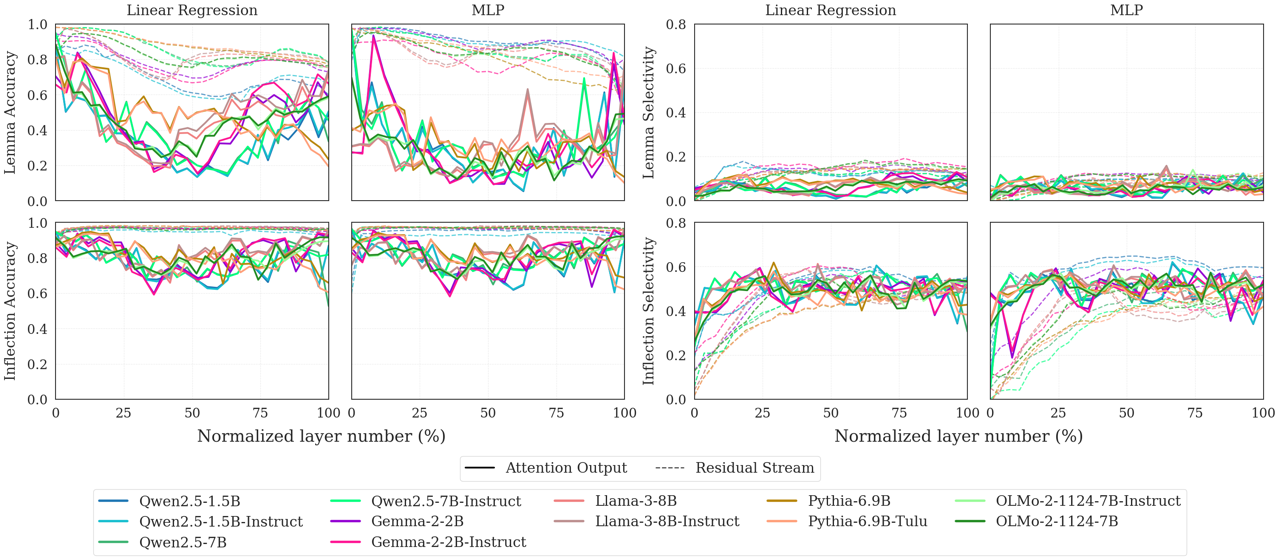}
  \caption{Linguistic task accuracy (left two columns) and classifier selectivity (right two columns) for attention head outputs (solid lines) versus residual stream representations (dashed lines) across contemporary model families. The top row corresponds to Lemma tasks, and the bottom row to Inflection tasks.}
  \label{fig:attention_other_combined}
\end{figure*}

\Cref{fig:attention_bert_combined,fig:attention_other_combined} compare probe accuracy and selectivity when trained on attention head outputs versus residual stream representations for lemma and inflection tasks across all English model families. Across both encoder-only and decoder-only architectures, residual stream representations consistently yield higher accuracy.

\section{Steering Vector Analysis}
\label{sec:steering_analysis_appendix}

We conducted steering vector experiments to test whether inflectional representations can be functionally manipulated and to understand model sensitivity to activation interventions.

\subsection{Methodology}

For each inflectional category, we computed steering vectors as:
\begin{equation}
\mathbf{s}_i = \mathbf{\mu}_i - \lambda \cdot \frac{1}{|C| - 1} \sum_{j \in C, j \neq i} \mathbf{\mu}_j
\end{equation}

We tested multiple values of $\lambda$ (1, 5, 10, 20, 100) and measured the impact on linear classifier performance when adding these steering vectors to existing activations for 1000 test words. We evaluated both mean probability change and prediction flip rate across all models.

\subsection{Results}

\begin{figure*}[htbp]
  \centering
  \includegraphics[width=\textwidth]{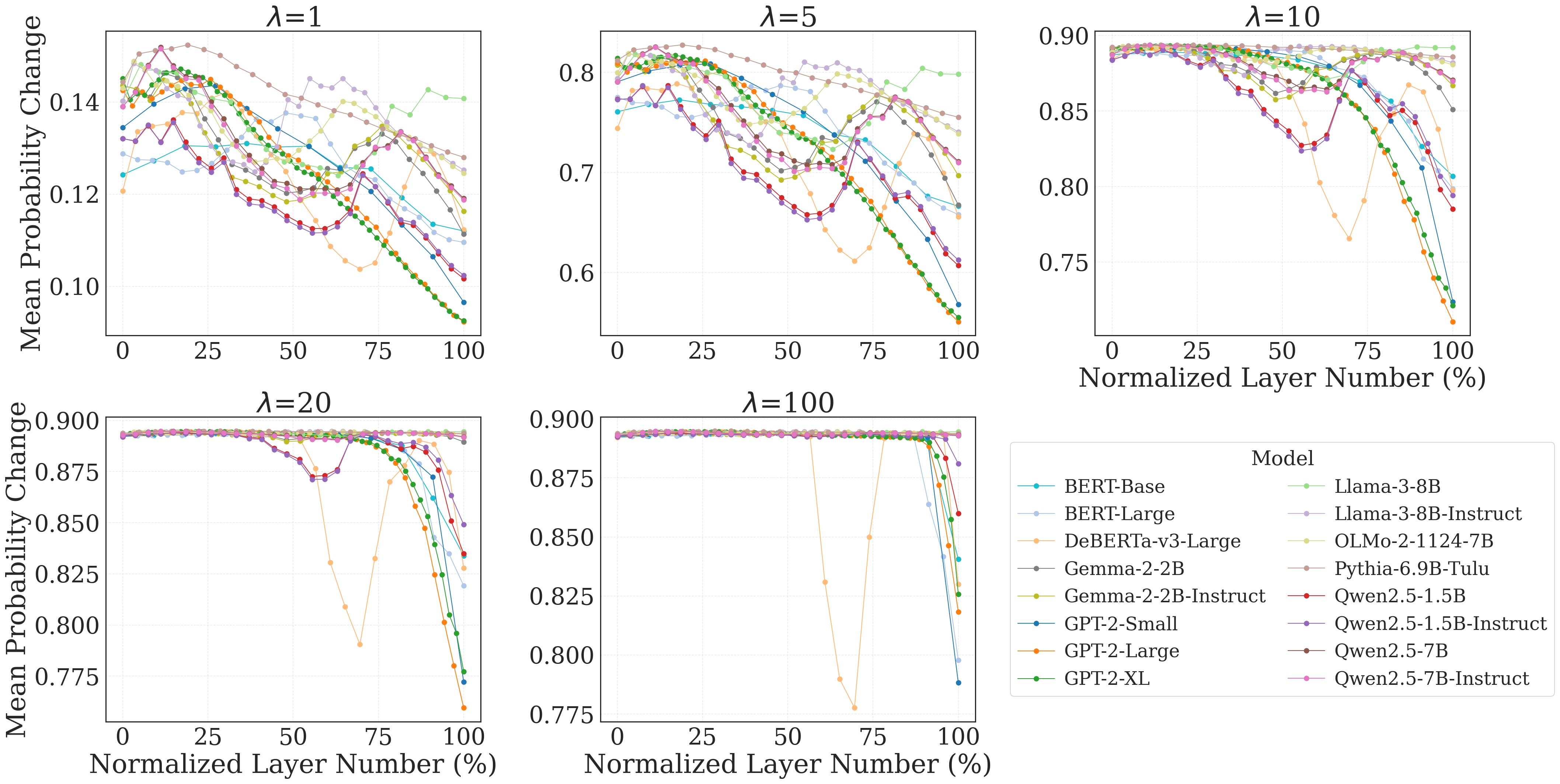}
  \caption{Mean probability change for inflection prediction when applying steering vectors across different $\lambda$ values. Five panels show results for $\lambda \in \{1, 5, 10, 20, 100\}$. All models start with high effectiveness ($\approx$0.9-1.0) at layer 0. Most models maintain stable performance across depth, with \texttt{DeBERTa-v3-Large} a notable exception, showing a sudden drop around 75\% of model depth. Higher $\lambda$ values increase steering effectiveness while preserving the overall pattern.}
  \label{fig:steering_probability_change}
\end{figure*}

\Cref{fig:steering_probability_change} shows the mean probability change for inflection prediction when applying steering vectors across different $\lambda$ values (1, 5, 10, 20, and 100). Each panel demonstrates that steering effectiveness is high across all models, with most maintaining stable performance throughout network depth. A notable exception is \texttt{DeBERTa-v3-Large}, which shows a sudden drop in steering effectiveness around 75\% of model depth.

\begin{figure*}[htbp]
  \centering
  \includegraphics[width=\textwidth]{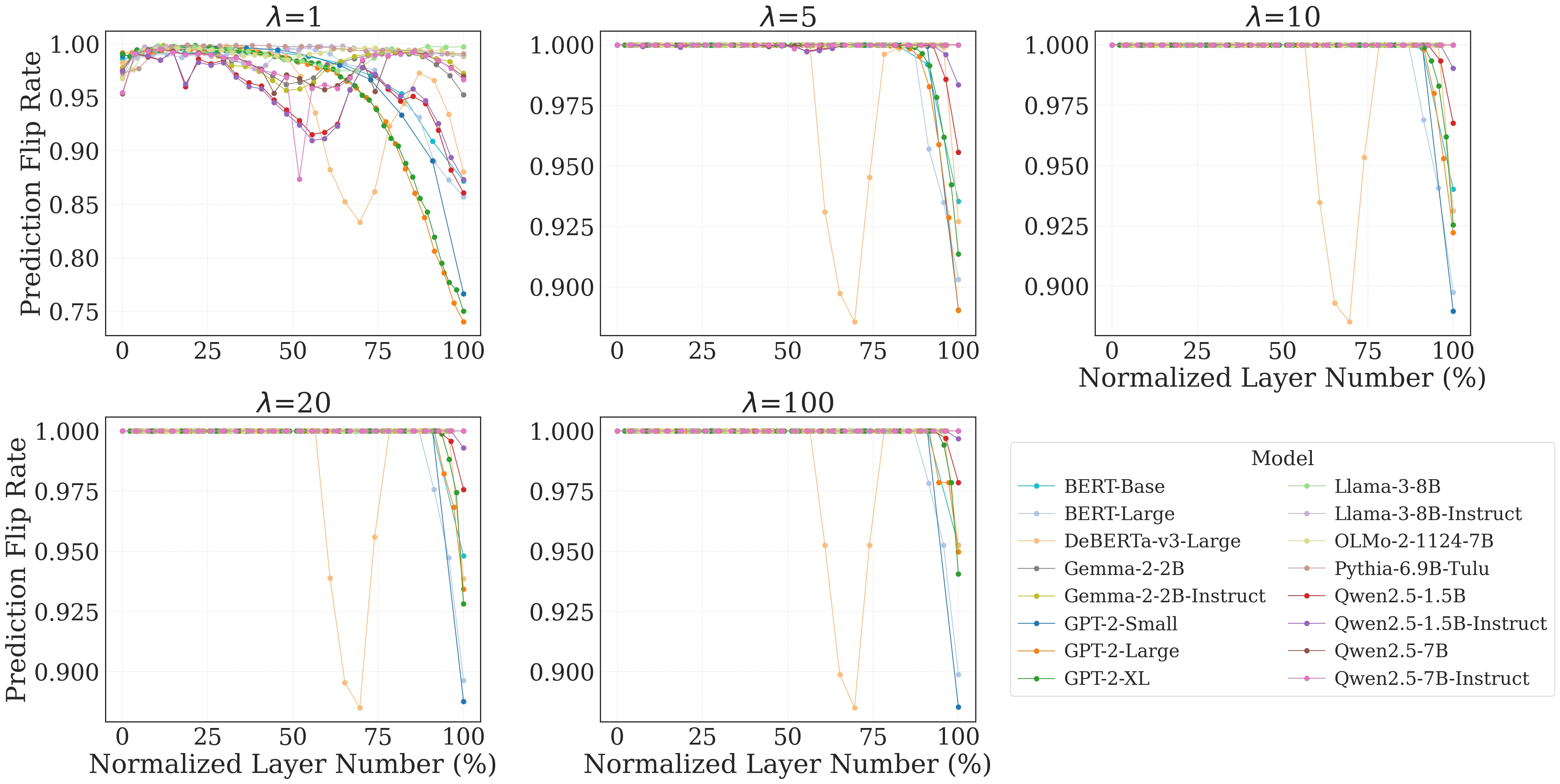}
  \caption{Prediction flip rate when applying steering vectors across different $\lambda$ values. The flip rate patterns mirror the probability change results, with most models maintaining high rates (0.98-1.00) throughout all layers. \texttt{DeBERTa-v3-Large} is an exception, showing a sudden drop around 75\% of model depth. The consistency across $\lambda$ values suggests that steering effectiveness depends more on model architecture than intervention strength.}
  \label{fig:steering_flip_rate}
\end{figure*}

\Cref{fig:steering_flip_rate} presents the corresponding prediction flip rates, which measure how often steering vectors successfully change the classifier's prediction. The patterns mirror the probability change results, with most models maintaining high flip rates (0.98--1.00) throughout all layers, except for \texttt{DeBERTa-v3-Large}, which exhibits a similar drop around 75\% of model depth.

\section{Classifier Error Analysis}
\label{sec:classifier_metrics_appendix}

We conducted a detailed error analysis of our classifiers to better understand their performance across different morphological features and languages. 

\paragraph{English Results}
\begin{itemize}[leftmargin=0.5cm, topsep=1pt, itemsep=1.5pt]
    \item \Cref{tab:inflection_breakdown_lr}: Inflection accuracy by morphological feature (linear probes)
    \item \Cref{tab:inflection_breakdown_mlp}: Inflection accuracy by morphological feature (MLP probes)
    \item \Cref{tab:lexeme_pos_breakdown_lr}: Lemma accuracy by part of speech (linear probes)
    \item \Cref{tab:lexeme_pos_breakdown_mlp}: Lemma accuracy by part of speech (MLP probes)
\end{itemize}

\paragraph{Chinese Results}
\begin{itemize}[leftmargin=0.5cm, topsep=1pt, itemsep=1.5pt]
    \item \Cref{tab:inflection_breakdown_combined_ud_zh_gsd_dataset}: Inflection accuracy by feature (linear and MLP)
    \item \Cref{tab:lexeme_pos_breakdown_lr_ud_zh_gsd_dataset}: Lemma accuracy by POS (linear probes)
    \item \Cref{tab:lexeme_pos_breakdown_mlp_ud_zh_gsd_dataset}: Lemma accuracy by POS (MLP probes)
\end{itemize}

\paragraph{German Results}
\begin{itemize}[leftmargin=0.5cm, topsep=1pt, itemsep=1.5pt]
    \item \Cref{tab:inflection_breakdown_lr_ud_de_gsd_dataset}: Inflection accuracy by feature (linear probes)
    \item \Cref{tab:inflection_breakdown_mlp_ud_de_gsd_dataset}: Inflection accuracy by feature (MLP probes)
    \item \Cref{tab:lexeme_pos_breakdown_combined_ud_de_gsd_dataset}: Lemma accuracy by POS (linear and MLP)
\end{itemize}

\paragraph{French Results}
\begin{itemize}[leftmargin=0.5cm, topsep=1pt, itemsep=1.5pt]
    \item \Cref{tab:inflection_breakdown_lr_ud_fr_gsd_dataset}: Inflection accuracy by feature (linear probes)
    \item \Cref{tab:inflection_breakdown_mlp_ud_fr_gsd_dataset}: Inflection accuracy by feature (MLP probes)
    \item \Cref{tab:lexeme_pos_breakdown_combined_ud_fr_gsd_dataset}: Lemma accuracy by POS (linear and MLP)
\end{itemize}

\paragraph{Russian Results}
\begin{itemize}[leftmargin=0.5cm, topsep=1pt, itemsep=1.5pt]
    \item \Cref{tab:inflection_breakdown_lr_ud_ru_syntagrus_dataset}: Inflection accuracy by feature (linear probes)
    \item \Cref{tab:inflection_breakdown_mlp_ud_ru_syntagrus_dataset}: Inflection accuracy by feature (MLP probes)
    \item \Cref{tab:lexeme_pos_breakdown_combined_ud_ru_syntagrus_dataset}: Lemma accuracy by POS (linear and MLP)
\end{itemize}

\paragraph{Turkish Results}
\begin{itemize}[leftmargin=0.5cm, topsep=1pt, itemsep=1.5pt]
    \item \Cref{tab:inflection_breakdown_lr_ud_tr_imst_dataset}: Inflection accuracy by feature (linear probes)
    \item \Cref{tab:inflection_breakdown_mlp_ud_tr_imst_dataset}: Inflection accuracy by feature (MLP probes)
    \item \Cref{tab:lexeme_pos_breakdown_combined_ud_tr_imst_dataset}: Lemma accuracy by POS (linear and MLP)
\end{itemize}

\begin{table*}[htbp]
\small
\centering
\renewcommand\arraystretch{1.2}
\setlength{\tabcolsep}{3pt}
\resizebox{\linewidth}{!}{%
\begin{tabular}{@{}lcccccccc@{}}
    \toprule
    \multirow{2}{*}{\textbf{Model}} & 3rd person & Base & Comparative & Past & Plural & Positive & Singular & Superlative \\
     & \footnotesize (n=249) & \footnotesize (n=1,833) & \footnotesize (n=76) & \footnotesize (n=1,003) & \footnotesize (n=1,247) & \footnotesize (n=1,785) & \footnotesize (n=3,587) & \footnotesize (n=52) \\
    \midrule
    \texttt{BERT-Base} & 0.960 & 0.965 & 0.817 & 0.967 & 0.983 & 0.946 & 0.971 & 0.759 \\
    \texttt{BERT-Large} & 0.956 & 0.964 & 0.861 & 0.968 & 0.982 & 0.950 & 0.971 & 0.768 \\
    \texttt{DeBERTa-v3-Large} & 0.938 & 0.974 & 0.831 & 0.961 & 0.986 & 0.954 & 0.977 & 0.706 \\
    \texttt{GPT-2-Small} & 0.828 & 0.958 & 0.840 & 0.956 & 0.974 & 0.941 & 0.964 & 0.754 \\
    \texttt{GPT-2-Large} & 0.812 & 0.958 & 0.826 & 0.951 & 0.975 & 0.936 & 0.967 & 0.792 \\
    \texttt{GPT-2-XL} & 0.817 & 0.959 & 0.813 & 0.948 & 0.977 & 0.940 & 0.968 & 0.788 \\
    \texttt{Pythia-6.9B} & 0.886 & 0.972 & 0.904 & 0.964 & 0.989 & 0.957 & 0.977 & 0.907 \\
    \texttt{Pythia-6.9B-Tulu} & 0.899 & 0.973 & 0.909 & 0.967 & 0.989 & 0.956 & 0.976 & 0.910 \\
    \texttt{OLMo-2-1124-7B} & 0.938 & 0.968 & 0.902 & 0.972 & 0.981 & 0.923 & 0.966 & 0.888 \\
    \texttt{OLMo-2-1124-7B-Instruct} & 0.927 & 0.967 & 0.896 & 0.971 & 0.981 & 0.923 & 0.965 & 0.872 \\
    \texttt{Gemma-2-2B} & 0.901 & 0.968 & 0.797 & 0.969 & 0.986 & 0.947 & 0.974 & 0.833 \\
    \texttt{Gemma-2-2B-Instruct} & 0.913 & 0.966 & 0.863 & 0.973 & 0.988 & 0.938 & 0.972 & 0.872 \\
    \texttt{Qwen2.5-1.5B} & 0.856 & 0.950 & 0.802 & 0.942 & 0.972 & 0.919 & 0.957 & 0.688 \\
    \texttt{Qwen2.5-1.5B-Instruct} & 0.774 & 0.954 & 0.647 & 0.945 & 0.972 & 0.921 & 0.965 & 0.630 \\
    \bottomrule
\end{tabular}
}%
\caption{Breakdown of inflection classification accuracy by morphological feature for each model using linear regression classifiers (English). Inflections are grouped by their morphological features (\eg Past, Plural, Comparative). For each group, the reported accuracy is the average of accuracies from classifiers trained at each model layer. All accuracy values are on a 0--1 scale. Comparative and superlative forms consistently show the lowest accuracy across all models, reflecting the challenges of these less frequent morphological categories.}
\label{tab:inflection_breakdown_lr}
\end{table*}

\begin{table*}[htbp]
\small
\centering
\renewcommand\arraystretch{1.2}
\setlength{\tabcolsep}{3pt}
\resizebox{\linewidth}{!}{%
\begin{tabular}{@{}lcccccccc@{}}
    \toprule
    \multirow{2}{*}{\textbf{Model}} & 3rd person & Base & Comparative & Past & Plural & Positive & Singular & Superlative \\
     & \footnotesize (n=249) & \footnotesize (n=1,833) & \footnotesize (n=76) & \footnotesize (n=1,003) & \footnotesize (n=1,247) & \footnotesize (n=1,785) & \footnotesize (n=3,587) & \footnotesize (n=52) \\
    \midrule
    \texttt{BERT-Base} & 0.973 & 0.969 & 0.910 & 0.972 & 0.989 & 0.959 & 0.974 & 0.939 \\
    \texttt{BERT-Large} & 0.967 & 0.970 & 0.910 & 0.973 & 0.988 & 0.961 & 0.975 & 0.931 \\
    \texttt{DeBERTa-v3-Large} & 0.954 & 0.976 & 0.925 & 0.966 & 0.989 & 0.962 & 0.979 & 0.867 \\
    \texttt{GPT-2-Small} & 0.921 & 0.963 & 0.928 & 0.952 & 0.972 & 0.930 & 0.963 & 0.870 \\
    \texttt{GPT-2-Large} & 0.857 & 0.962 & 0.872 & 0.955 & 0.976 & 0.942 & 0.967 & 0.854 \\
    \texttt{GPT-2-XL} & 0.921 & 0.963 & 0.928 & 0.952 & 0.972 & 0.930 & 0.963 & 0.870 \\
    \texttt{Pythia-6.9B} & 0.932 & 0.972 & 0.921 & 0.961 & 0.982 & 0.949 & 0.971 & 0.886 \\
    \texttt{Pythia-6.9B-Tulu} & 0.948 & 0.974 & 0.932 & 0.964 & 0.983 & 0.949 & 0.971 & 0.897 \\
    \texttt{OLMo-2-1124-7B} & 0.957 & 0.968 & 0.926 & 0.966 & 0.989 & 0.949 & 0.973 & 0.905 \\
    \texttt{OLMo-2-1124-7B-Instruct} & 0.939 & 0.967 & 0.903 & 0.967 & 0.988 & 0.949 & 0.973 & 0.873 \\
    \texttt{Gemma-2-2B} & 0.913 & 0.967 & 0.863 & 0.968 & 0.990 & 0.950 & 0.976 & 0.907 \\
    \texttt{Gemma-2-2B-Instruct} & 0.930 & 0.970 & 0.878 & 0.975 & 0.989 & 0.946 & 0.974 & 0.906 \\
    \texttt{Qwen2.5-1.5B} & 0.882 & 0.948 & 0.822 & 0.943 & 0.974 & 0.927 & 0.957 & 0.736 \\
    \texttt{Qwen2.5-1.5B-Instruct} & 0.808 & 0.953 & 0.697 & 0.947 & 0.974 & 0.930 & 0.965 & 0.682 \\
    \bottomrule
\end{tabular}
}%
\caption{Breakdown of inflection classification accuracy by morphological feature for each model using Multi-Layer Perceptron (MLP) classifiers (English). Inflections are grouped by their morphological features (\eg Past, Plural, Comparative). For each group, the reported accuracy is the average of accuracies from classifiers trained at each model layer. All accuracy values are on a 0--1 scale. MLP classifiers provide modest improvements over linear regression, particularly for comparative and superlative forms, though the relative ordering across morphological features remains consistent.}
\label{tab:inflection_breakdown_mlp}
\end{table*}

\begin{table*}[htbp]
\small
\centering
\renewcommand\arraystretch{1.2}
\setlength{\tabcolsep}{3pt}
\resizebox{\linewidth}{!}{%
\begin{tabular}{@{}lccccccccc@{}}
    \toprule
    \multirow{2}{*}{\textbf{Model}} & Noun & Verb & Adjective & Adverb & Pronoun & Preposition & Conjunction & Interjection & Other \\
     & \footnotesize (n=1,739) & \footnotesize (n=641) & \footnotesize (n=641) & \footnotesize (n=23) & \footnotesize (n=1) & \footnotesize (n=1) & \footnotesize (n=1) & \footnotesize (n=1) & \footnotesize (n=9) \\
    \midrule
    \texttt{BERT-Base} & 0.636 & 0.737 & 0.609 & 0.805 & 0.292 & 0.000 & 0.585 & 0.000 & 0.902 \\
    \texttt{BERT-Large} & 0.684 & 0.777 & 0.653 & 0.826 & 0.580 & 0.154 & 0.662 & 0.065 & 0.897 \\
    \texttt{DeBERTa-v3-Large} & 0.592 & 0.737 & 0.585 & 0.723 & 0.440 & 0.077 & 0.438 & 0.081 & 0.866 \\
    \texttt{GPT-2-Small} & 0.631 & 0.789 & 0.612 & 0.813 & 0.542 & 0.000 & 0.415 & 0.033 & 0.896 \\
    \texttt{GPT-2-Large} & 0.691 & 0.810 & 0.688 & 0.847 & 0.853 & 0.174 & 0.267 & 0.115 & 0.912 \\
    \texttt{GPT-2-XL} & 0.713 & 0.827 & 0.708 & 0.847 & 0.724 & 0.222 & 0.311 & 0.241 & 0.899 \\
    \texttt{Pythia-6.9B} & 0.856 & 0.926 & 0.836 & 0.926 & 0.938 & 0.443 & 0.566 & 0.488 & 0.934 \\
    \texttt{Pythia-6.9B-Tulu} & 0.864 & 0.930 & 0.843 & 0.930 & 0.923 & 0.514 & 0.651 & 0.476 & 0.936 \\
    \texttt{OLMo-2-1124-7B} & 0.798 & 0.875 & 0.794 & 0.913 & 0.697 & 0.339 & 0.363 & 0.495 & 0.913 \\
    \texttt{OLMo-2-1124-7B-Instruct} & 0.798 & 0.868 & 0.792 & 0.902 & 0.606 & 0.339 & 0.331 & 0.495 & 0.910 \\
    \texttt{Gemma-2-2B} & 0.757 & 0.869 & 0.736 & 0.876 & 0.667 & 0.179 & 0.205 & 0.288 & 0.891 \\
    \texttt{Gemma-2-2B-Instruct} & 0.749 & 0.844 & 0.742 & 0.872 & 0.620 & 0.137 & 0.152 & 0.247 & 0.912 \\
    \texttt{Qwen2.5-1.5B} & 0.652 & 0.801 & 0.650 & 0.828 & 0.526 & 0.082 & 0.223 & 0.068 & 0.867 \\
    \texttt{Qwen2.5-1.5B-Instruct} & 0.642 & 0.800 & 0.632 & 0.831 & 0.544 & 0.082 & 0.245 & 0.068 & 0.877 \\
    \texttt{Llama-3.1-8B} & 0.776 & 0.882 & 0.771 & 0.887 & 0.831 & 0.286 & 0.396 & 0.321 & 0.911 \\
    \texttt{Llama-3.1-8B-Instruct} & 0.796 & 0.892 & 0.788 & 0.896 & 0.908 & 0.300 & 0.443 & 0.357 & 0.917 \\
    \bottomrule
\end{tabular}
}%
\caption{Breakdown of lemma classification accuracy by Part of Speech (POS) for each model using linear regression classifiers (English). Lemmas are grouped by their POS tags (\eg Noun, Verb, Adjective). For each group, the reported accuracy is the average of accuracies from classifiers trained at each model layer. All accuracy values are on a 0--1 scale. Performance varies significantly with frequency: frequent categories like nouns and verbs achieve higher accuracy, while infrequent categories like pronouns and prepositions show lower performance due to limited training examples.}
\label{tab:lexeme_pos_breakdown_lr}
\end{table*}

\begin{table*}[htbp]
\small
\centering
\renewcommand\arraystretch{1.2}
\setlength{\tabcolsep}{3pt}
\resizebox{\linewidth}{!}{%
\begin{tabular}{@{}lccccccccc@{}}
    \toprule
    \multirow{2}{*}{\textbf{Model}} & Noun & Verb & Adjective & Adverb & Pronoun & Preposition & Conjunction & Interjection & Other \\
     & \footnotesize (n=1,739) & \footnotesize (n=641) & \footnotesize (n=641) & \footnotesize (n=23) & \footnotesize (n=1) & \footnotesize (n=1) & \footnotesize (n=1) & \footnotesize (n=1) & \footnotesize (n=9) \\
    \midrule
    \texttt{BERT-Base} & 0.775 & 0.831 & 0.748 & 0.873 & 0.458 & 0.125 & 0.756 & 0.267 & 0.898 \\
    \texttt{BERT-Large} & 0.813 & 0.863 & 0.785 & 0.884 & 0.540 & 0.231 & 0.725 & 0.323 & 0.897 \\
    \texttt{DeBERTa-v3-Large} & 0.689 & 0.803 & 0.682 & 0.802 & 0.700 & 0.115 & 0.662 & 0.242 & 0.861 \\
    \texttt{GPT-2-Small} & 0.678 & 0.792 & 0.665 & 0.765 & 0.042 & 0.000 & 0.610 & 0.000 & 0.830 \\
    \texttt{GPT-2-Large} & 0.754 & 0.837 & 0.755 & 0.827 & 0.347 & 0.188 & 0.596 & 0.385 & 0.871 \\
    \texttt{GPT-2-XL} & 0.774 & 0.844 & 0.771 & 0.827 & 0.561 & 0.232 & 0.561 & 0.431 & 0.860 \\
    \texttt{Pythia-6.9B} & 0.774 & 0.856 & 0.768 & 0.862 & 0.554 & 0.229 & 0.528 & 0.310 & 0.868 \\
    \texttt{Pythia-6.9B-Tulu} & 0.818 & 0.880 & 0.803 & 0.887 & 0.554 & 0.343 & 0.613 & 0.381 & 0.889 \\
    \texttt{OLMo-2-1124-7B} & 0.818 & 0.877 & 0.828 & 0.896 & 0.727 & 0.290 & 0.734 & 0.505 & 0.885 \\
    \texttt{OLMo-2-1124-7B-Instruct} & 0.822 & 0.874 & 0.829 & 0.897 & 0.667 & 0.306 & 0.750 & 0.473 & 0.886 \\
    \texttt{Gemma-2-2B} & 0.763 & 0.860 & 0.763 & 0.881 & 0.574 & 0.125 & 0.443 & 0.182 & 0.880 \\
    \texttt{Gemma-2-2B-Instruct} & 0.777 & 0.846 & 0.785 & 0.882 & 0.580 & 0.137 & 0.400 & 0.299 & 0.875 \\
    \texttt{Qwen2.5-1.5B} & 0.747 & 0.838 & 0.742 & 0.811 & 0.228 & 0.131 & 0.628 & 0.164 & 0.857 \\
    \texttt{Qwen2.5-1.5B-Instruct} & 0.749 & 0.840 & 0.738 & 0.818 & 0.211 & 0.098 & 0.564 & 0.123 & 0.860 \\
    \texttt{Llama-3.1-8B} & 0.798 & 0.879 & 0.807 & 0.886 & 0.800 & 0.214 & 0.679 & 0.393 & 0.882 \\
    \texttt{Llama-3.1-8B-Instruct} & 0.824 & 0.893 & 0.826 & 0.895 & 0.831 & 0.257 & 0.689 & 0.429 & 0.887 \\
    \bottomrule
\end{tabular}
}%
\caption{Breakdown of lemma classification accuracy by Part of Speech (POS) for each model using Multi-Layer Perceptron (MLP) classifiers (English). Lemmas are grouped by their POS tags (\eg Noun, Verb, Adjective). For each group, the reported accuracy is the average of accuracies from classifiers trained at each model layer. All accuracy values are on a 0--1 scale. MLP classifiers provide consistent improvements over linear regression across all POS categories, though the frequency-dependent performance patterns persist.}
\label{tab:lexeme_pos_breakdown_mlp}
\end{table*}

\begin{table*}[htbp]
\small
\centering
\renewcommand\arraystretch{1.2}
\setlength{\tabcolsep}{3pt}
\begin{minipage}{\linewidth}\centering
\begin{tabular}{@{}lcccccccc@{}}
    \toprule
    \multirow{3}{*}{\textbf{Model}} & \multicolumn{4}{c}{\textbf{Linear Regression}} & \multicolumn{4}{c}{\textbf{MLP}} \\
    \cmidrule(lr){2-5} \cmidrule(lr){6-9}
     & Positive & Base & Plural & Singular & Positive & Base & Plural & Singular \\
     & \footnotesize (n=300) & \footnotesize (n=2,074) & \footnotesize (n=3) & \footnotesize (n=3,947) & \footnotesize (n=300) & \footnotesize (n=2,074) & \footnotesize (n=3) & \footnotesize (n=3,947) \\
    \midrule
    mT5-Base & 0.739 & 0.913 & 0.436 & 0.962 & 0.783 & 0.919 & 0.231 & 0.961 \\
    Qwen2.5-1.5B & 0.785 & 0.929 & 0.034 & 0.969 & 0.801 & 0.924 & 0.092 & 0.967 \\
    Qwen2.5-1.5B-Instruct & 0.779 & 0.925 & 0.034 & 0.964 & 0.803 & 0.923 & 0.057 & 0.967 \\
    Qwen2.5-7B & 0.824 & 0.937 & 0.310 & 0.970 & 0.828 & 0.929 & 0.310 & 0.969 \\
    Qwen2.5-7B-Instruct & 0.819 & 0.936 & 0.299 & 0.970 & 0.823 & 0.928 & 0.276 & 0.969 \\
    Goldfish Chinese & 0.793 & 0.912 & 0.000 & 0.958 & 0.816 & 0.915 & 0.000 & 0.957 \\
    \bottomrule
\end{tabular}
\caption{Breakdown of inflection classification accuracy for each model by inflection type using Linear Regression and Multi-Layer Perceptron (MLP) classifiers (Chinese). Accuracies are calculated over all examples for a given group across all layers. Counts (n) are derived from a single representative layer for each group. All accuracy values are on a 0--1 scale.}
\label{tab:inflection_breakdown_combined_ud_zh_gsd_dataset}
\end{minipage}
\end{table*}

\begin{table*}[htbp]
\small
\centering
\renewcommand\arraystretch{1.2}
\setlength{\tabcolsep}{3pt}
\begin{minipage}{\linewidth}\centering
\begin{tabular}{@{}lcccccc@{}}
    \toprule
    \multirow{2}{*}{\textbf{Model}} & Noun & Verb & Adjective & Adverb & Preposition & Other \\
     & \footnotesize (n=1,179) & \footnotesize (n=564) & \footnotesize (n=108) & \footnotesize (n=22) & \footnotesize (n=20) & \footnotesize (n=50) \\
    \midrule
    mT5-Base & 0.838 & 0.828 & 0.786 & 0.762 & 0.920 & 0.726 \\
    Qwen2.5-1.5B & 0.810 & 0.797 & 0.746 & 0.715 & 0.872 & 0.699 \\
    Qwen2.5-1.5B-Instruct & 0.813 & 0.799 & 0.748 & 0.713 & 0.873 & 0.700 \\
    Qwen2.5-7B & 0.887 & 0.882 & 0.846 & 0.847 & 0.915 & 0.817 \\
    Qwen2.5-7B-Instruct & 0.886 & 0.877 & 0.843 & 0.835 & 0.913 & 0.811 \\
    Goldfish Chinese & 0.883 & 0.878 & 0.845 & 0.875 & 0.954 & 0.858 \\
    \bottomrule
\end{tabular}
\caption{Breakdown of lemma classification accuracy by Part of Speech (POS) for each model, using Linear Regression classifiers (Chinese). Lemmas are grouped by their POS tags (\eg Noun, Verb, Adjective). Accuracies are calculated over all examples for a given group across all layers. Counts (n) are derived from a single representative layer for each group. All accuracy values are on a 0--1 scale.}
\label{tab:lexeme_pos_breakdown_lr_ud_zh_gsd_dataset}
\end{minipage}
\end{table*}

\begin{table*}[htbp]
\small
\centering
\renewcommand\arraystretch{1.2}
\setlength{\tabcolsep}{3pt}
\begin{minipage}{\linewidth}\centering
\begin{tabular}{@{}lcccccc@{}}
    \toprule
    \multirow{2}{*}{\textbf{Model}} & Noun & Verb & Adjective & Adverb & Preposition & Other \\
     & \footnotesize (n=1,179) & \footnotesize (n=564) & \footnotesize (n=108) & \footnotesize (n=22) & \footnotesize (n=20) & \footnotesize (n=50) \\
    \midrule
    mT5-Base & 0.698 & 0.712 & 0.564 & 0.571 & 0.884 & 0.569 \\
    Qwen2.5-1.5B & 0.748 & 0.761 & 0.658 & 0.668 & 0.826 & 0.669 \\
    Qwen2.5-1.5B-Instruct & 0.735 & 0.745 & 0.643 & 0.643 & 0.814 & 0.655 \\
    Qwen2.5-7B & 0.815 & 0.826 & 0.749 & 0.745 & 0.848 & 0.750 \\
    Qwen2.5-7B-Instruct & 0.815 & 0.822 & 0.747 & 0.734 & 0.845 & 0.744 \\
    Goldfish Chinese & 0.766 & 0.771 & 0.647 & 0.621 & 0.912 & 0.682 \\
    \bottomrule
\end{tabular}
\caption{Breakdown of lemma classification accuracy by Part of Speech (POS) for each model, using Multi-Layer Perceptron (MLP) classifiers (Chinese). Lemmas are grouped by their POS tags (\eg Noun, Verb, Adjective). Accuracies are calculated over all examples for a given group across all layers. Counts (n) are derived from a single representative layer for each group. All accuracy values are on a 0--1 scale.}
\label{tab:lexeme_pos_breakdown_mlp_ud_zh_gsd_dataset}
\end{minipage}
\end{table*}

\begin{table*}[htbp]
\small
\centering
\renewcommand\arraystretch{1.2}
\setlength{\tabcolsep}{3pt}
\begin{minipage}{\linewidth}\centering
\begin{tabular}{@{}lcccccccc@{}}
    \toprule
    \multirow{2}{*}{\textbf{Model}} & Base & 3rd person & Positive & Past & Plural & Superlative & Singular & Comparative \\
     & \footnotesize (n=417) & \footnotesize (n=517) & \footnotesize (n=1,720) & \footnotesize (n=839) & \footnotesize (n=1,076) & \footnotesize (n=52) & \footnotesize (n=3,197) & \footnotesize (n=141) \\
    \midrule
    mT5-Base & 0.908 & 0.941 & 0.940 & 0.960 & 0.882 & 0.572 & 0.962 & 0.636 \\
    Qwen2.5-1.5B & 0.849 & 0.889 & 0.922 & 0.914 & 0.888 & 0.657 & 0.953 & 0.796 \\
    Qwen2.5-1.5B-Instruct & 0.844 & 0.887 & 0.922 & 0.910 & 0.889 & 0.659 & 0.952 & 0.795 \\
    Qwen2.5-7B & 0.892 & 0.922 & 0.939 & 0.947 & 0.909 & 0.826 & 0.962 & 0.878 \\
    Qwen2.5-7B-Instruct & 0.915 & 0.934 & 0.945 & 0.962 & 0.924 & 0.866 & 0.968 & 0.909 \\
    Goldfish German & 0.938 & 0.941 & 0.955 & 0.979 & 0.916 & 0.542 & 0.968 & 0.708 \\
    \bottomrule
\end{tabular}
\caption{Breakdown of inflection classification accuracy for each model by inflection type using Linear Regression classifiers (German). Accuracies are calculated over all examples for a given group across all layers. Counts (n) are derived from a single representative layer for each group. All accuracy values are on a 0--1 scale.}
\label{tab:inflection_breakdown_lr_ud_de_gsd_dataset}
\end{minipage}
\end{table*}

\begin{table*}[htbp]
\small
\centering
\renewcommand\arraystretch{1.2}
\setlength{\tabcolsep}{3pt}
\begin{minipage}{\linewidth}\centering
\begin{tabular}{@{}lcccccccc@{}}
    \toprule
    \multirow{2}{*}{\textbf{Model}} & Base & 3rd person & Positive & Past & Plural & Superlative & Singular & Comparative \\
     & \footnotesize (n=417) & \footnotesize (n=517) & \footnotesize (n=1,720) & \footnotesize (n=839) & \footnotesize (n=1,076) & \footnotesize (n=52) & \footnotesize (n=3,197) & \footnotesize (n=141) \\
    \midrule
    mT5-Base & 0.921 & 0.945 & 0.948 & 0.959 & 0.884 & 0.723 & 0.967 & 0.770 \\
    Qwen2.5-1.5B & 0.890 & 0.915 & 0.930 & 0.940 & 0.897 & 0.831 & 0.958 & 0.892 \\
    Qwen2.5-1.5B-Instruct & 0.888 & 0.914 & 0.930 & 0.938 & 0.898 & 0.825 & 0.957 & 0.897 \\
    Qwen2.5-7B & 0.912 & 0.932 & 0.944 & 0.956 & 0.913 & 0.868 & 0.964 & 0.924 \\
    Qwen2.5-7B-Instruct & 0.925 & 0.941 & 0.950 & 0.966 & 0.928 & 0.901 & 0.970 & 0.936 \\
    Goldfish German & 0.947 & 0.957 & 0.964 & 0.978 & 0.923 & 0.817 & 0.970 & 0.896 \\
    \bottomrule
\end{tabular}
\caption{Breakdown of inflection classification accuracy for each model by inflection type using Multi-Layer Perceptron (MLP) classifiers (German). Accuracies are calculated over all examples for a given group across all layers. Counts (n) are derived from a single representative layer for each group. All accuracy values are on a 0--1 scale.}
\label{tab:inflection_breakdown_mlp_ud_de_gsd_dataset}
\end{minipage}
\end{table*}

\begin{table*}[htbp]
\small
\centering
\renewcommand\arraystretch{1.2}
\setlength{\tabcolsep}{3pt}
\begin{minipage}{\linewidth}\centering
\begin{tabular}{@{}lcccccccc@{}}
    \toprule
    \multirow{3}{*}{\textbf{Model}} & \multicolumn{4}{c}{\textbf{Linear Regression}} & \multicolumn{4}{c}{\textbf{MLP}} \\
    \cmidrule(lr){2-5} \cmidrule(lr){6-9}
     & Noun & Verb & Adjective & Other & Noun & Verb & Adjective & Other \\
     & \footnotesize (n=1,262) & \footnotesize (n=395) & \footnotesize (n=406) & \footnotesize (n=12) & \footnotesize (n=1,262) & \footnotesize (n=395) & \footnotesize (n=406) & \footnotesize (n=12) \\
    \midrule
    mT5-Base & 0.685 & 0.662 & 0.568 & 0.750 & 0.611 & 0.602 & 0.486 & 0.723 \\
    Qwen2.5-1.5B & 0.743 & 0.725 & 0.715 & 0.775 & 0.721 & 0.700 & 0.687 & 0.711 \\
    Qwen2.5-1.5B-Instruct & 0.740 & 0.722 & 0.715 & 0.766 & 0.722 & 0.698 & 0.687 & 0.704 \\
    Qwen2.5-7B & 0.821 & 0.809 & 0.808 & 0.829 & 0.795 & 0.786 & 0.783 & 0.814 \\
    Qwen2.5-7B-Instruct & 0.815 & 0.803 & 0.803 & 0.821 & 0.795 & 0.785 & 0.782 & 0.813 \\
    Goldfish German & 0.720 & 0.747 & 0.701 & 0.769 & 0.758 & 0.772 & 0.742 & 0.769 \\
    \bottomrule
\end{tabular}
\caption{Breakdown of lemma classification accuracy by Part of Speech (POS) for each model, using Linear Regression and Multi-Layer Perceptron (MLP) classifiers (German). Lemmas are grouped by their POS tags (\eg Noun, Verb, Adjective). Accuracies are calculated over all examples for a given group across all layers. Counts (n) are derived from a single representative layer for each group. All accuracy values are on a 0--1 scale.}
\label{tab:lexeme_pos_breakdown_combined_ud_de_gsd_dataset}
\end{minipage}
\end{table*}

\begin{table*}[htbp]
\small
\centering
\renewcommand\arraystretch{1.2}
\setlength{\tabcolsep}{3pt}
\begin{minipage}{\linewidth}\centering
\begin{tabular}{@{}lcccccc@{}}
    \toprule
    \multirow{2}{*}{\textbf{Model}} & Base & 3rd person & Positive & Past & Plural & Singular \\
     & \footnotesize (n=688) & \footnotesize (n=776) & \footnotesize (n=1,833) & \footnotesize (n=857) & \footnotesize (n=1,457) & \footnotesize (n=5,169) \\
    \midrule
    mT5-Base & 0.934 & 0.912 & 0.879 & 0.908 & 0.954 & 0.970 \\
    Qwen2.5-1.5B & 0.933 & 0.858 & 0.896 & 0.903 & 0.958 & 0.967 \\
    Qwen2.5-1.5B-Instruct & 0.930 & 0.852 & 0.893 & 0.898 & 0.958 & 0.966 \\
    Qwen2.5-7B & 0.955 & 0.918 & 0.918 & 0.931 & 0.965 & 0.975 \\
    Qwen2.5-7B-Instruct & 0.951 & 0.913 & 0.915 & 0.928 & 0.964 & 0.974 \\
    Goldfish French & 0.942 & 0.955 & 0.937 & 0.930 & 0.968 & 0.976 \\
    \bottomrule
\end{tabular}
\caption{Breakdown of inflection classification accuracy for each model by inflection type using Linear Regression classifiers (French). Accuracies are calculated over all examples for a given group across all layers. Counts (n) are derived from a single representative layer for each group. All accuracy values are on a 0--1 scale.}
\label{tab:inflection_breakdown_lr_ud_fr_gsd_dataset}
\end{minipage}
\end{table*}

\begin{table*}[htbp]
\small
\centering
\renewcommand\arraystretch{1.2}
\setlength{\tabcolsep}{3pt}
\begin{minipage}{\linewidth}\centering
\begin{tabular}{@{}lcccccc@{}}
    \toprule
    \multirow{2}{*}{\textbf{Model}} & Base & 3rd person & Positive & Past & Plural & Singular \\
     & \footnotesize (n=688) & \footnotesize (n=776) & \footnotesize (n=1,833) & \footnotesize (n=857) & \footnotesize (n=1,457) & \footnotesize (n=5,169) \\
    \midrule
    mT5-Base & 0.957 & 0.937 & 0.911 & 0.935 & 0.957 & 0.977 \\
    Qwen2.5-1.5B & 0.954 & 0.905 & 0.914 & 0.925 & 0.965 & 0.968 \\
    Qwen2.5-1.5B-Instruct & 0.954 & 0.902 & 0.911 & 0.924 & 0.965 & 0.968 \\
    Qwen2.5-7B & 0.966 & 0.936 & 0.930 & 0.937 & 0.970 & 0.976 \\
    Qwen2.5-7B-Instruct & 0.962 & 0.931 & 0.926 & 0.934 & 0.970 & 0.975 \\
    Goldfish French & 0.974 & 0.967 & 0.945 & 0.942 & 0.973 & 0.979 \\
    \bottomrule
\end{tabular}
\caption{Breakdown of inflection classification accuracy for each model by inflection type using Multi-Layer Perceptron (MLP) classifiers (French). Accuracies are calculated over all examples for a given group across all layers. Counts (n) are derived from a single representative layer for each group. All accuracy values are on a 0--1 scale.}
\label{tab:inflection_breakdown_mlp_ud_fr_gsd_dataset}
\end{minipage}
\end{table*}

\begin{table*}[htbp]
\small
\centering
\renewcommand\arraystretch{1.2}
\setlength{\tabcolsep}{3pt}
\begin{minipage}{\linewidth}\centering
\begin{tabular}{@{}lcccccccc@{}}
    \toprule
    \multirow{3}{*}{\textbf{Model}} & \multicolumn{4}{c}{\textbf{Linear Regression}} & \multicolumn{4}{c}{\textbf{MLP}} \\
    \cmidrule(lr){2-5} \cmidrule(lr){6-9}
     & Noun & Verb & Adjective & Other & Noun & Verb & Adjective & Other \\
     & \footnotesize (n=1,496) & \footnotesize (n=406) & \footnotesize (n=358) & \footnotesize (n=15) & \footnotesize (n=1,496) & \footnotesize (n=406) & \footnotesize (n=358) & \footnotesize (n=15) \\
    \midrule
    mT5-Base & 0.708 & 0.577 & 0.605 & 0.799 & 0.755 & 0.560 & 0.636 & 0.820 \\
    Qwen2.5-1.5B & 0.754 & 0.725 & 0.673 & 0.824 & 0.807 & 0.765 & 0.751 & 0.853 \\
    Qwen2.5-1.5B-Instruct & 0.750 & 0.718 & 0.671 & 0.820 & 0.824 & 0.776 & 0.768 & 0.869 \\
    Qwen2.5-7B & 0.840 & 0.814 & 0.764 & 0.869 & 0.856 & 0.825 & 0.794 & 0.884 \\
    Qwen2.5-7B-Instruct & 0.833 & 0.805 & 0.758 & 0.860 & 0.851 & 0.818 & 0.792 & 0.883 \\
    Goldfish French & 0.749 & 0.758 & 0.661 & 0.811 & 0.894 & 0.869 & 0.813 & 0.888 \\
    \bottomrule
\end{tabular}
\caption{Breakdown of lemma classification accuracy by Part of Speech (POS) for each model, using Linear Regression and Multi-Layer Perceptron (MLP) classifiers (French). Lemmas are grouped by their POS tags (\eg Noun, Verb, Adjective). Accuracies are calculated over all examples for a given group across all layers. Counts (n) are derived from a single representative layer for each group. All accuracy values are on a 0--1 scale.}
\label{tab:lexeme_pos_breakdown_combined_ud_fr_gsd_dataset}
\end{minipage}
\end{table*}

\begin{table*}[htbp]
\small
\centering
\renewcommand\arraystretch{1.2}
\setlength{\tabcolsep}{3pt}
\begin{minipage}{\linewidth}\centering
\begin{tabular}{@{}lcccccccc@{}}
    \toprule
    \multirow{2}{*}{\textbf{Model}} & Base & 3rd person & Positive & Past & Plural & Superlative & Singular & Comparative \\
     & \footnotesize (n=690) & \footnotesize (n=456) & \footnotesize (n=1,192) & \footnotesize (n=455) & \footnotesize (n=1,333) & \footnotesize (n=3) & \footnotesize (n=3,316) & \footnotesize (n=23) \\
    \midrule
    mT5-Base & 0.930 & 0.978 & 0.975 & 0.957 & 0.877 & 0.000 & 0.977 & 0.799 \\
    Qwen2.5-1.5B & 0.925 & 0.946 & 0.974 & 0.938 & 0.923 & 0.015 & 0.966 & 0.835 \\
    Qwen2.5-1.5B-Instruct & 0.924 & 0.943 & 0.974 & 0.934 & 0.921 & 0.015 & 0.966 & 0.817 \\
    Qwen2.5-7B & 0.949 & 0.966 & 0.979 & 0.958 & 0.948 & 0.094 & 0.977 & 0.872 \\
    Qwen2.5-7B-Instruct & 0.951 & 0.974 & 0.980 & 0.970 & 0.948 & 0.080 & 0.980 & 0.918 \\
    Goldfish Russian & 0.940 & 0.950 & 0.976 & 0.931 & 0.921 & 0.000 & 0.976 & 0.867 \\
    \bottomrule
\end{tabular}
\caption{Breakdown of inflection classification accuracy for each model by inflection type using Linear Regression classifiers (Russian). Accuracies are calculated over all examples for a given group across all layers. Counts (n) are derived from a single representative layer for each group. All accuracy values are on a 0--1 scale.}
\label{tab:inflection_breakdown_lr_ud_ru_syntagrus_dataset}
\end{minipage}
\end{table*}

\begin{table*}[htbp]
\small
\centering
\renewcommand\arraystretch{1.2}
\setlength{\tabcolsep}{3pt}
\begin{minipage}{\linewidth}\centering
\begin{tabular}{@{}lcccccccc@{}}
    \toprule
    \multirow{2}{*}{\textbf{Model}} & Base & 3rd person & Positive & Past & Plural & Superlative & Singular & Comparative \\
     & \footnotesize (n=690) & \footnotesize (n=456) & \footnotesize (n=1,192) & \footnotesize (n=455) & \footnotesize (n=1,333) & \footnotesize (n=3) & \footnotesize (n=3,316) & \footnotesize (n=23) \\
    \midrule
    mT5-Base & 0.959 & 0.978 & 0.969 & 0.966 & 0.904 & 0.000 & 0.978 & 0.849 \\
    Qwen2.5-1.5B & 0.952 & 0.955 & 0.972 & 0.948 & 0.933 & 0.089 & 0.970 & 0.899 \\
    Qwen2.5-1.5B-Instruct & 0.950 & 0.954 & 0.973 & 0.947 & 0.933 & 0.089 & 0.969 & 0.911 \\
    Qwen2.5-7B & 0.963 & 0.964 & 0.978 & 0.960 & 0.951 & 0.246 & 0.979 & 0.910 \\
    Qwen2.5-7B-Instruct & 0.961 & 0.970 & 0.978 & 0.966 & 0.949 & 0.126 & 0.980 & 0.924 \\
    Goldfish Russian & 0.965 & 0.972 & 0.978 & 0.948 & 0.943 & 0.000 & 0.977 & 0.934 \\
    \bottomrule
\end{tabular}
\caption{Breakdown of inflection classification accuracy for each model by inflection type using Multi-Layer Perceptron (MLP) classifiers (Russian). Accuracies are calculated over all examples for a given group across all layers. Counts (n) are derived from a single representative layer for each group. All accuracy values are on a 0--1 scale.}
\label{tab:inflection_breakdown_mlp_ud_ru_syntagrus_dataset}
\end{minipage}
\end{table*}

\begin{table*}[htbp]
\small
\centering
\renewcommand\arraystretch{1.2}
\setlength{\tabcolsep}{3pt}
\begin{minipage}{\linewidth}\centering
\begin{tabular}{@{}lcccccccc@{}}
    \toprule
    \multirow{3}{*}{\textbf{Model}} & \multicolumn{4}{c}{\textbf{Linear Regression}} & \multicolumn{4}{c}{\textbf{MLP}} \\
    \cmidrule(lr){2-5} \cmidrule(lr){6-9}
     & Noun & Verb & Adjective & Other & Noun & Verb & Adjective & Other \\
     & \footnotesize (n=982) & \footnotesize (n=333) & \footnotesize (n=275) & \footnotesize (n=4) & \footnotesize (n=982) & \footnotesize (n=333) & \footnotesize (n=275) & \footnotesize (n=4) \\
    \midrule
    mT5-Base & 0.660 & 0.614 & 0.542 & 0.648 & 0.492 & 0.484 & 0.387 & 0.426 \\
    Qwen2.5-1.5B & 0.777 & 0.712 & 0.759 & 0.720 & 0.712 & 0.696 & 0.716 & 0.647 \\
    Qwen2.5-1.5B-Instruct & 0.772 & 0.704 & 0.756 & 0.720 & 0.710 & 0.689 & 0.717 & 0.643 \\
    Qwen2.5-7B & 0.854 & 0.790 & 0.843 & 0.812 & 0.798 & 0.794 & 0.813 & 0.749 \\
    Qwen2.5-7B-Instruct & 0.845 & 0.778 & 0.835 & 0.807 & 0.794 & 0.785 & 0.809 & 0.744 \\
    Goldfish Russian & 0.795 & 0.723 & 0.764 & 0.676 & 0.810 & 0.776 & 0.759 & 0.657 \\
    \bottomrule
\end{tabular}
\caption{Breakdown of lemma classification accuracy by Part of Speech (POS) for each model, using Linear Regression and Multi-Layer Perceptron (MLP) classifiers (Russian). Lemmas are grouped by their POS tags (\eg Noun, Verb, Adjective). Accuracies are calculated over all examples for a given group across all layers. Counts (n) are derived from a single representative layer for each group. All accuracy values are on a 0--1 scale.}
\label{tab:lexeme_pos_breakdown_combined_ud_ru_syntagrus_dataset}
\end{minipage}
\end{table*}

\begin{table*}[htbp]
\small
\centering
\renewcommand\arraystretch{1.2}
\setlength{\tabcolsep}{3pt}
\begin{minipage}{\linewidth}\centering
\begin{tabular}{@{}lcccccc@{}}
    \toprule
    \multirow{2}{*}{\textbf{Model}} & Base & 3rd person & Positive & Past & Plural & Singular \\
     & \footnotesize (n=154) & \footnotesize (n=51) & \footnotesize (n=401) & \footnotesize (n=168) & \footnotesize (n=33) & \footnotesize (n=632) \\
    \midrule
    mT5-Base & 0.860 & 0.911 & 0.928 & 0.966 & 0.837 & 0.952 \\
    Qwen2.5-1.5B & 0.808 & 0.802 & 0.721 & 0.928 & 0.861 & 0.892 \\
    Qwen2.5-1.5B-Instruct & 0.809 & 0.817 & 0.720 & 0.941 & 0.878 & 0.899 \\
    Qwen2.5-7B & 0.865 & 0.879 & 0.810 & 0.966 & 0.903 & 0.909 \\
    Qwen2.5-7B-Instruct & 0.850 & 0.874 & 0.796 & 0.960 & 0.886 & 0.900 \\
    Goldfish Turkish & 0.847 & 0.915 & 0.880 & 0.964 & 0.872 & 0.963 \\
    \bottomrule
\end{tabular}
\caption{Breakdown of inflection classification accuracy for each model by inflection type using Linear Regression classifiers (Turkish). Accuracies are calculated over all examples for a given group across all layers. Counts (n) are derived from a single representative layer for each group. All accuracy values are on a 0--1 scale.}
\label{tab:inflection_breakdown_lr_ud_tr_imst_dataset}
\end{minipage}
\end{table*}

\begin{table*}[htbp]
\small
\centering
\renewcommand\arraystretch{1.2}
\setlength{\tabcolsep}{3pt}
\begin{minipage}{\linewidth}\centering
\begin{tabular}{@{}lcccccc@{}}
    \toprule
    \multirow{2}{*}{\textbf{Model}} & Base & 3rd person & Positive & Past & Plural & Singular \\
     & \footnotesize (n=154) & \footnotesize (n=51) & \footnotesize (n=401) & \footnotesize (n=168) & \footnotesize (n=33) & \footnotesize (n=632) \\
    \midrule
    mT5-Base & 0.755 & 0.760 & 0.848 & 0.922 & 0.515 & 0.949 \\
    Qwen2.5-1.5B & 0.770 & 0.767 & 0.667 & 0.919 & 0.765 & 0.914 \\
    Qwen2.5-1.5B-Instruct & 0.762 & 0.757 & 0.662 & 0.917 & 0.766 & 0.913 \\
    Qwen2.5-7B & 0.853 & 0.845 & 0.791 & 0.956 & 0.875 & 0.937 \\
    Qwen2.5-7B-Instruct & 0.845 & 0.844 & 0.786 & 0.956 & 0.875 & 0.932 \\
    Goldfish Turkish & 0.832 & 0.879 & 0.870 & 0.957 & 0.834 & 0.957 \\
    \bottomrule
\end{tabular}
\caption{Breakdown of inflection classification accuracy for each model by inflection type using Multi-Layer Perceptron (MLP) classifiers (Turkish). Accuracies are calculated over all examples for a given group across all layers. Counts (n) are derived from a single representative layer for each group. All accuracy values are on a 0--1 scale.}
\label{tab:inflection_breakdown_mlp_ud_tr_imst_dataset}
\end{minipage}
\end{table*}

\begin{table*}[htbp]
\small
\centering
\renewcommand\arraystretch{1.2}
\setlength{\tabcolsep}{3pt}
\begin{minipage}{\linewidth}\centering
\begin{tabular}{@{}lcccccccc@{}}
    \toprule
    \multirow{3}{*}{\textbf{Model}} & \multicolumn{4}{c}{\textbf{Linear Regression}} & \multicolumn{4}{c}{\textbf{MLP}} \\
    \cmidrule(lr){2-5} \cmidrule(lr){6-9}
     & Noun & Verb & Adjective & Other & Noun & Verb & Adjective & Other \\
     & \footnotesize (n=221) & \footnotesize (n=53) & \footnotesize (n=104) & \footnotesize (n=13) & \footnotesize (n=221) & \footnotesize (n=53) & \footnotesize (n=104) & \footnotesize (n=13) \\
    \midrule
    mT5-Base & 0.866 & 0.823 & 0.921 & 0.955 & 0.215 & 0.421 & 0.374 & 0.637 \\
    Qwen2.5-1.5B & 0.834 & 0.805 & 0.866 & 0.877 & 0.307 & 0.439 & 0.449 & 0.693 \\
    Qwen2.5-1.5B-Instruct & 0.816 & 0.791 & 0.860 & 0.874 & 0.305 & 0.439 & 0.448 & 0.691 \\
    Qwen2.5-7B & 0.871 & 0.850 & 0.900 & 0.904 & 0.595 & 0.625 & 0.695 & 0.809 \\
    Qwen2.5-7B-Instruct & 0.850 & 0.823 & 0.883 & 0.885 & 0.579 & 0.613 & 0.678 & 0.800 \\
    Goldfish Turkish & 0.929 & 0.904 & 0.940 & 0.969 & 0.386 & 0.550 & 0.477 & 0.808 \\
    \bottomrule
\end{tabular}
\caption{Breakdown of lemma classification accuracy by Part of Speech (POS) for each model, using Linear Regression and Multi-Layer Perceptron (MLP) classifiers (Turkish). Lemmas are grouped by their POS tags (\eg Noun, Verb, Adjective). Accuracies are calculated over all examples for a given group across all layers. Counts (n) are derived from a single representative layer for each group. All accuracy values are on a 0--1 scale.}
\label{tab:lexeme_pos_breakdown_combined_ud_tr_imst_dataset}
\end{minipage}
\end{table*}

\end{document}